%% file: main.tex
\pdfoutput=1
\documentclass[sigconf, screen]{acmart}

\usepackage[linesnumbered,ruled,vlined]{algorithm2e}
\usepackage{subcaption}
\newcommand{\method}{ReForge}
\usepackage{listings}
\lstdefinestyle{policy}{basicstyle=\ttfamily\scriptsize, columns=fullflexible, keepspaces=true, breaklines=true, frame=single, framesep=3pt, xleftmargin=4pt, xrightmargin=4pt}

\usepackage{booktabs}
\usepackage{siunitx}
\usepackage{wrapfig}

\begin{document}

\title{ReForge: Keeping ABR Algorithms Never Finished with Verified Large Language Model Revisions}

\author{Zhiqiang He}
\email{hezhiqiang@ieee.org}
\affiliation{%
  \institution{The University of Electro-Communications}
  \city{Tokyo}
  \country{Japan}
}

\author{Zhi Liu}
\authornote{Corresponding author}
\email{liu@ieee.org}
\affiliation{%
  \institution{The University of Electro-Communications}
  \city{Tokyo}
  \country{Japan}
}

\renewcommand{\shortauthors}{He et al.}

\begin{abstract}
	Designing an ABR algorithm for a network scenario takes an engineer months, and large language models (LLMs) now do this work in hours. But either way, the algorithm is fixed once it is deployed. It fits the networks seen during its design and falls behind on the networks that arrive afterwards. We ask whether an ABR algorithm can keep being designed after deployment, revised as each new network arrives, without becoming worse on any network it already serves. We propose ReForge, which keeps a pool of existing ABR policies unchanged and lets an LLM revise only the rules that decide which policy chooses the next bitrate. When a new network arrives, the LLM reads how the current rules and every policy perform on the networks served so far and rewrites the rules. The rewrite is deployed only if a replay in simulation shows that it improves the served networks as a whole and makes none of them worse. A fast GPU simulator that replays every episode exactly makes this acceptance test cheap enough to run for every revision. We evaluate ReForge on nine real-world network datasets, with new networks arriving one at a time. The revisions raise the mean QoE from 1.43 to 1.79, past the best single policy at 1.66 and close to the 1.86 of selecting the best policy for each dataset, while the QoE on the networks served first does not drop. The gain also holds in a real video player. All code, data, and records will be available.
\end{abstract}




\begin{teaserfigure}
	\includegraphics[width=\textwidth]{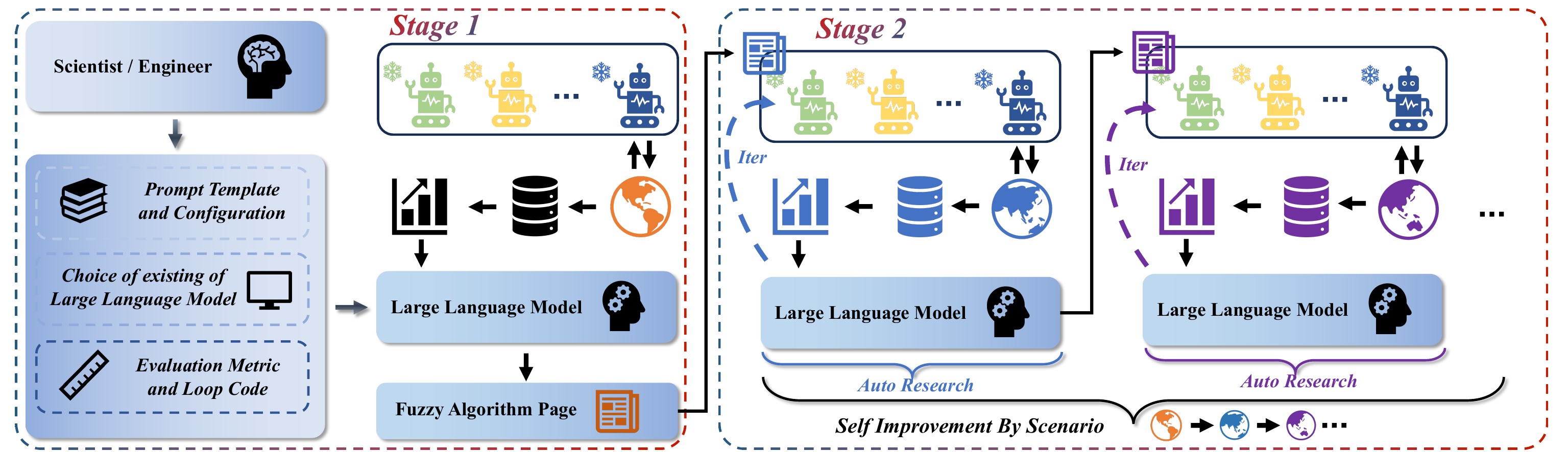}
	\caption{The Framework of ReForge.}
	\label{fig:framework}
\end{teaserfigure}

\maketitle

\section{Introduction}

In a single decade, adaptive bitrate (ABR) algorithms have gone through three generations. Engineers hand-built controllers from buffer and throughput models~\cite{huang2014buffer, spiteri2020bola, yin2015control} and from fuzzy logic~\cite{vergados2015fdash}. Reinforcement learning then trained policies end to end~\cite{mao2017neural, kan2025merina+, 11498444}. Now LLMs write ABR algorithms themselves~\cite{he2024designing, jia2026crucible}. NADA has an LLM draft candidate designs for a learned policy, then trains and filters them~\cite{he2024designing}. Crucible has an LLM tune the logic and parameters of existing controllers, once per scenario~\cite{jia2026crucible}. LLM algorithm design in other fields follows the same pattern~\cite{romera2024mathematical, karimi2025robustheuristicalgorithmdesign, hamadanian2026glia, karimi2026improving}. An LLM writes candidates, a search loop scores and refines them, and the best candidate at the end of the budget is delivered. The designer keeps getting better, but in all three generations the algorithm is fixed at delivery, while the networks it serves keep changing. An algorithm fitted to the networks seen during design therefore falls behind on the networks that arrive later~\cite{liu2024systematic}.

Two remedies exist, and neither removes this limitation. Generalization trains one policy on as many networks as possible in advance~\cite{mao2017neural, kan2025merina+}, but its performance still falls on networks outside the training data. Continual learning retrains the policy on new data~\cite{mccloskey1989catastrophic}. Retraining changes all weights of the policy, so its decisions on earlier networks change as well, and methods against this forgetting can reduce it~\cite{kirkpatrick2017overcoming, zenke2017continual, rolnick2019experience, buzzega2020dark, rusu2016progressive} but cannot rule it out. LLM agents that improve themselves change the artifact the agent produces~\cite{novikov2025alphaevolve}, the harness around the model~\cite{zhang2026self}, or the model weights~\cite{zhao2026absolute}. Those that change the artifact, however, stop doing so at delivery~\cite{jia2026crucible}.

We take a different view. The design of an ABR algorithm should continue after deployment, and every revision should be checked on all networks served so far before it is deployed, in the same way that a software change is deployed only after it passes a regression test. Two conditions make this practical. First, the part that is revised must be small and easy to read, so that a revision can improve a new network while leaving the behavior on earlier networks untouched. Second, this acceptance test must be fast and precise, since it runs for every revision and must tell small improvements apart from noise.

\method{} meets both conditions. It keeps five existing ABR policies unchanged as experts and revises only a small set of IF--THEN rules that decide, from recent throughput and buffer level, which expert chooses the next bitrate. When a new network arrives, \method{} records a few episodes on it and measures how every expert and the current rules perform. An LLM reads these measurements and rewrites the rules. The rewritten rules are then replayed in simulation on every network served so far, and they are deployed only if they improve these networks as a whole and make none of them worse. The acceptance test is cheap because our GPU simulator plays thousands of episodes in under a second and replays every episode exactly, so the current and the rewritten rules are compared on identical episodes.

We evaluate \method{} on nine real-world network datasets, starting from five 3G datasets and letting four further networks arrive one at a time. The revisions raise the mean QoE over the nine datasets from 1.43 to 1.79, past the best single expert at 1.66 and close to the 1.86 of selecting the best expert for each dataset (Table~\ref{tab:main}). The revised rules keep their lead over the best single expert when they control a real hls.js player (Section~\ref{sec:eval-real}).

Our contributions are as follows.
\begin{itemize}
	\item \textbf{Continual ABR algorithm design.} Existing ABR design, by hand, by training, or by LLM, stops at delivery. We formulate algorithm design itself as a process that continues while the algorithm serves, with networks arriving sequentially.
	\item \textbf{\method{}.} We propose a framework that keeps the costly components of an algorithm fixed and lets an LLM keep rewriting a small readable part, and that deploys a rewrite only after verifying it on every network served so far. No earlier network can thus fall by more than a fixed margin, and the deployed algorithm needs no LLM at run time.
	\item \textbf{A GPU simulator for ABR.} We build an ABR simulator that runs thousands of environments in parallel on a GPU and produces exactly the same episodes as the CPU simulator. It evaluates 8192 episodes 415 times faster than the CPU simulator, which makes it affordable to verify every revision.
	\item \textbf{Verification prevents forgetting, and the LLM makes revision efficient.} On nine real-world datasets, \method{} raises the mean QoE from 1.43 to 1.79 without lowering earlier networks, while an MLP gate retrained without verification forgets, and revisers without an LLM end lower under the same verification. In a real video player, \method{} keeps its lead over the best single expert.
\end{itemize}

\section{Motivation}
\label{sec:motivation}

Two observations led to the design of \method{}. No single ABR policy is the best on every network, and a selection among policies that is fixed in advance falls behind when new networks arrive.

\subsection{No single policy is the best}
\label{sec:motivation-grid}

\begin{figure}[!htbp]
	\centering
	\includegraphics[width=\columnwidth]{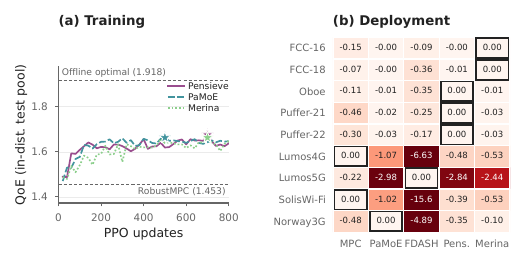}
	\caption{Trained on 3G networks, deployed on many. (a)~QoE of the three learned agents on 3G test traces during training, with the checkpoint used marked by a star and RobustMPC and the offline optimum dashed. (b)~Regret of each policy on nine datasets, i.e.\ its QoE minus that of the best policy on the dataset (boxed zero). The top five rows are from the training distribution. Details in Appendix~\ref{app:train-deploy}.}
	\label{fig:train-deploy}
\end{figure}

Learned ABR policies are trained on one distribution of network conditions but deployed on many. Figure~\ref{fig:train-deploy}(a) shows that, on test traces from the training distribution, three learned agents with different architectures (Pensieve, Merina and PaMoE) reach almost the same QoE, between 0.210 and 0.222 above RobustMPC. Judged on this test set, the choice of agent hardly matters. Figure~\ref{fig:train-deploy}(b) deploys these agents, together with the model-based controller RobustMPC and the fuzzy-logic controller FDASH, on nine trace families. On the five families of the training distribution an agent is always the best policy. On the four families from other datasets the ranking breaks up. RobustMPC is the best policy on Lumos4G and SolisWi-Fi, FDASH on Lumos5G and PaMoE on Norway3G, and on Lumos5G the best agent trails the best policy by 2.441. No single policy is the best across deployment conditions, and a test set drawn from the training distribution does not reveal this. We therefore keep the five policies as experts and choose among them according to the network. Every bitrate is then still chosen by one of the experts.

\subsection{A selection fixed in advance falls behind}
\label{sec:motivation-oneshot}

The choice among the experts can be written as IF--THEN rules over throughput and buffer statistics, and an LLM can write such rules from measurements of the experts on the networks known so far. Rules written in this way from the 3G training data come close to the best expert on the 3G datasets, but once new networks arrive they fall below running RobustMPC alone over the nine datasets (1.427 against 1.664, Table~\ref{tab:main}). Outside the 3G datasets, all nine policies that three LLMs wrote in three forms fall far below RobustMPC alone (Appendix~\ref{app:llm-res-ood}). The cause is not how the rules were written but when. At that time the measurements contained no 4G, 5G or Wi-Fi network, and no prompt can supply facts that have not yet been observed. Retraining a learned selector such as an MLP gate on each new network is no remedy, because it also changes the choices on earlier networks and makes them worse (Section~\ref{sec:eval-past}). The rules must therefore be revised as new networks arrive, and every revision must be checked on the earlier networks before it is deployed.

\section{\method{}}
\label{sec:system}

\method{} keeps a pool of ABR policies fixed and lets an LLM write and revise the rules that select among them (Figure~\ref{fig:framework}). In Stage~1 the LLM writes the first rules from measurements on the networks known at deployment. In Stage~2, each time a new network arrives, the LLM revises the rules from a few recorded episodes, and an acceptance test in simulation decides whether the revision is deployed. This section describes the rules (Section~\ref{sec:firstpage}), the evidence the LLM receives (Section~\ref{sec:evidence}), the revision and its acceptance test (Section~\ref{sec:revise}), and the simulator that makes the test affordable (Section~\ref{sec:gpusim}).

\subsection{Experts and Rules}
\label{sec:firstpage}

The pool $\mathcal{M}$ holds five experts, the learned policies Pensieve~\cite{mao2017neural}, Merina~\cite{kan2025merina+} and PaMoE~\cite{11498444}, trained on 3G traces to convergence (Appendix~\ref{app:converge}), and the controllers RobustMPC~\cite{yin2015control} and FDASH~\cite{vergados2015fdash}. Before each chunk the rules select one expert, and that expert chooses the bitrate. The rules are a JSON document with four parts, all of which the LLM writes. \emph{Features} are statistics of recent chunks, such as the mean throughput over the last five chunks, built from a fixed list of operators. We write $x_j$ for the current value of feature $j$. \emph{Intervals} give each feature named states $l$ such as \texttt{Low} and \texttt{High}. State $l$ of feature $j$ has a trapezoidal membership function $\phi_{j,l}$, and its value $\phi_{j,l}(x_j) \in [0,1]$ is the degree to which $x_j$ is in state $l$, so that a value near a boundary belongs partly to both neighboring states~\cite{vergados2015fdash}. \emph{Rules} have the form IF $x_{j_1}$ is $l_1$ and $x_{j_2}$ is $l_2$ THEN expert $m$. For a rule $q$, we write $(j,l) \in q$ for each of its conditions ``$x_j$ is $l$'' and $m_q$ for the expert it names. The rule fires with the product of the membership degrees of its conditions. Each expert $m$ receives as its score $w_m$ the firing strength of its strongest rule, and the expert $m^\star$ with the highest score is selected,
\begin{equation}
	w_m = \max_{q\,:\,m_q = m} \ \prod_{(j,l) \in q} \phi_{j,l}(x_j), \qquad m^\star = \arg\max_{m \in \mathcal{M}} w_m .
	\label{eq:score}
\end{equation}
\emph{Additional parameters} limit switching between experts and handle the start of an episode. A new expert replaces the current one only if its score is higher by a margin and the current one has been used for a minimum number of chunks, and a default expert plays the first chunks, before the statistics are meaningful. We call rules of this form Form~I. Two further forms let the features also read the experts' proposals (Form~II) or let the LLM write the selection as Python code (Form~III), and are defined in Appendix~\ref{app:llm-forms}. Before any rules are used, a program checks that every feature, state and expert they refer to exists and that every membership function is well formed (Appendix~\ref{app:llm-check}).

This form keeps the part that is revised small and readable. A rule set has a handful of rules, so a human can read what each revision changed, and a revision can add a rule for a new network while leaving the others as they are. The deployed algorithm consists only of the rules and the five experts, and evaluating the rules takes a few arithmetic operations per chunk, so no LLM is called at run time. In Stage~1, three LLMs, Claude Opus~4.7, gpt-oss-120b and Qwen3.8-27B, each wrote rules in all three forms from the same measurements of the experts on the five 3G training sets. Only Opus~4.7 followed the measurements in every form, while the other two wrote rules that the measurements do not support or code that does not compute what it states (Appendix~\ref{app:llm-results}). We therefore use Opus~4.7 with Form~I. Its rules select Pensieve when the mean throughput of the last five chunks is not very high and Merina when it is very high (prompt and answer in Appendix~\ref{app:prompt}).

\subsection{Evidence from a New Network}
\label{sec:evidence}

When a new network arrives, the deployed rules play 16 episodes on it, and for each episode we record the trace and the time in the trace at which it starts. Any expert or rule set can then be replayed on the same episodes, meeting the same link capacity. The download time of every chunk also contains random delay noise, so each recorded episode is replayed 512 times with different noise, and the results are averaged (Appendix~\ref{app:continual-replay}). The 16 episodes are taken from the training split of the network, and its test split is used only for evaluation.

From these replays and from the measurements made on the 3G training sets in Stage~1, the LLM receives tables for every network served so far (Appendix~\ref{app:continual-evidence}). They give the QoE, rebuffering and bitrate change of each expert alone, the distribution of throughput and of the rule features, the QoE of the current rules together with the experts they select, and on how many of the 16 episodes each expert is the best. The last table shows whether an expert's lead rests on most episodes or on a few. The new networks are named Network-1 to Network-4 in the prompt, so that the revision rests on the measurements rather than on what the LLM may know about a dataset. A log of the earlier rounds lists each candidate, the reason the LLM gave for it, and the result of its acceptance test. The complete prompt of one round is given in Appendix~\ref{app:roundprompt}.

\subsection{Revision and Acceptance}
\label{sec:revise}

In each round the LLM writes a complete new rule set, which must pass the same validity check as in Stage~1. We denote this candidate rule set by $\pi'$ and the currently deployed rules by $\pi$. After the $s$-th new network has arrived, the router serves the set of entries $\mathcal{E}_s$, which contains each of the five 3G datasets and each of the $s$ arrived networks as one entry $u$. The candidate is compared with the deployed rules on every entry $u \in \mathcal{E}_s$. Let $\mathrm{QoE}(\pi, u)$ be the mean QoE on entry $u$ in a validation simulation of the training data, and $\mathrm{QoE}^{\mathrm{best}}_u$ the highest value any deployed rules have reached on $u$. The candidate passes the acceptance test, and is deployed, if and only if
\begin{align}
	\frac{1}{|\mathcal{E}_s|} \sum_{u \in \mathcal{E}_s} \mathrm{QoE}(\pi', u) &\ \ge\ \frac{1}{|\mathcal{E}_s|} \sum_{u \in \mathcal{E}_s} \mathrm{QoE}(\pi, u) + \varepsilon, \label{eq:gain}\\
	\mathrm{QoE}(\pi', u) &\ \ge\ \mathrm{QoE}^{\mathrm{best}}_u - \varepsilon \quad \text{for every } u \in \mathcal{E}_s, \label{eq:floor}
\end{align}
with $\varepsilon = 0.01$. Condition~\eqref{eq:gain} requires a real gain over all entries, and condition~\eqref{eq:floor} forbids forgetting, since no entry may fall more than $\varepsilon$ below the best it has had. The validation simulation uses a different random seed from the tables the LLM read, so that a candidate tuned to the particular noise of those tables is not rewarded. A rejected candidate is discarded and the deployed rules stay. A stage ends after five rounds or after two consecutive rejections. Algorithm~\ref{alg:cont} in Appendix~\ref{app:continual-revise} summarizes the procedure.

\subsection{Fast and Exact Simulation}
\label{sec:gpusim}

A loop of revisions is practical only if every trial is cheap to repeat~\cite{xiao2026enpire}. Each candidate is evaluated on $8192$ episodes of every entry, so the acceptance test must be fast, and it must be precise enough to resolve $\varepsilon = 0.01$. Our simulator implements the trace-driven ABR environment of Pensieve~\cite{mao2017neural} (Appendix~\ref{sec:system-model}) on the GPU, stepping thousands of environments at once as tensors (Appendix~\ref{sec:gpu_sim}). Every random quantity, i.e.\ the trace, the start time and the delay noise of each chunk, is obtained by hashing the seed together with the indices of the environment, the episode and the chunk. The GPU simulator therefore produces exactly the same episodes as the CPU simulator, and any episode can be replayed. It evaluates 8192 episodes in 0.53\,s, against 220\,s on the CPU, 415 times faster, and it also trains the learned experts 3.75 times faster (Table~\ref{tab:gpu_summary}). The mean QoE of a single policy over 8192 episodes varies by 0.026 between seeds (Table~\ref{tab:eval_noise}), which exceeds $\varepsilon$. The acceptance test, however, compares the candidate and the deployed rules on identical episodes with identical noise, so this variation affects both alike and largely cancels in their difference.

\section{Evaluation}
\label{sec:eval}

\begin{figure*}[t]
	\centering
	\includegraphics[width=\textwidth]{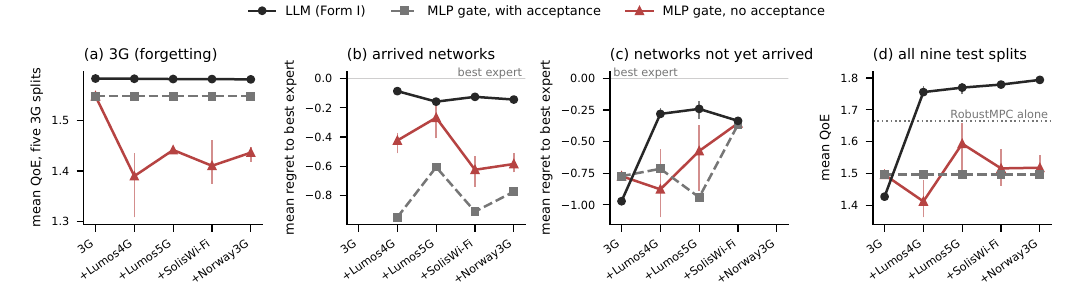}
	\caption{QoE on the test splits after each arrival, mean over the three repetitions with their range as whiskers. (a)~Mean QoE on the five 3G datasets. (b)~Mean regret on the networks that have arrived. (c)~Mean regret on the networks that have not yet arrived. (d)~Mean QoE on all nine datasets, with RobustMPC alone dotted. LLM (Form~I) is \method{}. Regret is the QoE minus that of the best expert on a dataset.}
	\label{fig:contmain}
\end{figure*}

\begin{figure*}[t]
	\centering
	\includegraphics[width=\textwidth]{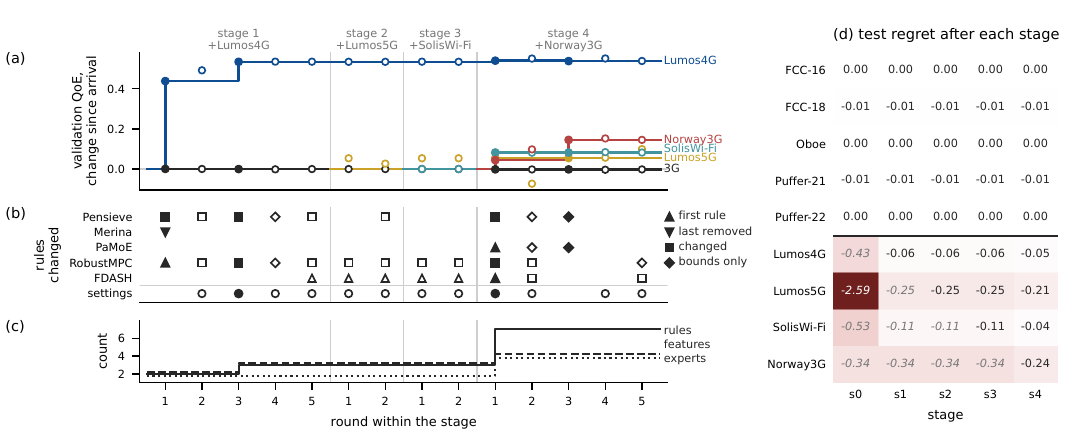}
	\caption{The 14 rounds of \method{} in one repetition. (a)~Validation QoE in the acceptance test after each round, as the change since the network arrived. Filled markers are accepted candidates and hollow markers rejected ones. (b)~Experts whose rules a candidate changed (markers explained in Appendix~\ref{app:continual-run}). (c)~Number of rules, features and selected experts. (d)~Regret on the test splits after each stage.}
	\label{fig:cont-runmain}
\end{figure*}

\subsection{Setup}
\label{sec:setup}

\textbf{Networks.} The router starts on five 3G datasets, FCC-16 and FCC-18~\cite{fcc2016rawdata}, Oboe~\cite{akhtar2018oboe}, and Puffer-21 and Puffer-22~\cite{yan2020learning}, on whose training splits the learned experts were trained and the first rules were written. Four networks then arrive one at a time, Lumos4G and Lumos5G~\cite{narayanan2021variegated}, SolisWi-Fi~\cite{GreenLv} and Norway3G~\cite{riiser2013commute}. Every dataset has disjoint training and test splits. The revisions use only the 16 episodes recorded on the training split of each arriving network, and after every arrival the rules are evaluated on the test splits of all nine datasets, with 8192 episodes per dataset. QoE is the standard linear form, the bitrate minus penalties for rebuffering and bitrate changes, averaged over chunks~\cite{yin2015control}. The LLM is Claude Opus~4.7, selected in Stage~1 (Section~\ref{sec:firstpage}). The evidence tables, the acceptance test and the final evaluation each use their own random seed, and the whole procedure is repeated three times with different recorded episodes (Appendix~\ref{app:continual-replay}).

\textbf{Baselines.} The classical controllers RateBased~\cite{li2014probe}, BBA~\cite{huang2014buffer} and BOLA~\cite{spiteri2020bola} and the five experts are each used alone. An MLP gate, a small network trained to predict the return of each expert (Appendix~\ref{app:llm-mlp-gate}), is retrained on the same data in every round, once with the acceptance test and once deployed without it. Three revisers without an LLM replace the LLM under the same acceptance test (Appendix~\ref{app:continual-baselines}). Evolutionary search mutates the rules and keeps the fittest candidate, a network lookup assigns each network the expert that is best on it, and a value tree splits on throughput statistics to select an expert. Two references bound what is achievable, selecting the best expert for each dataset and the offline optimum, the best bitrate sequence of each test episode computed with the whole episode known in advance (Appendix~\ref{app:continual-eval}).

\subsection{Does \method{} keep improving?}
\label{sec:eval-main}

\begin{table}[t]
	\caption{Test QoE. 3G is the mean over the five 3G datasets, and All the mean over all nine datasets. Superscripts 1 to 4 give the order in which the new networks arrive. Revised methods are shown after the last stage, as the mean over the three collection seeds. Bold marks the best value in each column, references excluded.}
	\label{tab:main}
	\footnotesize
	\setlength{\tabcolsep}{2.5pt}
	\begin{tabular}{@{}lrrrrrr@{}}
		\toprule
		Method & 3G & L4G\textsuperscript{1} & L5G\textsuperscript{2} & Wi-Fi\textsuperscript{3} & N3G\textsuperscript{4} & All \\
		\midrule
		\multicolumn{7}{@{}l}{\textit{Classical controllers, not in the expert pool}} \\
		\quad RateBased & 0.957 & 2.680 & 3.905 & 1.938 & $-0.912$ & 1.377 \\
		\quad BBA & 1.117 & 2.263 & 3.674 & 1.333 & $-0.418$ & 1.382 \\
		\quad BOLA & $-0.484$ & 2.543 & 3.691 & 1.820 & $-0.874$ & 0.529 \\
		\addlinespace
		\multicolumn{7}{@{}l}{\textit{Experts, each used alone}} \\
		\quad Pensieve & \textbf{1.584} & 2.291 & 1.179 & 1.767 & $-0.465$ & 1.410 \\
		\quad Merina & 1.572 & 2.242 & 1.583 & 1.623 & $-0.224$ & 1.454 \\
		\quad PaMoE & 1.573 & 1.698 & 1.051 & 1.139 & $\mathbf{-0.130}$ & 1.291 \\
		\quad RobustMPC & 1.368 & \textbf{2.766} & 3.808 & \textbf{2.156} & $-0.592$ & 1.664 \\
		\quad FDASH & 1.342 & $-3.823$ & \textbf{4.026} & $-13.481$ & $-4.983$ & $-1.284$ \\
		\addlinespace
		\multicolumn{7}{@{}l}{\textit{Selecting among the experts, not revised}} \\
		\quad One-shot LLM rules & 1.583 & 2.336 & 1.433 & 1.626 & $-0.465$ & 1.427 \\
		\quad MLP gate, with acceptance & 1.548 & 1.818 & 3.763 & 0.638 & $-0.495$ & 1.496 \\
		\addlinespace
		\multicolumn{7}{@{}l}{\textit{Selecting among the experts, revised without the acceptance test}} \\
		\quad MLP gate, no acceptance & 1.436 & 2.134 & 3.815 & 0.942 & $-0.410$ & 1.518 \\
		\addlinespace
		\multicolumn{7}{@{}l}{\textit{Selecting among the experts, revised with the acceptance test}} \\
		\quad Evolutionary search & 1.580 & 2.561 & 3.813 & 2.131 & $-0.465$ & 1.771 \\
		\quad Network lookup & 1.580 & 2.722 & 3.803 & 1.887 & $-0.468$ & 1.761 \\
		\quad Value tree & 1.583 & 2.463 & 2.924 & 1.857 & $-0.465$ & 1.633 \\
		\quad \method{}, LLM (Form~I) & 1.581 & 2.701 & 3.829 & 2.117 & $-0.405$ & \textbf{1.794} \\
		\midrule
		\multicolumn{7}{@{}l}{\textit{References}} \\
		\quad Best expert per dataset & 1.587 & 2.766 & 4.026 & 2.156 & $-0.130$ & 1.861 \\
		\quad Offline optimum & 1.815 & 3.295 & 4.209 & 2.371 & 0.151 & 2.123 \\
		\bottomrule
	\end{tabular}
\end{table}

Yes. \method{} starts from the one-shot LLM rules, and over the nine test splits it raises the mean QoE from their 1.427 to 1.794 (column All of Table~\ref{tab:main}). The three repetitions end between 1.791 and 1.798 (Table~\ref{tab:cont-res} in Appendix~\ref{app:continual-res}). The mean rises at every arrival in which a revision is accepted and never falls (Figure~\ref{fig:contmain}d). It ends 0.13 above RobustMPC, the best single expert, and at 96\% of the 1.861 of selecting the best expert for each dataset. The distance to the offline optimum of 2.123 falls from 0.70 to 0.33, i.e.\ by about half. Table~\ref{tab:main} also shows why selecting among experts pays off. The best method changes from column to column, and no expert is close to the best on every network, whereas \method{} is close to the best on most. It stays clearly behind only on Lumos5G, where FDASH obtains 4.026 and RateBased, which is not in the pool, 3.905, and on Norway3G, where PaMoE is the best expert.

Revisions made for one network also help networks that have not yet arrived. The revisions of the first stage, made from 16 Lumos4G episodes, raised Lumos5G from 1.433 to 3.760 and SolisWi-Fi from 1.626 to 1.916 before either network arrived, so the regret on the networks yet to arrive shrinks sharply after the first stage (Figure~\ref{fig:cont}c, Appendix~\ref{app:continual-res}).

\subsection{Do earlier networks get worse?}
\label{sec:eval-past}

No. The mean QoE of \method{} on the 3G datasets is 1.583 before the first arrival and 1.581 after the last, and in no repetition does it move by more than 0.003 at any stage (Table~\ref{tab:main}, Figure~\ref{fig:contmain}a). Condition~\eqref{eq:floor} contributed to this by rejecting four LLM candidates that lowered an earlier network. The MLP gate behaves differently. Each of its 24 retrained versions improved the network that had just arrived but lowered at least one 3G dataset more than $\varepsilon$ below its best, so none passed the acceptance test and it stayed at 1.496. Deployed without acceptance, it drops from 1.548 to 1.436 on the 3G datasets and ends at 1.518 overall, still below RobustMPC alone.

\subsection{What does the LLM contribute?}
\label{sec:eval-llm}

To see what the LLM contributes, we replace it with the three revisers of Section~\ref{sec:setup}, under the same acceptance test and with the same evidence (Appendix~\ref{app:continual-baselines}). The network lookup keeps one row per network, which holds the expert that is best on that network, and plays for each episode the row whose throughput statistics are nearest. The value tree is a decision tree on throughput statistics, fitted to the expert rollouts of all networks. Evolutionary search makes random changes to the deployed rules and submits the best changed rule set. All three end below \method{}, at 1.761, 1.633 and 1.771 (Table~\ref{tab:main}). The lookup and the tree fail on the earlier networks. The tree is refitted from scratch at every arrival, so it also routes the 3G datasets differently. The lookup adds a row for each new network, and some episodes of an earlier network lie closer to this row and are given its expert. Most of their candidates therefore violate condition~\eqref{eq:floor}, 18 of 24 rejected candidates of the lookup and 23 of 24 of the tree. Each was accepted in only 3 of 27 rounds, the lookup only at the first arrival and the tree only at the first two. Evolutionary search changes the deployed rules only slightly and so keeps the earlier networks, but its gains become too small. At the last two arrivals none of its candidates passed condition~\eqref{eq:gain}, although it simulates about 63 rule sets in every round. The LLM writes a single candidate per round, and 12 of its 40 candidates were accepted, including one at the last arrival in every repetition. A smaller LLM gives smaller gains, 1.782 with Sonnet and 1.739 with Haiku (Appendix~\ref{app:continual-models}). The acceptance test thus makes revision safe for every reviser, and the LLM makes it efficient. Results for other arrival orders and other forms of the rules are given in Appendices~\ref{app:continual-order} and~\ref{app:continual-forms}.

\subsection{What does a revision cost?}
\label{sec:eval-cost}

A round of \method{} takes 2.6 minutes, most of it spent waiting for the LLM, and its acceptance test takes about 40 seconds (Table~\ref{tab:cost}). Evolutionary search needs 32 times as long per round, because it scores about 63 rule sets for each candidate, and still ends lower. All these times rely on the GPU simulator. The CPU simulator is 415 times slower (Section~\ref{sec:gpusim}), so on it a single acceptance test of about 40 seconds would take about 4.6 hours, for every reviser.

\begin{table}[!h]
	\caption{Revisers' cost per round, mean over all rounds of the three repetitions. Time is wall-clock time on one RTX A6000, including the acceptance test. Simulated is the number of rule sets or models scored in simulation to produce one candidate.}
	\label{tab:cost}
	\footnotesize
	\setlength{\tabcolsep}{3pt}
	\begin{tabular}{@{}lrrrr@{}}
		\toprule
		Reviser & Time (min) & Simulated & Accepted & All \\
		\midrule
		\method{}, LLM (Form~I) & 2.6 & 1 & 12 of 40 & \textbf{1.794} \\
		Evolutionary search & 83.2 & 63 & 7 of 32 & 1.771 \\
		Network lookup & 0.8 & 1 & 3 of 27 & 1.761 \\
		Value tree & 1.1 & 1 & 3 of 27 & 1.633 \\
		MLP gate, retrained & 12.8 & 1 & 0 of 24 & 1.496 \\
		\bottomrule
	\end{tabular}
\end{table}

\subsection{What happens in one repetition?}
\label{sec:eval-anatomy}

Figure~\ref{fig:cont-runmain} follows one repetition round by round. Panel~(a) shows the validation QoE of each candidate, filled if the acceptance test accepted it and hollow if it rejected it. Panel~(b) shows which experts each candidate changed, panel~(c) the size of the rules, and panel~(d) the regret on the test splits after each stage. At the first arrival, the first candidate moves the high-throughput range from Merina to RobustMPC (panel~b), which raises Lumos4G while the 3G datasets stay flat (panel~a). At the next two arrivals, every candidate gives FDASH a rule aimed at the very high throughput of Lumos5G. Each raises only Lumos5G, and by too little to satisfy condition~\eqref{eq:gain} once the gain is averaged over all entries, so all four are rejected. At the last arrival, the LLM combines this rule with two changes aimed at SolisWi-Fi and Norway3G, so that the gains add up. This candidate is accepted, and the rules grow from three to seven (panel~c). The rounds and the notes of the LLM are described in Appendix~\ref{app:continual-run}.

\subsection{Does it hold in a real player?}
\label{sec:eval-real}

We also let the final rules control an unmodified hls.js player in a browser, with every segment throttled in real time by a test trace \cite{he2026nsma} (Appendix~\ref{app:realworld}). On three traces of each of the nine datasets, \method{} obtains a mean QoE of 2.519, above RobustMPC with 2.309, the one-shot rules with 2.264 and Pensieve with 1.856. This is the order that the simulator gives for the same 108 episodes. \method{} plays about the same bitrate as RobustMPC but stalls for 0.7 seconds per episode instead of 1.9. The values are higher than in Table~\ref{tab:main} because the player uses one video on all datasets, whereas the simulation of Table~\ref{tab:main} uses a video with larger segments on the four arriving networks.

\section{Conclusion and Discussion}
\label{sec:limits}

ABR algorithms have so far been designed before deployment and then left unchanged, while the networks they serve keep changing. \method{} continues the design after deployment. It keeps the experts fixed, lets an LLM rewrite the rules that select among them whenever a new network arrives, and deploys a rewrite only after a fast simulation shows that no network served so far becomes worse. With four networks arriving one at a time, this raised the mean QoE over nine datasets from 1.43 to 1.79 without lowering the networks served first, and the revised rules kept their lead in a real hls.js player. The GPU simulator evaluates 8192 episodes in about half a second, so a candidate can be tested on every network served so far within seconds. Together with the acceptance test, this makes the framework a basis for future work in which LLMs improve ABR algorithms recursively, revising their own earlier revisions, or discover new ABR algorithms.

Two limits remain. First, \method{} can only select among its experts. Even selecting the best expert for each dataset reaches only 1.861, while the offline optimum is 2.123, and a controller outside the pool such as RateBased can be better on a network such as Lumos5G. Second, the replay uses the link capacity of the recorded episodes. A trace-driven simulator provides it, and a real deployment can estimate it from the download logs of the episodes with causal method~\cite{alomar2023causalsim}, at the cost of an approximate replay. The first limit points to a larger revision space. Besides the rules, the LLM could tune the parameters of the experts, add a controller such as RateBased to the pool when the rules can no longer close the gap, or define new features from what the player observes. Every such revision would pass the same acceptance test, and the fast simulator keeps this test affordable as the space grows.

\bibliographystyle{ACM-Reference-Format}
\bibliography{main}

\clearpage
\appendix

\onecolumn
\newpage

\section{Environment}

\subsection{System Model}
\label{sec:system-model}

\paragraph{\textbf{Video model:}} Following the adaptive video streaming model \cite{yin2015control}, a video is divided into $I$ consecutive chunks $\{V_1, V_2, \dots, V_I\}$, each with a fixed playback duration of $L$ seconds. Each chunk is encoded at $M$ bitrate levels $\mathbb{B} = \{b^{(1)}, b^{(2)}, \dots, b^{(M)}\}$ in ascending order, and $d_i(b)$ denotes the size (in bytes) of chunk $V_i$ encoded at bitrate $b \in \mathbb{B}$. For each chunk, the agent selects a bitrate $b_i \in \mathbb{B}$ to maximize the QoE.

\paragraph{\textbf{Network model:}} 
The network is driven by a dataset of bandwidth traces $\mathcal{D} = \{\mathcal{T}^{(1)}, \dots, \mathcal{T}^{(N)}\}$. At the start of each episode, a trace $\mathcal{T} = \{(t_k, c_k)\}_{k=1}^{K}$ is sampled uniformly from $\mathcal{D}$, where $t_0 = 0$ marks the start of the trace and $c_k$ (Mbps) is the link capacity during $(t_{k-1}, t_k]$. To diversify the network conditions experienced across episodes, the episode does not necessarily start from the beginning of the trace: the initial trace time $\tau_1$ is drawn uniformly from $\{t_0, t_1, \dots, t_{K-2}\}$. Within an episode, the sampled trace is replayed cyclically with period $T = t_K$ if the episode outlasts it. Since only a fraction $\rho$ of each packet carries payload, the download speed (bytes/s) is
\begin{equation}
  S(t) = \rho \cdot \frac{10^{6}}{8}\, c_k, \quad t \in (t_{k-1}, t_k],
  \qquad S(t + T) = S(t).
\end{equation}
where the factor $10^{6}/8$ converts the link capacity from Mbps to bytes/s, consistent with the chunk size $d_i(b)$ measured in bytes. Let $\tau_i$ denote the trace time at which the download of $V_i$ starts. The transfer time $T_i^{\mathrm{tr}}$ is the time required to receive $d_i(b_i)$
bytes over the time-varying link. Since a download may span multiple trace
intervals, let $\Delta_{i,k}$ denote the portion of the download interval
$[\tau_i, \tau_i + T_i^{\mathrm{tr}}]$ that falls in the $k$-th trace interval
(with cyclic replay). Then $T_i^{\mathrm{tr}}$ is determined by
\begin{equation}
  \sum_{k} S_k \, \Delta_{i,k} = d_i(b_i),
  \qquad
  T_i^{\mathrm{tr}} = \sum_{k} \Delta_{i,k},
\end{equation}
where $S_k = \rho \cdot \frac{10^{6}}{8}\, c_k$. The effective throughput is thus
the time-weighted average speed over the intervals traversed by the download:
\begin{equation}
  C_i = \frac{d_i(b_i)}{T_i^{\mathrm{tr}}}
      = \frac{\sum_{k} S_k \, \Delta_{i,k}}{\sum_{k} \Delta_{i,k}} .
\end{equation}
To account for protocol overhead and network variability, the download delay perceived by the player includes the round-trip time $t_{\mathrm{RTT}}$ and a multiplicative noise factor:
\begin{equation}
  t_i^{\mathrm{delay}} = \left( T_i^{\mathrm{tr}} + t_{\mathrm{RTT}} \right) \eta_i,
  \qquad \eta_i \sim \mathcal{U}[\eta_{\min}, \eta_{\max}] \ \text{i.i.d.}
\end{equation}
Note that $t_{\mathrm{RTT}}$ and $\eta_i$ only affect the player-side delay.

\paragraph{\textbf{Buffer dynamics:}}
Let $B_i$ be the buffer occupancy (in seconds) when the download of $V_i$ starts, with $B_1 = 0$. The buffer drains during the download, and the player stalls if the buffer runs out before the chunk arrives. The rebuffering (stall) time is
\begin{equation}
  R_i = \left( t_i^{\mathrm{delay}} - B_i \right)_+ ,
\end{equation}
where $(x)_+ = \max(0, x)$. Once $V_i$ is downloaded, it adds $L$ seconds of content:
\begin{equation}
  \tilde{B}_{i+1} = \left( B_i - t_i^{\mathrm{delay}} \right)_+ + L .
\end{equation}
To prevent overflow beyond the buffer capacity $B^{\max}$, the player waits before requesting the next chunk. The waiting time is quantized to multiples of a drain step $\delta$:
\begin{equation}
  \Delta t_i^{w} = \delta \left\lceil
    \frac{\left( \tilde{B}_{i+1} - B^{\max} \right)_+}{\delta}
  \right\rceil ,
  \qquad
  B_{i+1} = \tilde{B}_{i+1} - \Delta t_i^{w} .
\end{equation}
Hence $B_{i+1} \in [0, B^{\max}]$, and $B_{i+1} \in (B^{\max} - \delta, B^{\max}]$ whenever waiting is triggered. After downloading $V_i$ and waiting for $\Delta t_i^{w}$, the next download starts at
\begin{equation}
  \tau_{i+1} = \left( \tau_i + T_i^{\mathrm{tr}} + \Delta t_i^{w} \right) \bmod T ,
\end{equation}
where the modulo wraps the position back to the beginning of the trace once its duration $T$ is exhausted.

\paragraph{\textbf{QoE objective:}}
The reward for chunk $V_i$ balances video quality, rebuffering, and quality smoothness:
\begin{equation}
  r_i = \mu_1 b_i - \mu_3 R_i - \mu_2 \left| b_i - b_{i-1} \right| ,
\end{equation}
where bitrates are measured in Mbps and $R_i$ in seconds.

\paragraph{\textbf{Episode:}}
At the start of each episode, a trace is sampled uniformly from the dataset and the initial trace time $\tau_1$ is drawn uniformly from the trace breakpoints. The first chunk $V_1$ is downloaded at the lowest bitrate $b^{(1)}$ as a warm-up step without reward, and this bitrate serves as the reference $b_{1} = b^{(1)}$ for the smoothness term of $V_{2}$. The agent then selects bitrates for $V_2, \dots, V_I$, i.e. it makes $I-1 = 47$ decisions per episode; we index them by $t=1, \cdots, I-1$, where decision $t$ selects the bitrate of $V_{t+1}$. The episode terminates after $V_{I}$ is downloaded.

Table~\ref{tab:sim_params} summarizes the notation used in the system model. The upper part lists the simulation constants, which follow the widely used trace-driven setting of Pensieve~\cite{mao2017neural}; the trace dataset $\mathcal{D}$ and chunk sizes $d_i(b)$ are loaded from the corresponding datasets. The lower part lists the per-chunk variables updated at each decision step. Note that the implementation measures all time quantities in milliseconds, whereas the system model uses seconds; the conversion is applied when computing the reward, i.e., $R_i$ enters the QoE in seconds and bitrates in Mbps.

\begin{table}[!htbp]
\centering
\caption{Notation and simulation parameters.}
\label{tab:sim_params}
\begin{tabular}{lll}
\toprule
Symbol & Description & Value / Unit \\
\midrule
\multicolumn{3}{l}{\textit{Video}} \\
$I$ & Number of chunks per video & 48 \\
$L$ & Chunk duration & 4\,s \\
$M$ & Number of bitrate levels & 6 \\
$\mathbb{B}$ & Bitrate ladder & \{0.3, 0.75, 1.2, 1.85, 2.85, 4.3\}\,Mbps (300–4300 kbps) \\
$d_i(b)$ & Size of chunk $i$ at bitrate $b$ & bytes \\
\midrule
\multicolumn{3}{l}{\textit{Network}} \\
$\mathcal{D}$ & Bandwidth trace dataset & -- \\
$(t_k, c_k)$ & Timestamp and capacity of a trace & s, Mbps \\
$T$ & Total duration of a trace & $t_K$ \\
$\rho$ & Payload ratio & 0.95 \\
$S_k$ & Download speed in interval $k$ & $\rho \cdot \frac{10^6}{8} c_k$\,bytes/s \\
$t_{\mathrm{RTT}}$ & Round-trip time & 80\,ms \\
$[\eta_{\min}, \eta_{\max}]$ & Delay noise range & [0.9, 1.1] \\
\midrule
\multicolumn{3}{l}{\textit{Buffer and QoE}} \\
$B^{\max}$ & Buffer capacity & 60\,s \\
$\delta$ & Waiting step & 0.5\,s \\
$\mu_1$ & Bitrate weight & 1 \\
$\mu_2$ & Smoothness weight & 1 \\
$\mu_3$ & Rebuffering weight & 4.3 \\
\midrule
\multicolumn{3}{l}{\textit{Per-chunk variables}} \\
$b_i$ & Selected bitrate for chunk $i$ & Mbps \\
$\tau_i$ & Trace time at download start & s \\
$T_i^{\mathrm{tr}}$ & Transfer time & s \\
$C_i$ & Effective throughput & bytes/s \\
$t_i^{\mathrm{delay}}$ & Perceived download delay & s \\
$R_i$ & Rebuffering time & s \\
$\Delta t_i^{w}$ & Waiting time & s \\
$B_i$ & Buffer occupancy & s \\
\bottomrule
\end{tabular}
\end{table}

\newpage

\input{parallel_env}

\input{parallel_env_train}

\input{policy_converge}

\section{Training Versus Deployment}
\label{app:train-deploy}

Figure~\ref{fig:train-deploy} asks whether the comparison of ABR policies on test traces from the training distribution carries over to the conditions under which the policies are deployed. This section describes the policies, how each panel is produced, and what the figure shows in more detail.

\paragraph{\textbf{Policies:}}
The three learned agents, Pensieve, Merina and PaMoE, are trained as in Appendix~\ref{app:converge}: on 3G-Train-Pool with five training seeds each, 4096 parallel environments and 800 PPO updates per run. Each agent is represented by its best single checkpoint over all seeds and evaluations: Pensieve seed 2 after 700 updates, Merina seed 3 after 700 updates and PaMoE seed 0 after 500 updates. The two other policies are not learned: \emph{RobustMPC} plans over the next chunks with the chunk sizes of the video being played and a throughput estimate. \emph{FDASH} is a fuzzy-logic controller. Its inputs are the buffer level and its change since the previous chunk; nine Mamdani rules map them to a factor that multiplies a smoothed throughput estimate (an exponential moving average, weight 0.8 on the newest value, of the mean throughput of the last three chunks), and the highest bitrate not above the product is chosen. Two rules suppress switches: no switch up while the previous buffer level is at most 15\,s, and no switch down while the buffer is predicted to stay at or above 30\,s. We port the pyDash implementation to the batched environment; its fuzzy inference reproduces that of scikit-fuzzy.

\paragraph{\textbf{Training panel:}} Panel (a) shows, for each agent, the single training run that contains its best checkpoint, evaluated every 20 updates on 3G-Test-Pool with evaluation seed 123 (update 0, the random initial policy, is omitted). The star marks the highest point of each curve, which is the agent's best checkpoint. RobustMPC (1.453) and the offline optimum (1.918) are computed on the same episodes. Unlike Figure~\ref{fig:converge}, which averages over the five training seeds, the panel shows individual runs, so it includes the fluctuation between evaluations: from update 200 on, the curves stay within 1.602--1.675 (Pensieve), 1.555--1.663 (Merina) and 1.626--1.663 (PaMoE), and after 800 updates they end at 1.637, 1.649 and 1.647.

\paragraph{\textbf{Deployment panel:}} Panel (b) evaluates the five policies on nine trace families, all on their test splits: \emph{within the training distribution}: FCC-16, FCC-18, Oboe, Puffer-21 and Puffer-22, the five families of 3G-Test-Pool, with the Envivio video used in training; \emph{outside the training distribution}: Lumos4G, Lumos5G, SolisWi-Fi and Norway3G, traces from other datasets, with the Big Buck Bunny video (Appendix~\ref{app:converge}). Every policy is evaluated with five evaluation seeds (123 to 127), each giving $4096 \times 2$ episodes; under a given seed all policies see the same episodes. A cell shows the regret of a policy on a family: its QoE minus the highest QoE among the five policies on that family, both averaged over the five seeds. The regret is thus measured against the best available policy, not against the offline optimum. The checkpoints were selected on 3G-Test-Pool, which contains the test traces of the five families within the training distribution; the four families outside it played no part in the selection.

\paragraph{Within the training distribution:}
An agent is the best policy on every family: Merina on FCC-16 and FCC-18, Pensieve on Oboe, Puffer-21 and Puffer-22. No agent is more than 0.033 behind the best. The two non-learned controllers are behind on every family, RobustMPC by 0.07 to 0.46 and FDASH by 0.09 to 0.36.

\paragraph{Outside the training distribution:} A learned agent is the best policy on only one of the four families. On Lumos4G and SolisWi-Fi, RobustMPC performs best, leading Pensieve, the strongest of the three learned agents, by 0.48 and 0.39, respectively. On Lumos5G, FDASH performs best, scoring 0.22 higher than RobustMPC and 2.44 higher than Merina, the strongest learned agent there. Only on Norway3G is a learned agent the best policy: PaMoE leads Merina by 0.10.

\paragraph{\textbf{Summary:}}
Each of the five policies is the best on at least one family (Pensieve on three, Merina and RobustMPC on two, PaMoE and FDASH on one). The worst regret of each policy is 2.84 for Pensieve, 2.44 for Merina and 2.98 for PaMoE (all on Lumos5G), 0.48 for RobustMPC (Norway3G) and 15.595 for FDASH (SolisWi-Fi). No single policy wins everywhere. The three agents are nearly tied on the in-distribution test traces, and these results do not tell us which one will do best on other traces.

\newpage

\section{LLM-Generated Routing Policies}
\label{app:llm-routing}

This section describes how a Large Language Model writes a routing policy over a pool of frozen ABR experts: the setting, the generation procedure, the prompt, the three forms in which a policy can be written, how an answer is checked, and how a policy is evaluated. All JSON fragments, code and rationales attributed to the LLM in this section are illustrative examples of the format; the policies the LLM actually produces may differ.

\subsection{Setting}
\label{app:llm-setting}

\paragraph{\textbf{Expert pool:}} The pool $\mathcal{M}$ contains $|\mathcal{M}| = 5$ experts whose parameters are frozen: the three learned agents Pensieve, Merina and PaMoE (best single checkpoints, Appendix~\ref{app:converge}) and the two controllers without learned parameters, RobustMPC and FDASH (Appendix~\ref{app:train-deploy}). The bitrate ladder $\mathbb{B}$ has six levels, 300, 750, 1200, 1850, 2850 and 4300\,kbps (Appendix~\ref{sec:system-model}), indexed $0$ to $5$; in this and the following sections bitrates and throughputs are given in kbps, as in the prompt.


\paragraph{\textbf{Router:}} At each decision, every expert receives the observation and proposes a bitrate level, its \emph{proposal}. The router selects one expert and plays that expert's proposal; it never chooses a bitrate itself. The level actually played is passed back to all experts, so that each expert's record of the previous bitrate is the one that was played. Because all experts run at every chunk, an expert that is not selected keeps its internal state up to date and can take over at any chunk.

\paragraph{\textbf{Observation}:} Following the observation space of Pensieve and PaMoE \cite{mao2017neural, 11498444}, after each chunk the router observes seven quantities: the download time of that chunk (ms), the time the player slept because the buffer was full (ms), the buffer level (ms), the stall time caused by the chunk (ms), the size of the chunk (bytes), the number of chunks left, and the sizes of the next chunk at the six levels (bytes).

\paragraph{\textbf{Goal:}} The LLM writes a routing policy $\pi$ offline. At deployment $\pi$ is a fixed artifact executed by a program; the LLM is not called. While writing $\pi$, the LLM sees only measurements on the training splits of the trace families; no test split is used. At run time the router does not know which trace family it is playing.

\subsection{Routing Policies and Policy Forms}
\label{app:llm-policy}

A routing policy $\pi$ maps the information available at each decision step to an expert. Specifically, at decision $t \in \{1, \dots, I-1\}$ of an episode ($I - 1 = 47$, Appendix~\ref{sec:system-model}), the router holds the history $h_t$, which consists of the observations up to step $t$, the bitrates played so far, the experts' proposals for the current chunk, and the router's own previous selections, and it selects $\pi(h_t) \in \mathcal{M}$. Since $\pi$ is executed at deployment without invoking the LLM, the LLM cannot express $\pi$ in free-form text; instead, it must write $\pi$ in a language that the deployment program can parse and execute. We provide three such languages, called the \emph{policy forms} $F \in \{\mathrm{I}, \mathrm{II}, \mathrm{III}\}$.
A form specifies two components: its \emph{inputs}, i.e., which parts of $h_t$ the policy may access; and its \emph{decision procedure}, which maps these inputs to an expert and is authored either by us, as a fixed rule-scoring program, or by the LLM, as generated code.
\textbf{Form~I}: the LLM's features may read only the observations and the bitrates played, and are combined by the fixed rule-scoring program; 
\textbf{Form~II}: the features may in addition read the experts' proposals and the router's switching history, with the same program; it differs from Form~I only in its inputs; \textbf{Form~III} uses the same inputs as Form~II, with code written by the LLM; it differs from Form~II only in its decision procedure. A difference in QoE between Forms~I and~II can therefore be attributed to the additional inputs, and a difference between Forms~II and~III to the decision procedure, i.e.\ to whether the fixed rule structure limits the policies the LLM can express. Since every policy is written by the LLM from the format description $D_F$, both differences also include how well the LLM uses the given form and how the form is described to it. The forms are summarised in the rest of this subsection and detailed in Appendix~\ref{app:llm-forms}.

\emph{\textbf{Form~I: fuzzy rule base over features of the observation.}}
A fuzzy rule base is a set of if--then rules over named ranges of numeric quantities, such as ``if the recent throughput is volatile and the buffer is low, select RobustMPC''. Unlike a hard threshold, a condition such as ``the buffer is low'' can hold partially, to a degree between 0 and 1. A Form~I rule base consists of three components: features, intervals, and rules.
\emph{\textbf{Features}} are the quantities on which the rules are conditioned. Each feature is constructed by composing operators from a fixed set and is recomputed at every decision from the observations of the current episode; for example,
$\texttt{tp\_cv}$ is the coefficient of variation of the throughputs measured over
the five most recent chunks.

\emph{\textbf{Intervals}} are named fuzzy ranges defined over a feature, such as
$\texttt{Low}$ and $\texttt{High}$. Each interval is specified by four breakpoints $a \le b \le c \le d$ that define a trapezoidal membership function: a feature value has membership degree~1 on $[b, c]$, a degree that varies linearly on $(a, b)$ and $(c, d)$, and degree~0 beyond $a$ and $d$; an outermost interval can instead extend to the end of the axis.

\emph{\textbf{Rules}} map conditions to experts. A rule base contains at most 24 rules. Each rule pairs one or more conditions with a single expert, where each condition states that a feature falls in one of its intervals, for example, "if throughput variability is High and the buffer is Low, then select RobustMPC".

\emph{\textbf{Form~II: fuzzy rule base with router-side features.}}
Form~II is identical to Form~I in its components, its decision procedure, its parameters and the program that executes it. The only difference is the set of operators from which features are composed: Form~II adds four operators that read information available to the router but not contained in the observation.
\emph{\textbf{Added operators.}} (i) The bitrate, or level, that a given expert proposes for the current chunk; (ii) the spread of the levels proposed by the five experts, i.e.\ the highest minus the lowest, from~0 when all experts agree to~5; (iii) the number of chunks since the router last switched experts; and (iv) the stall time accumulated since that switch. The proposals are available at decision time because every expert computes its proposal before the router selects one of them.
\emph{\textbf{What they add.}} A Form~I rule can refer only to the network and the buffer; a Form~II rule can also refer to what the experts intend to do and to how the current expert has performed since it was selected, for example, ``if the proposal spread is $\texttt{Large}$ and $\texttt{buffer}$ is $\texttt{Low}$, then select RobustMPC''.

\emph{\textbf{Form~III: Python function.}}
In Form~III the LLM writes the decision logic itself, as a Python module with two functions of fixed interface: $\texttt{route(obs, proposals, ctx, state)}$, which the router calls at every decision and which returns the name of an expert, and $\texttt{init\_state()}$, which creates, at the start of every episode, the dictionary $\texttt{state}$ in which $\texttt{route}$ keeps information across the decisions of that episode. Within this interface nothing is prescribed: no features, intervals or rules. The code must be written in a restricted subset of Python; the interface, the restrictions and an example are given in Appendix~\ref{app:llm-form3}.

\subsection{Generation Procedure}
\label{app:llm-procedure}

\paragraph{\textbf{Overall procedure.}}
Algorithm~\ref{alg:llm-gen} generates a policy of a given form $F$ in three steps.
(1)~\emph{Measure}: each expert plays the training split of each trace family alone, and the results are summarised in three tables.
(2)~\emph{Build the prompt}: the tables are combined with a description of form $F$ into a prompt.
(3)~\emph{Query and check}: the LLM answers the prompt with a policy; a program checks whether the answer is a valid policy of form $F$ and, if it is not, queries the LLM again with the reason, up to three attempts in total. All three forms share the same procedure and differ only in two form-specific elements: the description of the form included in the prompt, $D_F$, and the check applied to each candidate answer, $\textsc{Check}_F$. The procedure returns the first candidate that passes $\textsc{Check}_F$, or reports a failure if none does. The three steps are detailed in turn.

\begin{algorithm}[!htbp]
\caption{One-shot generation of a routing policy of form $F$}
\label{alg:llm-gen}
\small
\SetKwComment{Comment}{$\triangleright$\ }{}
\SetCommentSty{textit}
\DontPrintSemicolon
\KwIn{Training splits of the trace families $\mathcal{F} = \{$FCC-16, FCC-18, Oboe, Puffer-21, Puffer-22$\}$; expert pool $\mathcal{M}$; form $F$ with format description $D_F$ and check $\textsc{Check}_F$.}
\KwOut{A routing policy $\pi$ of form $R$, or failure.}
\Comment{(1) Measure}
\ForEach{$f \in \mathcal{F}$}{
  \ForEach{$m \in \mathcal{M}$}{
    play the 8192 episodes of $f$ with $m$ alone; record the row $(f, m)$ of Table~A and the values of the four quantities of Table~B2\;
  }
  compute the row $f$ of Table~B from the trace files\;
}
\Comment{(2) Prompt}
$\mathit{sys} \gets$ role statement $\Vert\; D_F$\;
$\mathit{usr}_0 \gets$ expert descriptions $\Vert$ Tables A, B, B2 as text $\Vert$ task sentence\;
$\mathit{usr} \gets \mathit{usr}_0$\;
\Comment{(3) Query and check}
\For{$k \gets 1$ \KwTo $3$}{
  $y \gets \textsc{LLM}(\mathit{sys}, \mathit{usr})$ \Comment*[r]{temperature}
  $(\pi, \mathit{msg}) \gets \textsc{Check}_F(y)$ \Comment*[r]{Algorithms~\ref{alg:llm-check-fuzzy}, \ref{alg:llm-check-python}}
  \lIf{$y$ passed}{\Return $\pi$}
  $\mathit{usr} \gets \mathit{usr}_0 \;\Vert$ ``Your previous answer was rejected'' $\Vert\; y \;\Vert$ ``Reason:'' $\mathit{msg} \;\Vert$ ``Answer again with a corrected policy, in the same format, nothing else.''\;
}
\Return failure\;
\end{algorithm}

\paragraph{\textbf{Step 1: Measure (Algorithm~\ref{alg:llm-gen}, block (1)).}}
\label{app:llm-meas}
The LLM receives no routing heuristic; it has to derive the policy from measurements that we place in the prompt. A routing policy requires two kinds of knowledge: (i)~which expert performs best under which network conditions, and (ii)~how the router can recognise these conditions at run time, when it sees only its own observations and not the name of the trace family. Step~1 measures both on the training splits of the five trace families. For~(i), each expert plays every family on its own. For~(ii), each family is described in two ways. The first is the link capacity in its trace files, e.g.\ the median bandwidth; it characterises the network, but the router never observes it. The second is what the router does observe: quantities that it can compute from its observations before each decision, such as the mean throughput of the last five chunks.

\emph{Table~A: which expert is best on which family.} For each family $f$ and expert $m$, expert $m$ plays the episodes of $f$ on its own, i.e.\ it chooses the bitrate of every chunk without a router. The playback follows the protocol of the evaluation (Appendix~\ref{app:llm-eval}): 4096 parallel environments each play two consecutive episodes, 8192 episodes in total, with seed 123. For each of the $5 \times 5$ pairs $(f, m)$, Table~A reports three averages over all chunks of these episodes: the QoE $r_i$, the rebuffering time $R_i$ (s), and the bitrate change $\left| b_i - b_{i-1} \right|$ (Mbps). The QoE ranks the experts on each family. The other two columns are the quantities penalised and show where an expert loses QoE. Below the table, one line per family names the best expert and the QoE difference between the best and the worst expert. For example, on Puffer-21 (Table~\ref{tab:llm-profile}) Pensieve has the highest QoE ($1.151$) and RobustMPC the lowest ($0.749$). The difference of $0.402$ comes mainly from rebuffering: RobustMPC rebuffers $0.142$\,s per chunk against $0.065$\,s for Pensieve, which costs $4.3 \times 0.077 \approx 0.33$. Table~A thus tells the LLM not only to avoid RobustMPC on networks like those of Puffer-21, but also that the risk on these networks is rebuffering.

\begin{table}[!htbp]
  \centering
  \caption{Row of Table A for Puffer-21 (training split), as shown to the LLM: each expert playing alone.}
  \label{tab:llm-profile}
  \small
  \begin{tabular}{lccc}
    \toprule
    Expert & QoE $r_i$ & $R_i$ (s) & $\left| b_i - b_{i-1} \right|$ (Mbps) \\
    \midrule
    Pensieve  & $+1.151$ & 0.065 & 0.180 \\
    Merina    & $+1.128$ & 0.068 & 0.185 \\
    PaMoE     & $+1.131$ & 0.068 & 0.188 \\
    RobustMPC & $+0.749$ & 0.142 & 0.312 \\
    FDASH     & $+0.951$ & 0.066 & 0.263 \\
    \bottomrule
  \end{tabular}
\end{table}

\emph{Table~B: link capacity.} Table~B is computed, without playback, from the trace files of the training split of each family. A trace is a sequence of samples $(t_k, c_k)$, $k = 1, \dots, K$, where $c_k$ is the link capacity during $(t_{k-1}, t_k]$ (Appendix~\ref{sec:system-model}); Table~B uses only the capacities $c_k$, converted to kbps. For each family it reports three groups of numbers:
\begin{itemize}
  \item \emph{Size}: the number of traces in the training split, e.g.\ 500 for Puffer-21.
  \item \emph{Level}: the capacity samples $c_k$ of all traces of the split are pooled, each sample counting once regardless of the length of its interval, and the table gives their mean and their 10th, 25th, 50th, 75th and 90th percentiles.
  \item \emph{Variability}: for each sample $k \ge 2$ of each trace, the coefficient of variation of the last five samples,
  \begin{equation*}
    \mathrm{cv}_k = \frac{\sigma(c_{j}, \dots, c_k)}{\bar{c}(c_{j}, \dots, c_k)}, \qquad j = \max(1, k-4),
  \end{equation*}
  where $\bar{c}$ is the mean and $\sigma$ the (population) standard deviation; at the start of a trace the window contains fewer than five samples, and windows with zero mean are skipped. A trace with $K$ samples thus yields $K-1$ values $\mathrm{cv}_k$. We put the values from all traces in the split together, so longer traces count more, and report the median and 90th percentile.
\end{itemize}
Table B summarizes how the network conditions of the families differ. The families differ widely: for example, Puffer-21 has a median bandwidth of 1188 kbps and a median cv of 0.643, while FCC-18 has 4272 kbps and 0.046. Two caveats apply. First, the sampling interval differs between families (median 5 s on FCC-18, 2 s on Puffer-21 and Puffer-22, 1.9 s on FCC-16, and 0.37 s on Oboe), so a five-sample window covers different durations and the variability columns are only roughly comparable; the low cv of FCC-18, for instance, partly reflects its long interval. Second, the table describes link capacity, whereas the router sees only the throughput of downloaded chunks, which also depends on chunk sizes and download timing and can differ from capacity by a large factor. The table therefore cannot provide thresholds for the router. Instead, the prompt points the LLM to Table B2, which reports the same statistics on the observed chunk throughput.

\emph{Table~B2: quantities the router can compute.} Table~B2 describes the trace families in the units the router observes. Four quantities are computed before each decision from the current and past observations of the episode:
\begin{itemize}
  \item $\texttt{tp\_mean}$: the mean throughput of the last five chunks, where the throughput of a chunk is its size in bytes times 8 divided by its download time in ms (kbps);
  \item $\texttt{tp\_cv}$: the standard deviation of the same five values divided by their mean;
  \item $\texttt{buffer}$: the buffer level (s);
  \item $\texttt{buffer\_diff}$: the buffer level minus that at the previous decision (s).
\end{itemize}

At the start of an episode, fewer than five throughputs are available; the available ones are used, $\texttt{tp\_cv}$ is set to 0 while there is only one, and $\texttt{buffer\_diff}$ is 0 at the first decision. The values are recorded during the playback for Table~A, without additional runs: whenever an expert is about to choose a bitrate, the four quantities are computed and stored. The values depend on which expert is playing, because its bitrate choices change the chunk sizes and the buffer. Table~B2 should show the values that the router will encounter at run time. These values depend on which experts the policy selects, but the policy does not yet exist when the table is computed. We therefore pool the values of all five experts, so that the table covers the values that each of them produces. The pooled values are only an approximation. Under an actual policy, some experts may be selected more often than others, and the policy may switch experts in the middle of an episode. Neither happens when each expert plays alone, so the values the router actually sees can differ from the table.

\begin{table}[!htbp]
	\centering
	\caption{Two rows of Table B2 as shown to the LLM: distribution of $\texttt{tp\_cv}$ over all decisions of the five experts playing alone (training splits).}
	\label{tab:llm-observed}
	\small
	\setlength{\tabcolsep}{4pt}
	\begin{tabular}{lccccccc}
		\toprule
		Family & min & p5 & p25 & p50 & p75 & p95 & max \\
		\midrule
		FCC-18    & 0.000 & 0.024 & 0.047 & 0.064 & 0.097 & 0.249 & 1.710 \\
		Puffer-21 & 0.000 & 0.054 & 0.207 & 0.345 & 0.488 & 0.733 & 1.752 \\
		\bottomrule
	\end{tabular}
\end{table}

Each trace family thus contributes $5$ experts $\times$ $8192$ episodes $\times$ $47$ decisions $= 1{,}925{,}120$ values per quantity, and Table~B2 gives their minimum, 5th, 25th, 50th, 75th and 95th percentiles and maximum (Table~\ref{tab:llm-observed}). With these percentiles, the LLM can judge how often a threshold is exceeded on each family. For example, $\texttt{tp\_cv} = 0.3$ lies above the 95th percentile on FCC-18 ($0.249$) but below the median on Puffer-21 ($0.345$); a rule conditioned on $\texttt{tp\_cv} > 0.3$ therefore applies to fewer than 5\% of the decisions on FCC-18 and to more than half of those on Puffer-21, and thus separates the two families.

\paragraph{\textbf{Step 2: Build the prompt (block (2))}.}
\label{app:llm-prompt}
The prompt consists of a system prompt, which fixes the task and the answer format, and a user prompt, which carries the measurements.

\emph{\textbf{System prompt.}} The system prompt has two parts: a role statement, which is identical for all forms, followed by the format description $D_F$ of form $F$.

\emph{(i) Role statement.} The role statement consists of three fixed paragraphs:
\begin{lstlisting}[style=policy]
You are designing, from scratch, the controller of a router that picks, once per video chunk, which of several frozen pre-trained adaptive-bitrate (ABR) experts acts. The router only delegates: the quality played is the chosen expert's own decision for this chunk; the router never chooses a bitrate itself. Every expert sees every chunk, so an expert that is not chosen still keeps its state up to date, and it is told the quality actually played.

You are given ONLY measurements: how each expert scores on each training trace family, the bandwidth statistics of those families, and the observed ranges of quantities the router can compute. You are NOT given any routing heuristic. Infer what to observe and how to route from the measurements.

At run time the router does not know which trace family it is on; it only sees what it observes. A regime is only actionable if the observations can tell it apart from the others.
\end{lstlisting}
The first paragraph defines what the policy controls: it selects an expert, never a bitrate. The second states that the LLM must derive the policy from the measurements alone. The third states a constraint. The router does not know the trace family, so a difference between families is useful only if it shows up in the observations. Table~B2 lets the LLM check whether it does.

\emph{(ii) Format description $D_F$.} The rest of the system prompt is $D_F$, a text that we wrote for each form. $D_F$ is the LLM's only description of the language in which the policy must be written. It states three things: (1) the syntax of an answer; (2) how the program executes an answer, so that the LLM can predict how its policy will behave, i.e.\ which expert
will be selected in a given situation. For Forms~I and~II, this is how the membership degrees of the conditions are combined into a score for each expert, and how the selected expert follows from these scores. For Form~III, it is when $\texttt{init\_state}$ and $\texttt{route}$ are called; (3) all the limits that $\textsc{Check}_F$ enforces, such as a maximum of 24 rules, so that the LLM knows them before it writes its answer. Knowing them in advance, the LLM can write an answer that passes the
check without having to revise it. For example, $D_F$ for Form~III reads as follows (shortened; [\dots] marks omitted lines):

\begin{lstlisting}[style=policy]
== How the deployed router runs your controller ==
You write two plain Python functions. For every video session (episode) the router calls
init_state() once to make a fresh `state`; then, once per chunk, it calls
    route(obs, proposals, ctx, state) -> "<expert name>"
and plays the quality that expert proposed for this chunk.
  obs        dict, the observation after the chunk just downloaded: [...]
  proposals  dict expert name -> level (int 0..5) that expert proposes for THIS chunk. [...]
  state      your own dict from init_state(), kept between calls of the same session; [...]
== Language ==
Plain Python, restricted:
  - no import (the `math` module is available as `math`), no class, no global / nonlocal, no with,
    no name or attribute starting with "_"; [...]
  - at most 200 lines. [...]
== Output ==
A SINGLE ```python fenced block containing the whole module, nothing else outside it.
Return one of: pensieve, merina, pamoe, robustmpc, fdash.
\end{lstlisting}
The format descriptions of Forms~I and~II are organised in the same way; their excerpts are given with each form in Appendix~\ref{app:llm-forms}.

\emph{\textbf{User prompt.}} The user prompt consists of three parts: one sentence per expert describing how it works, e.g.\ for RobustMPC ``model-predictive control: harmonic-mean throughput prediction discounted by the worst recent prediction error, then a search over the bitrates of the next 3 chunks (no learned parameters)''; Tables~A, B and~B2 as text; and the task sentence \emph{``Write the routing controller in the format described in the system prompt. Route each measured regime to the expert the measurements say is strongest there.''}

\paragraph{\textbf{Step 3: Query and check (block (3)).}}
The LLM is queried with the two prompts at temperature 0, and its answer $y$ is passed to $\textsc{Check}_F$ (Appendix~\ref{app:llm-check}). The check decides whether $y$ is a valid policy of form $F$: it either returns the parsed policy $\pi$, or rejects $y$ with a message $\mathit{msg}$ naming the violated constraint (e.g.\ \texttt{rule[7] uses undeclared label buffer.Medium}). If $y$ is rejected, the LLM is queried again with the original user prompt extended by $y$ and $\mathit{msg}$ (Algorithm~\ref{alg:llm-gen}, last line of the loop). At most three attempts are made; if none passes, generation fails for this form.

\paragraph{One-shot generation.}
The LLM never receives the QoE of a policy it has written; the only feedback is whether its answer is a valid policy. The returned $\pi$ is fixed and evaluated afterwards with the protocol of Appendix~\ref{app:llm-eval}.

\subsection{Policy Forms in Detail}
\label{app:llm-forms}

\subsubsection{Form I: fuzzy rule base}
\label{app:llm-form1}

\paragraph{Overview.}

A routing policy answers two questions: what to observe, and which expert to select given the observation. The raw observation consists of seven low-level quantities (in bytes and milliseconds) that describe only the chunk just downloaded. A good decision, however, depends on quantities derived over several chunks, such as how high and how variable the recent throughput is, or whether the buffer is growing. Form~I therefore selects an expert in three steps, each defined by one part of the LLM's answer, followed by a fixed $\arg\max$:
\begin{equation*}
	\underbrace{\text{observation}}_{7\ \text{quantities}}
	\xrightarrow{\ \text{(1) features}\ } x \in \mathbb{R}^{J}
	\xrightarrow{\ \text{(2) intervals}\ } \phi_{j,l}(x_j) \in [0,1]
	\xrightarrow{\ \text{(3) rules}\ } w_m \ \ (m \in \mathcal{M})
	\xrightarrow{\ \arg\max\ } \text{expert}.
\end{equation*}
\emph{Features} specify which quantities are computed from the observation. \emph{Intervals} define named states of each feature, such as ``low'' or ``high'', and the degree to which the feature's value is in each state. \emph{Rules} combine these degrees into a score $s_m$ for each expert $m$, and the expert with the highest score is selected, subject to the adjustments of part~(4).

\paragraph{(1) Features.}
Network conditions can be told apart only by statistics over several chunks. Which statistics are useful, and over how many chunks, should follow from the measurements (Tables~A and~B2), so we let the LLM declare the features itself instead of giving it a fixed set. It builds them from a fixed list of operators rather than writing arbitrary code, for three reasons. First, a declaration is data, so it can be validated before it runs. Second, every operator has a batched GPU implementation, so the features of all episodes are computed at once. Third, the quantities in Table~B2 are defined with the same operators, so the LLM's features are computed exactly like the values it is shown, and it can read thresholds directly from Table~B2.

\emph{Prompt.} The format description specifies the features as follows (condensed; the full prompt is given in Appendix~\ref{app:prompt}):
\begin{lstlisting}[style=policy]
1. Evaluate the `features` section in declaration order to get one number per feature.
...
A feature is one node. `of` / `num` / `den` may only reference a feature declared EARLIER in the object (no forward references, no cycles). Operators:
  {"op":"obs", "field":<name>, "scale":<float>, "index":<int>}   a raw observation field, multiplied by `scale`
  {"op":"throughput"}             goodput of the chunk just downloaded, kbps (= chunk_bytes*8/delay_ms)
  {"op":"last_bitrate_kbps"}      the bitrate actually played on the previous chunk, kbps
  {"op":"diff", "of":<feature>}                          value now minus value last chunk
  {"op":"window_mean",  "of":<feature>, "window":<int>}  over the last `window` chunks (1..64)
  {"op":"window_std" | "window_cv" | "window_slope", ...} std, std/mean, least-squares slope per chunk
  {"op":"ema",   "of":<feature>, "alpha":<(0,1]>}        alpha on the new sample
  {"op":"ratio", "num":<feature>, "den":<feature>}
Every video session (episode) starts afresh: windows are empty, diff is 0 on its first chunk.
You may declare intermediate features without membership functions (e.g. a raw throughput series feeding a windowed statistic). Only features in `membership` can appear in rules.
In `notes`, say why each feature earns its place. A feature that no measurement justifies enlarges the rule space a human must read without buying discrimination.
\end{lstlisting}

\emph{Answer (example).} The features are given as an ordered map from names to operators:
\begin{lstlisting}[style=policy]
"features": {
  "tp":     {"op": "throughput"},
  "tp_cv":  {"op": "window_cv", "of": "tp", "window": 5},
  "buffer": {"op": "obs", "field": "buffer_size_ms", "scale": 0.001}
}
\end{lstlisting}
Here $\texttt{tp}$ is an intermediate feature: it has no membership functions, so it cannot appear in rules and serves only as the input of $\texttt{tp\_cv}$. The \texttt{notes} could justify $\texttt{tp\_cv}$ as follows: Table~A shows that the best expert differs across trace families, and Table~B2 shows that $\texttt{tp\_cv}$ is distributed very differently across these families; $\texttt{tp\_cv}$ can therefore help the policy recognise at run time which kind of network it is on, and hence which expert to select.

At each decision, the program computes all features for all episodes at once, in declaration order. Operators that depend on past chunks (window statistics, $\texttt{diff}$ and $\texttt{ema}$) keep a separate history for each episode, which is cleared when the episode ends. At the start of an episode, window statistics use only the chunks seen so far and $\texttt{diff}$ is 0; the warm-up of part~(4) covers this period.

\paragraph{(2) Intervals.}
Rules such as ``if the buffer is low, use RobustMPC'' require named states of continuous features. A hard threshold (``low'' if $\texttt{buffer} < 7.5$\,s) makes membership jump between 0 and 1. A feature that fluctuates around the threshold can then make the selected expert alternate from chunk to chunk, and a small shift of the threshold can change many decisions at once. We therefore give each state transition bands in which membership changes gradually, so that scores vary continuously with both the feature value and the thresholds.

\emph{Definition.} A named state $l$ of feature $j$ is given by four numbers $a \le b \le c \le d$, which define the trapezoidal membership function
\begin{equation}
	\phi_{j,l}(x) =
	\begin{cases}
		1 & b \le x \le c,\\
		1 & x < b \text{ and } a = b \quad\text{(left shoulder)},\\
		1 & x > c \text{ and } c = d \quad\text{(right shoulder)},\\
		(x-a)/(b-a) & a < x < b,\\
		(d-x)/(d-c) & c < x < d,\\
		0 & \text{otherwise}.
	\end{cases}
	\label{eq:llm-trapezoid}
\end{equation}
Membership is 1 on $[b, c]$, rises linearly on $(a, b)$, falls linearly on $(c, d)$, and is 0 elsewhere, except that an equality $a = b$ makes the state hold with degree~1 for every $x \le b$, and $c = d$ for every $x \ge c$. The lowest and highest states use these shoulders so that the two ends of the axis are covered. A hard threshold is therefore not a special case (a state $[a,a,c,c]$ has degree~1 everywhere); a near-hard boundary is written with a narrow band, e.g.\ $[0, 0, 7.5, 7.6]$ for a boundary at $7.5$.

\emph{Prompt.}
\begin{lstlisting}[style=policy]
2. For every feature that appears in `membership`, each label's trapezoid gives a degree in [0,1]:
     trapezoid(x; a,b,c,d) = 0 for x<=a or x>=d; 1 for b<=x<=c; linear on (a,b) and (c,d).
     Shoulders: a == b means degree 1 for every x <= b; c == d means degree 1 for every x >= c.
     Use shoulders on the outermost labels so that the whole axis is covered.
...
  - Breakpoints satisfy a <= b <= c <= d, in that feature's own units.
\end{lstlisting}

\emph{Answer (example).}
\begin{lstlisting}[style=policy]
"membership": {"tp_cv":  {"Stable": [0, 0, 0.1, 0.3],  "Volatile": [0.1, 0.3, 1e9, 1e9]},
                "buffer": {"Low":    [0, 0, 5, 10],     "High":     [5, 10, 1e9, 1e9]}}
\end{lstlisting}
For example, $\texttt{Stable} = [0, 0, 0.1, 0.3]$ has $a = b$, so it has no rising side: its degree is 1 for $\texttt{tp\_cv} \le 0.1$ and falls to 0 at $0.3$, e.g.\ $0.5$ at $0.2$. $\texttt{Volatile} = [0.1, 0.3, 10^9, 10^9]$ is its mirror image: its degree is 0 up to $0.1$, rises to 1 at $0.3$ and, since $c = d$, stays 1 for all larger values; $10^9$ only stands for ``no upper end''. Between $0.1$ and $0.3$ a value thus belongs partly to both labels. The breakpoints can be read from Table~B2: the median of $\texttt{tp\_cv}$ is $0.064$ on FCC-18, below $0.1$, and $0.345$ on Puffer-21, above $0.3$. More than half of the decisions on FCC-18 are therefore fully $\texttt{Stable}$, and more than half of those on Puffer-21 fully $\texttt{Volatile}$.

At each decision the program computes $\mu_{j,l}$ for every state used by a rule. Adjacent states may overlap: at $\texttt{buffer} = 7.5$\,s both $\texttt{Low}$ and $\texttt{High}$ have membership 0.5. If two adjacent states leave a gap between them, any value in that gap has membership 0 in both states. Such a value can leave every rule with a score of 0; in that case, the policy falls back to the default expert.

\paragraph{(3) Rules.}
\emph{Definition.} A rule $q$ has the form $\textsc{if}\ x_{j_1} \text{ is } l_1 \ \textsc{and}\ x_{j_2} \text{ is } l_2 \ \textsc{and}\ \dots \ \textsc{then}\ m_q$. Its conditions form the \emph{antecedent set} $A_q = \{(j_1, l_1), (j_2, l_2), \dots\}$, in which each feature appears at most once, and the expert it selects is its \emph{consequent} $m_q \in \mathcal{M}$. The rule ignores all features not in $A_q$. Its score is the product of the memberships of its conditions, $w_q = \prod_{(j,l) \in A_q} \phi_{j,l}(x_j)$, which equals 1 if every condition is fully satisfied, 0 if any condition is not satisfied at all, and lies strictly between 0 and 1 otherwise. At a given decision, several rules usually have positive scores. Rules that select the same expert are alternatives, so the expert's score is the highest score among them:
\begin{equation}
	w_m = \max_{q:\, m_q = m} w_q, \qquad w_m = 0 \text{ if no rule selects } m.
	\label{eq:llm-score}
\end{equation}
Rules that select different experts compete: the policy selects the expert with the highest score, $\arg\max_{m \in \mathcal{M}} w_m$, and breaks exact ties by the fixed pool order (Pensieve, Merina, PaMoE, RobustMPC, FDASH). If every expert scores 0, the default expert is used. Since conflicts are resolved by score at every decision, the LLM does not need to make its rules mutually exclusive.

\emph{Duplicate antecedents.} Two rules with the same antecedent set always have equal scores. If they select the same expert, one of them is redundant; if they select different experts, the tie between them is always broken by pool order, which carries no information about the current state. The check therefore rejects any rule set that contains two rules with the same antecedent set.

\emph{Prompt.}
\begin{lstlisting}[style=policy]
3. Rule firing strength = PRODUCT of its antecedent degrees.
4. Per expert, aggregate across rules by MAX.
...
  - Every label used in a rule is declared in `membership` for that feature.
  - Every consequent is one of: pensieve, merina, pamoe, robustmpc, fdash.
  - Every rule has at least one antecedent; it ignores the features it omits.
  - No two rules have the same antecedents (their scores would always be equal).
  - At most 24 rules. Prefer the smallest controller that covers the measured regimes; every rule is one a human has to audit.
\end{lstlisting}

\emph{Answer (example).}
\begin{lstlisting}[style=policy]
"rules": [
  {"antecedents": {"tp_cv": "Stable"},                     "consequent": "pensieve"},
  {"antecedents": {"tp_cv": "Volatile", "buffer": "Low"},  "consequent": "robustmpc"},
  {"antecedents": {"tp_cv": "Volatile", "buffer": "High"}, "consequent": "pamoe"}
]
\end{lstlisting}

\paragraph{(4) Additional parameters.}
Three mechanisms adjust the expert chosen by the rules.
\begin{itemize}
	\item \emph{Fence.} A list of constraints of the form ``expert $m$ may be selected only while $x_j \ge a$'' (or $x_j \le a$), e.g.\ FDASH only while $\texttt{buffer} \ge 4$\,s. When the current features violate an expert's constraint, the expert is \emph{fenced out}: it cannot be selected, whatever its score. The fence lets the LLM exclude an expert from states in which it considers that expert harmful.
	\item \emph{Stickiness.} Two parameters, $d_{\min}$ and $\theta$. The policy switches from the current expert to the highest-scoring one only if the current expert has been used for at least $d_{\min}$ consecutive chunks and the new expert's score exceeds the current one's by more than $\theta$. The only exception is a current expert that is fenced out, which is replaced immediately. Stickiness limits switching because a switch often changes the bitrate, which QoE penalises through the term $\mu_2 \left| b_i - b_{i-1} \right|$.
	\item \emph{Warm-up.} A number of decisions $t_{\mathrm{w}}$ and a default expert $m_{\mathrm{def}}$. The default expert is used during the first $t_{\mathrm{w}}$ decisions of an episode, while the feature windows are not yet full, and whenever no expert has a positive score. If $m_{\mathrm{def}}$ is itself fenced out, the highest-scoring expert that is not fenced out is used instead.
\end{itemize}

\paragraph{(5) Decision procedure.}
Algorithm~\ref{alg:llm-fuzzy} gives one decision of an episode. It first scores every expert with the rules (parts (1)--(3)) and then applies the fence, stickiness and warm-up (part (4)). In machine-learning terms, each expert's score is a maximum over products of soft indicator functions, so the rules form a fuzzy rule-based classifier with max--product inference.

\begin{algorithm}[!htbp]
	\caption{One decision of a Form~I policy (one episode, decision $t = 1, \dots, I-1$)}
	\label{alg:llm-fuzzy}
	\small
	\SetKwComment{Comment}{$\triangleright$\ }{}
	\SetCommentSty{textit}
	\DontPrintSemicolon
	\KwData{Current expert $m_{\mathrm{cur}}$ (initially $m_{\mathrm{def}}$); number $d$ of consecutive decisions using it (initially 0).}
	compute all features $x$ in declaration order \Comment*[r]{part (1)}
	\ForEach{$m \in \mathcal{M}$}{
		$w_m \gets$ Eq.~\eqref{eq:llm-score} with $\phi$ from Eq.~\eqref{eq:llm-trapezoid} \Comment*[r]{parts (2), (3)}
		\lIf{$m$ is fenced out}{$w_m \gets -1$ \Comment*[r]{fence}}
	}
	$m^\star \gets \arg\max_{m} w_m$ \Comment*[r]{ties: first in pool order}
	\If{$w_{m_{\mathrm{cur}}} = -1$ \KwSty{or} $(d \ge d_{\min}$ \KwSty{and} $w_{m^\star} > w_{m_{\mathrm{cur}}} + \theta)$}{
		$m_{\mathrm{new}} \gets m^\star$ \Comment*[r]{stickiness}
	}
	\lElse{$m_{\mathrm{new}} \gets m_{\mathrm{cur}}$}
	\If{$t \le t_{\mathrm{w}}$ \KwSty{or} $\max_m w_m \le 0$}{
		$m_{\mathrm{new}} \gets m_{\mathrm{def}}$ if it is not fenced out, else the highest-scoring expert that is not fenced out \Comment*[r]{warm-up}
	}
	$d \gets d + 1$ if $m_{\mathrm{new}} = m_{\mathrm{cur}}$, else $1$;\quad $m_{\mathrm{cur}} \gets m_{\mathrm{new}}$\;
	play the proposal of $m_{\mathrm{cur}}$ for this chunk\;
\end{algorithm}

\paragraph{Worked example.}
We apply the example answers of parts (1)--(3) to one decision after the warm-up period. Suppose the last five chunks of an episode had throughputs of 2000, 1500, 2500, 1000 and 3000\,kbps, and the buffer holds 6000\,ms.
\begin{enumerate}
	\item \emph{Features.} The throughput mean is 2000\,kbps and its standard deviation is 707.1\,kbps, so $\texttt{tp\_cv} = 707.1 / 2000 = 0.354$; $\texttt{buffer} = 6.0$\,s.
	\item \emph{Memberships.} $\phi_{\texttt{Stable}} = 0$ because $0.354 > 0.3$, and $\phi_{\texttt{Volatile}} = 1$; $\phi_{\texttt{Low}} = (10-6)/(10-5) = 0.8$ and $\phi_{\texttt{High}} = (6-5)/(10-5) = 0.2$.
	\item \emph{Rule scores.} Rule 1: $0$; rule 2: $1 \times 0.8 = 0.8$; rule 3: $1 \times 0.2 = 0.2$.
	\item \emph{Expert scores.} RobustMPC 0.8 (from rule 2), PaMoE 0.2 (from rule 3), all others 0.
	\item \emph{Selection.} Rules 2 and 3 both fire but select different experts. Rule 2 scores higher because the buffer is closer to \texttt{Low} than to \texttt{High}, so RobustMPC has the highest score. Neither expert is fenced out. If RobustMPC is already the current expert, or stickiness permits the switch, the policy plays RobustMPC's proposal for this chunk; otherwise it keeps the current expert.
\end{enumerate}

\paragraph{Properties that can be checked directly.}
(i) \emph{Size.} A policy has at most 24 rules, each readable as one sentence of the form ``if \dots, then use \dots''.
(ii) \emph{Coverage.} A program can enumerate all combinations of states and list those that no rule covers.
(iii) \emph{Attribution.} Every decision can be traced to its cause: either the highest-scoring rule and its score, as in the worked example, or the mechanism of part~(4) that overrode the rules.

\subsubsection{Form II: fuzzy rule base with router-side features}
\label{app:llm-form2}

Form~II has the same answer format and decision procedure as Form~I
(Algorithm~\ref{alg:llm-fuzzy}). It adds four operators to the feature list.
Unlike the operators of Form~I, which compute only from the observation,
these read information that the router has beyond the observation:
\begin{itemize}
  \item \emph{proposal of $m$}: the bitrate (kbps) or level (0--5) that
    expert $m$ proposes for the current chunk;
  \item \emph{proposal spread}: the highest minus the lowest level proposed
    by the experts in the pool, from 0 (all agree) to 5 (maximal
    disagreement);
  \item \emph{chunks since switch}: the number of chunks downloaded since
    the router last changed expert (0 at the start of a episode);
  \item \emph{stall since switch}: the stall (rebuffering) time accumulated
    over those chunks, in seconds.
\end{itemize}
With these operators, the LLM can write rules that Form~I cannot express, because they depend on information that is not in the observation. For example: ``if the proposals spread by at least 3 levels and the buffer is low, select RobustMPC''; ``if FDASH proposes a bitrate more than 1.2 times the recent mean throughput, do not select FDASH''. Such a rule can be expressed by defining the ratio of the two as a feature and putting a fence on it; or ``if more than 0.5\,s of stall has accumulated since the last switch, select RobustMPC''. The prompt appends the following
description to the operator list (verbatim):
\begin{lstlisting}[style=policy]
Router-side operators: what the router knows beyond the observation.
  {"op":"proposal", "expert":<expert>, "unit":"kbps"|"level"}
        the quality that expert proposes for THIS chunk: its bitrate in kbps (default),
        or its level 0..5 on the ladder [300, 750, 1200, 1850, 2850, 4300] kbps.
  {"op":"proposal_spread"}
        highest minus lowest proposed level over the whole pool this chunk (0 = all agree, 5 = max disagreement).
  {"op":"chunks_since_switch"}
        the number of chunks downloaded since the router's first selection in the episode or, after a switch, since its last switch (0 at the first decision).
  {"op":"rebuffer_since_switch"}
        stall seconds accumulated over those chunks.
\end{lstlisting}

\subsubsection{Form III: Python function}
\label{app:llm-form3}

In Form~III the LLM writes a Python module with two functions. $\texttt{init\_state()}$ is called once at the start of every episode and returns a dictionary $\texttt{state}$, which persists across the decisions of that episode. $\texttt{route(obs, proposals, ctx, state)}$ is called once per decision and returns the name of the expert to select. The arguments of $\texttt{route}$ are the seven observed quantities ($\texttt{obs}$), the level each expert proposes for the current chunk ($\texttt{proposals}$), and facts kept by the router ($\texttt{ctx}$): the index of the decision within the episode, the expert selected at the previous decision, the level played on the previous chunk, and the bitrate ladder. The answer must contain the module in a single Python code block.

These arguments describe only the current decision, whereas a policy may need information from earlier decisions, e.g.\ the throughputs of the last five chunks. Such information is kept in $\texttt{state}$: the router passes the dictionary returned by $\texttt{init\_state()}$ to every call of $\texttt{route}$ in that episode, without reading or modifying it, and discards it when the episode ends. For example, $\texttt{init\_state}$ can return $\texttt{\{"tp": []\}}$, and $\texttt{route}$ can append the throughput of the chunk just downloaded to $\texttt{state["tp"]}$ and use the mean of the last five entries. Each episode has its own dictionary, so episodes do not affect one another.

Within $\texttt{route}$, the LLM may compute any feature, keep any counter or history in $\texttt{state}$, add named constants and helper functions, and structure the decision freely. The language it may use, however, is restricted, and the check enforces these restrictions, e.g.\ no imports (the $\texttt{math}$ module is provided), no classes, no $\texttt{global}$, $\texttt{nonlocal}$ or $\texttt{with}$, no name or attribute starting with an underscore, and only a fixed set of builtins, e.g.\ $\texttt{abs}$, $\texttt{all}$, $\texttt{any}$, $\texttt{bool}$, $\texttt{dict}$, $\texttt{enumerate}$, $\texttt{float}$, $\texttt{int}$, $\texttt{isinstance}$, $\texttt{len}$, $\texttt{list}$, $\texttt{max}$, $\texttt{min}$, $\texttt{range}$, $\texttt{reversed}$, $\texttt{round}$, $\texttt{set}$, $\texttt{sorted}$, $\texttt{sum}$, $\texttt{tuple}$ and $\texttt{zip}$. In addition, the prompt asks for code that is deterministic and at most 200 lines long (a longer module is accepted with a warning), that uses named constants for all thresholds, and that has a comment on each branch saying which situation it handles. The program extracts the code block from the answer and checks it with Algorithm~\ref{alg:llm-check-python} (Appendix~\ref{app:llm-check}); it then loads the two functions and, for every episode, calls $\texttt{init\_state}$ once and $\texttt{route}$ at every decision, without calling the LLM again.

The following example keeps a short ban list of experts that have just caused a stall, and then chooses by buffer level and by how much the experts' proposals agree:
\begin{lstlisting}[style=policy, language=Python]
LOW_BUFFER_S = 8.0     # below this buffer level, play safe
AGREE_LEVELS = 1       # proposals within this many levels count as agreement
BAN_CHUNKS = 10        # an expert that just caused a stall is skipped this long

def init_state():                           # once per session
    return {"banned": {}}

def route(obs, proposals, ctx, state):      # once per decision; returns an expert name
    if obs["rebuffer_ms"] > 0 and ctx["last_expert"] is not None:
        state["banned"][ctx["last_expert"]] = BAN_CHUNKS
    state["banned"] = {m: n - 1 for m, n in state["banned"].items() if n > 1}

    if obs["buffer_size_ms"] / 1000 < LOW_BUFFER_S:
        pick = "robustmpc"                                  # low buffer: conservative planner
    elif max(proposals.values()) - min(proposals.values()) <= AGREE_LEVELS:
        pick = ctx["last_expert"] or "pensieve"             # experts agree: do not switch
    else:
        pick = min(proposals, key=proposals.get)            # disagreement: most conservative proposal
    return "robustmpc" if pick in state["banned"] else pick
\end{lstlisting}

\subsubsection{Comparison of the three forms}

Table~\ref{tab:llm-forms} summarises the forms. Forms~I and~II share the decision procedure and differ only in their inputs, which measures the effect of the router-side information. Forms~II and~III have the same inputs and differ only in the decision procedure, which measures whether the fixed rule structure limits the policy.

\begin{table}[!htbp]
  \centering
  \caption{The three forms of a routing policy.}
  \label{tab:llm-forms}
  \small
  \setlength{\tabcolsep}{3pt}
  \begin{tabular}{p{2.3cm}p{3.6cm}p{3.6cm}p{3.6cm}}
    \toprule
     & Form I & Form II & Form III \\
    \midrule
    LLM writes & features, intervals, rules, fence, stickiness, warm-up (JSON) & as Form I & two Python functions \\
    Inputs & observation, previous bitrate & as Form I, plus proposals and the router's past choices & as Form II \\
    Decision & Algorithm~\ref{alg:llm-fuzzy} & Algorithm~\ref{alg:llm-fuzzy} & written by the LLM \\
    Execution & batched on the GPU, all episodes at once & as Form I & on the CPU, one call per episode and decision \\
    \bottomrule
  \end{tabular}
\end{table}

\subsection{Checking an Answer}
\label{app:llm-check}

An answer is checked before it is used (Algorithms~\ref{alg:llm-check-fuzzy} and~\ref{alg:llm-check-python}). A rejection message is sent back to the LLM verbatim as $\mathit{msg}$ in Algorithm~\ref{alg:llm-gen}; warnings are recorded but do not reject.

\begin{algorithm}[!htbp]
\caption{$\textsc{Check}$ for Forms I and II}
\label{alg:llm-check-fuzzy}
\small
\DontPrintSemicolon
\KwIn{Answer text $y$; allowed operators (Form~I: those of part~(1);
  Form~II: also the four router-side operators).}
parse the text between the first ``\{'' and the last ``\}'' of $y$ as JSON;
  reject on failure\;
reject unless features, membership, rules and the additional parameters are
  all present\;
\ForEach{feature}{
  reject unless its operator is allowed and every feature it references is
    declared before it\;
  reject if it reads an observation field that does not exist, or if it has
    a window outside $\{1, \dots, 64\}$\;
}
\ForEach{state (membership function)}{
  reject unless it has exactly four numbers $a \le b \le c \le d$\;
}
reject unless there are at most 24 rules\;
\ForEach{rule}{
  reject unless it has at least one antecedent, every antecedent names a
    declared feature and state, and its consequent is in $\mathcal{M}$\;
}
reject if two rules have the same set of antecedents
  \tcp*{same consequent: duplicate; different consequents: always tied}
reject unless the default expert is in $\mathcal{M}$ and every expert and
  feature named in the fence is declared\;
\textbf{warn} if some expert is the consequent of no rule\;
\textbf{warn} with the fraction of uncovered points, computed as follows\;
\Indp
for every feature in the membership section, take 21 equally spaced points strictly between the lowest $a$ and the highest $d$ of its states\;
form the grid of all combinations of these points; if it has more than
200{,}000 points, use a random sample of 200{,}000 instead\;
count the points at which every expert scores 0, i.e.\ at which the default
expert would be used\;
\Indm
\end{algorithm}

\begin{algorithm}[!htbp]
\caption{$\textsc{Check}$ for Form III}
\label{alg:llm-check-python}
\small
\DontPrintSemicolon
\KwIn{Answer text $y$.}
extract the Python code block of $y$\;
parse it; reject on a syntax error\;
reject on forbidden constructs, e.g., \texttt{import}, \texttt{class}, \texttt{global},
  \texttt{nonlocal} or \texttt{with}, and on any name or attribute starting
  with ``\_''\;
reject unless top-level functions \texttt{init\_state} and \texttt{route}
  are defined\;
\textbf{warn} if the module is longer than 200 lines\;
\For{each of 20 synthetic episodes of 48 chunks}{
  draw a base bandwidth from $\{200, 800, 2000, 5000, 20000\}$\,kbps, and
    scale it at every chunk by a factor drawn uniformly from $[0.2, 1.5]$\;
  simulate download times, buffer and stalls; draw every proposal uniformly
    from $\{0, \dots, 5\}$\;
  call \texttt{init\_state} once, and \texttt{route} at every chunk, with
    only the allowed builtins available\;
  reject if a call raises an exception or returns anything other than the
    name of an expert in $\mathcal{M}$
    \tcp*{the message gives the line and the inputs}
}
\textbf{warn} if a call takes more than 200\,$\mu$s on average\;
\textbf{warn} if some expert is never returned\;
\end{algorithm}

\subsection{Evaluation Protocol}
\label{app:llm-eval}

A policy is evaluated on every training trace family with 4096 parallel environments, each playing two consecutive episodes, which gives 8192
episodes per family; the random seed is 123. At every decision the router executes $\pi$ and passes the level actually played back to all experts, so that every expert, including those not selected, tracks the true playback state. We report the mean QoE per chunk, its three terms (bitrate, rebuffering penalty and smoothness penalty), and the fraction of decisions on which each expert was selected. Table~A uses the same protocol and seed, so it was computed on exactly the same episodes; its rows are therefore the single-expert baselines for the evaluated policy. These episodes are the ones whose statistics the LLM saw in Tables~A and~B2, so this evaluation is in-sample. Each policy is also evaluated, with the same protocol, on the test split of every family, which is used neither for the measurements nor for selecting a policy.

\newpage

\section{Results of the One-Shot Generation}
\label{app:llm-results}

Three LLMs, Claude Opus~4.7, gpt-oss-120b and Qwen3.8-27B, each generated one policy of each of the three forms with Algorithm~\ref{alg:llm-gen}, which gives nine policies. All nine generations received the same measurements (Tables~A, B and~B2 on the training splits). We first analyse the policies themselves, i.e.\ what the LLMs wrote and whether it agrees with the measurements they were given (Appendix~\ref{app:llm-res-gen}), and then their QoE on the training and test splits of the five trace families (Appendix~\ref{app:llm-res-qoe}).

\subsection{The Generated Policies}
\label{app:llm-res-gen}

We first ask whether each LLM could write a policy in each form at all. Recall from Appendix~\ref{app:llm-check} that every answer of the LLM is passed to $\textsc{Check}_F$ before it is used: an answer that is not a valid policy of form $F$ is rejected, and the LLM is queried again with the rejection message, up to three attempts in total; generation fails if none of the three answers passes. 

\begin{table}[!htbp]
	\centering
	\caption{The nine generated policies. Attempts: number of queries until the answer passed the check. Size: number of rules (Forms~I, II) or lines of the module including comments and blank lines (Form~III). Throughput and variation are the mean and the coefficient of variation of the throughput over the last five chunks; buffer change is the change of the buffer level since the previous chunk. For Forms~I and~II, each state is a membership function $[a, b, c, d]$ (Eq.~\eqref{eq:llm-trapezoid}), with $\infty$ for a state without upper end, and the expert is selected by Eq.~\eqref{eq:llm-score}; Pensieve is also selected during warm-up and when no rule fires. For Form~III, the branches of the code are given as written.}
	\label{tab:llm-res-gen}
	\small
	\setlength{\tabcolsep}{4pt}
	\begin{tabular}{llcc>{\raggedright\arraybackslash}p{10.4cm}}
		\toprule
		LLM & Form & Attempts & Size & Policy \\
		\midrule
		Opus 4.7     & I   & 1 & 2 rules  & \textbf{States:} throughput (Mbps) \textit{NotVeryHigh} $[0, 0, 5.5, 7.5]$, \textit{VeryHigh} $[5.5, 7.5, \infty, \infty]$ \newline
		\textbf{Rules:} throughput \textit{NotVeryHigh} $\to$ Pensieve; throughput \textit{VeryHigh} $\to$ Merina \\\addlinespace
		& II  & 1 & 3 rules  & \textbf{States:} throughput (Mbps) \textit{Normal} $[0, 0, 6, 9]$, \textit{High} $[6, 9, \infty, \infty]$; variation \textit{Stable} $[0, 0, 0.12, 0.2]$, \textit{Volatile} $[0.12, 0.2, \infty, \infty]$ \newline
		\textbf{Rules:} throughput \textit{Normal} $\to$ Pensieve; throughput \textit{High}, variation \textit{Stable} $\to$ Merina; throughput \textit{High}, variation \textit{Volatile} $\to$ Pensieve \\\addlinespace
		& III & 1 & 56 lines & After five chunks: if throughput $> 6$\,Mbps and variation $< 0.15$: Merina; otherwise Pensieve. \\
		\midrule
		gpt-oss-120b & I   & 1 & 5 rules  & \textbf{States:} throughput (Mbps) \textit{low} $[0, 0, 0.8, 1.5]$, \textit{medium} $[0.8, 1.5, 4, 6]$, \textit{high} $[4, 6, \infty, \infty]$; variation \textit{low} $[0, 0, 0.1, 0.2]$; buffer (s) \textit{low\_buf} $[0, 0, 12, 20]$; buffer change (s) \textit{draining} $[-\infty, -\infty, -5, -1]$ \newline
		\textbf{Rules:} throughput \textit{low} $\to$ Pensieve; throughput \textit{medium} $\to$ Pensieve; throughput \textit{high}, variation \textit{low} $\to$ Merina; buffer \textit{low\_buf} $\to$ FDASH; buffer \textit{low\_buf}, buffer change \textit{draining} $\to$ RobustMPC \newline
		\textbf{Fence:} FDASH only if buffer change $\ge -2$\,s \\\addlinespace
		& II  & 1 & 4 rules  & \textbf{States:} throughput (Mbps) \textit{low} $[0, 0, 2, 3]$, \textit{med} $[2, 3, 8, 10]$, \textit{high} $[8, 10, \infty, \infty]$; variation \textit{low} $[0, 0, 0.2, 0.4]$, \textit{high} $[0.3, 0.5, \infty, \infty]$ \newline
		\textbf{Rules:} throughput \textit{low} $\to$ Pensieve; throughput \textit{med} $\to$ Pensieve; throughput \textit{high}, variation \textit{low} $\to$ Merina; throughput \textit{high}, variation \textit{high} $\to$ Pensieve \\\addlinespace
		& III & 2 & 77 lines & If throughput $> 3000$ and variation $< 0.12$: Merina; otherwise Pensieve. Throughput is computed in bit/s. \\
		\midrule
		Qwen3.8-27B  & I   & 2 & 7 rules  & \textbf{States:} throughput (Mbps) \textit{Low} $[0, 0, 1, 2]$, \textit{Mid} $[1, 2, 3, 4]$, \textit{High} $[3, 4, \infty, \infty]$; variation \textit{Stable} $[0, 0, 0.15, 0.3]$, \textit{Volatile} $[0.15, 0.3, \infty, \infty]$; buffer (s) \textit{Low} $[0, 0, 10, 20]$; buffer change (s) \textit{Draining} $[-\infty, -\infty, -0.5, 0.5]$ \newline
		\textbf{Rules:} throughput \textit{Low}, variation \textit{Stable} $\to$ Pensieve; throughput \textit{Mid}, variation \textit{Stable} $\to$ Pensieve; throughput \textit{High}, variation \textit{Stable} $\to$ Merina; throughput \textit{Low} / \textit{Mid} / \textit{High}, variation \textit{Volatile} $\to$ RobustMPC (3 rules); buffer \textit{Low}, buffer change \textit{Draining} $\to$ FDASH \newline
		\textbf{Fence:} RobustMPC only if buffer $\ge 5$\,s; FDASH only if buffer $\ge 2$\,s \\\addlinespace
		& II  & 2 & 7 rules  & Identical to Form~I. \\\addlinespace
		& III & 1 & 111 lines & After five chunks: if throughput $\ge 2500$ and variation $\le 0.10$: Merina if at most 10 chunks remain, otherwise Pensieve; if throughput $\le 1200$ and variation $\ge 0.25$: RobustMPC if buffer $\le 12$\,s, otherwise Pensieve; in all other cases: Pensieve. Throughput is computed in Mbps. \\
		\bottomrule
	\end{tabular}
\end{table}

Table~\ref{tab:llm-res-gen} (column ``Attempts'') shows that all nine generations succeeded, with at most one retry. All three Opus~4.7 policies, the Form~I and~II policies of gpt-oss-120b and the Form~III policy of Qwen3.8-27B needed a single query. The other three were first rejected for format violations only: a Python function name with a leading underscore (gpt-oss-120b), and a derived feature declared before its source (Qwen3.8-27B, Forms~I and~II). Each second answer fixed the named violation without changing the decision logic. The differences reported below therefore lie in the policies' decision logic, not in whether they could be generated.

The policies are small and built from the same few quantities. The rule bases use one to four features and two to seven rules, far below the limit of 24, and the Python modules have 56 to 111 lines. Every feature in a rule is one of the four quantities of Table~B2 (the mean and coefficient of variation of the throughput over five chunks, the buffer level and its change). No LLM used the router-side operators that Form~II adds, no Form~III module reads the values of the experts' proposals, and the Form~II policy of Qwen3.8-27B is identical to its Form~I policy. Opus~4.7 gives the reason in the notes of its Form~II answer: "Buffer level, buffer change and the proposal features were left out because no measurement links them to a change in which expert is best." The statement is accurate: the prompt describes each expert by one average per family, and nothing in it relates the proposals or the switching history to the QoE of an expert.

\subsection{QoE on the Training and Test Splits}
\label{app:llm-res-qoe}

\begin{table}[!htbp]
  \centering
  \caption{QoE of the experts and of the nine policies. Mean QoE per chunk on each family; $\Delta$: mean over the families minus that of Pensieve; choice share: fraction of decisions on which each expert is selected, averaged over the families (P:Pensieve, M:Merina, Pa:PaMoE, R:RobustMPC, F:FDASH). The Form~I and~II policies of Qwen3.8-27B are identical and share one row.}
  \label{tab:llm-res-qoe}
  \resizebox{\textwidth}{!}{%
  \begin{tabular}{ll*{5}{c}cc*{5}{r}}
    \toprule
    & & \multicolumn{5}{c}{QoE} & & & \multicolumn{5}{c}{Choice share (\%)} \\
    \cmidrule(lr){3-7} \cmidrule(lr){10-14}
    \multicolumn{2}{l}{Policy} & FCC-16 & FCC-18 & Oboe & Puffer-21 & Puffer-22 & Mean & $\Delta$ & P & M & Pa & R & F \\
    \midrule
    \multicolumn{14}{l}{\emph{(a) Training splits (in-sample: the sessions of Table~A)}} \\
    \multicolumn{2}{l}{Pensieve}  & 0.8657 & 3.0055 & 2.2693 & 1.1509 & 0.8485 & 1.6280 & 0 & & & & & \\
    \multicolumn{2}{l}{Merina}    & 0.8613 & 3.0179 & 2.2626 & 1.1276 & 0.8321 & 1.6203 & $-0.0077$ & & & & & \\
    \multicolumn{2}{l}{PaMoE}     & 0.8604 & 3.0056 & 2.2616 & 1.1308 & 0.8349 & 1.6187 & $-0.0093$ & & & & & \\
    \multicolumn{2}{l}{RobustMPC} & 0.7174 & 2.9339 & 2.1774 & 0.7489 & 0.6151 & 1.4385 & $-0.1894$ & & & & & \\
    \multicolumn{2}{l}{FDASH}     & 0.7518 & 2.6583 & 1.9099 & 0.9508 & 0.7070 & 1.3956 & $-0.2324$ & & & & & \\
    \multicolumn{2}{l}{Family oracle} & 0.8657 & 3.0179 & 2.2693 & 1.1509 & 0.8485 & 1.6305 & $+0.0025$ & & & & & \\
    \cmidrule(lr){1-14}
    Opus 4.7     & I   & 0.8657 & 3.0061 & 2.2693 & 1.1501 & 0.8486 & 1.6280 & $-0.0000$ & 94 & 6  & 0 & 0  & 0 \\
                 & II  & 0.8657 & 3.0054 & 2.2693 & 1.1512 & 0.8485 & 1.6280 & $+0.0000$ & 96 & 4  & 0 & 0  & 0 \\
                 & III & 0.8657 & 3.0057 & 2.2693 & 1.1504 & 0.8485 & 1.6279 & $-0.0001$ & 94 & 6  & 0 & 0  & 0 \\
    gpt-oss-120b & I   & 0.8554 & 2.9870 & 2.2428 & 1.1478 & 0.8346 & 1.6135 & $-0.0145$ & 89 & 8  & 0 & 1  & 2 \\
                 & II  & 0.8657 & 3.0059 & 2.2693 & 1.1509 & 0.8485 & 1.6281 & $+0.0001$ & 97 & 3  & 0 & 0  & 0 \\
                 & III & 0.8588 & 3.0176 & 2.2659 & 1.1430 & 0.8419 & 1.6254 & $-0.0025$ & 51 & 49 & 0 & 0  & 0 \\
    Qwen3.8-27B  & I, II & 0.8310 & 2.9551 & 2.1765 & 0.9487 & 0.7573 & 1.5337 & $-0.0943$ & 58 & 14 & 0 & 21 & 7 \\
                 & III & 0.8657 & 3.0039 & 2.2653 & 1.1542 & 0.8486 & 1.6275 & $-0.0005$ & 96 & 0  & 0 & 4  & 0 \\
    \midrule
    \multicolumn{14}{l}{\emph{(b) Test splits (traces not used in the generation)}} \\
    \multicolumn{2}{l}{Pensieve}  & 0.7341 & 3.1005 & 2.1454 & 0.9942 & 0.9449 & 1.5838 & 0 & & & & & \\
    \multicolumn{2}{l}{Merina}    & 0.7372 & 3.1131 & 2.1336 & 0.9619 & 0.9153 & 1.5722 & $-0.0116$ & & & & & \\
    \multicolumn{2}{l}{PaMoE}     & 0.7342 & 3.1070 & 2.1337 & 0.9671 & 0.9228 & 1.5730 & $-0.0109$ & & & & & \\
    \multicolumn{2}{l}{RobustMPC} & 0.5789 & 3.0440 & 2.0384 & 0.5306 & 0.6468 & 1.3677 & $-0.2161$ & & & & & \\
    \multicolumn{2}{l}{FDASH}     & 0.6441 & 2.7488 & 1.8016 & 0.7448 & 0.7710 & 1.3421 & $-0.2418$ & & & & & \\
    \multicolumn{2}{l}{Family oracle} & 0.7372 & 3.1131 & 2.1454 & 0.9942 & 0.9449 & 1.5870 & $+0.0031$ & & & & & \\
    \cmidrule(lr){1-14}
    Opus 4.7     & I   & 0.7341 & 3.1007 & 2.1454 & 0.9883 & 0.9448 & 1.5827 & $-0.0012$ & 93 & 7  & 0 & 0  & 0 \\
                 & II  & 0.7341 & 3.1005 & 2.1454 & 0.9923 & 0.9449 & 1.5834 & $-0.0004$ & 96 & 4  & 0 & 0  & 0 \\
                 & III & 0.7341 & 3.1008 & 2.1454 & 0.9926 & 0.9448 & 1.5835 & $-0.0003$ & 94 & 6  & 0 & 0  & 0 \\
    gpt-oss-120b & I   & 0.7305 & 3.0819 & 2.1127 & 0.9921 & 0.9382 & 1.5711 & $-0.0128$ & 89 & 8  & 0 & 1  & 2 \\
                 & II  & 0.7341 & 3.1008 & 2.1454 & 0.9938 & 0.9449 & 1.5838 & $-0.0000$ & 97 & 3  & 0 & 0  & 0 \\
                 & III & 0.7292 & 3.1128 & 2.1337 & 0.9842 & 0.9299 & 1.5780 & $-0.0058$ & 53 & 47 & 0 & 0  & 0 \\
    Qwen3.8-27B  & I, II & 0.7124 & 3.0473 & 2.0481 & 0.7557 & 0.8278 & 1.4783 & $-0.1056$ & 57 & 14 & 0 & 22 & 8 \\
                 & III & 0.7358 & 3.0982 & 2.1399 & 0.9958 & 0.9424 & 1.5824 & $-0.0014$ & 95 & 0  & 0 & 5  & 0 \\
    \bottomrule
  \end{tabular}}
\end{table}

Each policy is evaluated with the protocol of Appendix~\ref{app:llm-eval} on the training and on the test splits of the five families. The training splits contain exactly the episodes from which Table~A was computed, so this evaluation is in-sample; the test splits were not used at any step of the generation. The five experts are evaluated alone with the same protocol. The environment draws the trace, the start position within it and the link noise of every episode from the seed, the index of the environment and the index of the episode, independently of the bitrates played. All policies and experts therefore play the same episodes on a given split, and their QoE values differ only through the routing decisions. Table~\ref{tab:llm-res-qoe} reports the mean QoE per chunk on each family and, as $\Delta$, its mean over the families minus that of Pensieve, the best single expert on both splits.  As a reference, we include the \emph{family oracle}, which uses the best expert on each family. It corresponds to a router that identifies every family correctly, and thus marks the best QoE achievable from the family-level measurements.

The single-expert rows set the scale. Pensieve is the best expert on four training and three test families, and Merina, the best on the others, leads by at most $0.013$, so the family oracle exceeds Pensieve by only $0.0025$ (training) and $0.0031$ (test). Playing RobustMPC or FDASH throughout instead costs $0.19$ to $0.24$. The QoE of a policy is therefore decided mainly by how often it selects these two experts.

The four policies that select only Pensieve and Merina (the three of Opus 4.7 and the Form II policy of gpt-oss-120b) obtain the QoE of Pensieve, with $\Delta$ between $-0.0012$ and $+0.0001$. They do not recover Merina's advantage on FCC-18: on its training split the policies of Opus 4.7 obtain $3.0054$ to $3.0061$, against $3.0055$ for Pensieve and $3.0179$ for Merina. Of the three policies with rules for RobustMPC or FDASH, two lose QoE. Form I and II policy of The Qwen3.8-27B hands 21–22\% of the decisions to RobustMPC and 7–8\% to FDASH and loses $0.094$ (training) and $0.106$ (test), most on Puffer-21 ($0.949$ against $1.151$ for Pensieve on the training split); the Form I policy of gpt-oss-120b policy selects them on 3\% of the decisions and loses $0.013$ to $0.015$. The third, the Form III policy of Qwen3.8-27B selects RobustMPC on 4–5\% of the decisions and stays within $0.0014$ of Pensieve.

In summary, every LLM produced a valid policy of every form within two attempts. Opus 4.7 followed the measurements in all three forms and matched the best single expert, with little more available; the other two LLMs wrote rules that Table A does not support or code that does not compute what it states. Since no LLM used the inputs that Form II adds, and Opus 4.7 wrote essentially the same rule in Forms II and III, the experiment does not separate the effect of the form itself.

\subsection{QoE outside the Training Distribution}
\label{app:llm-res-ood}

The nine policies were generated from measurements on the five 3G families only. We evaluate them unchanged on the seven datasets outside the training distribution of Appendix~\ref{app:converge}. The protocol is that of Appendix~\ref{app:llm-res-qoe} (seed 123), so the values of the experts differ slightly from those of Table~\ref{tab:breakdown}, which average five seeds. Table~\ref{tab:llm-res-ood} has the layout of Table~\ref{tab:llm-res-qoe}, with the means taken over the seven datasets; $\Delta$ is still measured from Pensieve, although Pensieve is no longer the best single expert.

\begin{table}[!htbp]
	\centering
	\caption{QoE of the experts and of the nine policies outside the training distribution. Mean QoE per chunk on each dataset; $\Delta$: mean over the datasets minus that of Pensieve; choice share: fraction of decisions on which each expert is selected, averaged over the datasets (P:Pensieve, M:Merina, Pa:PaMoE, R:RobustMPC, F:FDASH). The Form~I and~II policies of Qwen3.8-27B are identical and share one row.}
	\label{tab:llm-res-ood}
	\resizebox{\textwidth}{!}{%
		\begin{tabular}{ll*{7}{c}cc*{5}{r}}
			\toprule
			& & \multicolumn{7}{c}{QoE} & & & \multicolumn{5}{c}{Choice share (\%)} \\
			\cmidrule(lr){3-9} \cmidrule(lr){12-15}
			\multicolumn{2}{l}{Policy} & HSR & Norway3G & Lumos4G & Lumos5G & SolisWi-Fi & Ghent & Lab & Mean & $\Delta$ & P & M & Pa & R & F \\
			\midrule
			\multicolumn{2}{l}{Pensieve} & 3.0405 & $-0.4654$ & 2.2911 & 1.1793 & 1.7671 & 2.7600 & 3.3670 & 1.9914 & 0 & & & & & \\
			\multicolumn{2}{l}{Merina} & 3.0286 & $-0.2238$ & 2.2420 & 1.5828 & 1.6230 & 1.9753 & 3.1916 & 1.9171 & $-0.0743$ & & & & & \\
			\multicolumn{2}{l}{PaMoE} & 3.0327 & $-0.1301$ & 1.6981 & 1.0509 & 1.1388 & 1.9571 & 2.4969 & 1.6063 & $-0.3850$ & & & & & \\
			\multicolumn{2}{l}{RobustMPC} & 2.9065 & $-0.5917$ & 2.7657 & 3.8077 & 2.1557 & 3.0156 & 3.6649 & 2.5320 & $+0.5407$ & & & & & \\
			\multicolumn{2}{l}{FDASH} & 2.9427 & $-4.9827$ & $-3.8234$ & 4.0255 & $-13.4813$ & $-3.0205$ & 1.3407 & $-2.4284$ & $-4.4198$ & & & & & \\
			\multicolumn{2}{l}{Family oracle} & 3.0405 & $-0.1301$ & 2.7657 & 4.0255 & 2.1557 & 3.0156 & 3.6649 & 2.6482 & $+0.6569$ & & & & & \\
			\cmidrule(lr){1-15}
			Opus 4.7     & I     & 3.0517 & $-0.4654$ & 2.3357 & 1.4332 & 1.6263 & 2.2637 & 3.0503 & 1.8994 & $-0.0920$ & 31 & 69 & 0 & 0 & 0 \\
			& II    & 3.0420 & $-0.4654$ & 2.3045 & 1.3492 & 1.7338 & 2.5799 & 3.2066 & 1.9644 & $-0.0270$ & 68 & 32 & 0 & 0 & 0 \\
			& III   & 3.0424 & $-0.4654$ & 2.2533 & 1.2333 & 1.7210 & 2.5235 & 3.1736 & 1.9259 & $-0.0654$ & 70 & 30 & 0 & 0 & 0 \\
			gpt-oss-120b & I     & 3.0292 & $-0.4057$ & 2.0848 & 1.2388 & 1.6807 & 2.4064 & 2.9510 & 1.8550 & $-0.1363$ & 52 & 34 & 0 & 3 & 10 \\
			& II    & 3.0451 & $-0.4654$ & 2.2734 & 1.3704 & 1.6467 & 2.2548 & 3.0531 & 1.8826 & $-0.1088$ & 54 & 46 & 0 & 0 & 0 \\
			& III   & 3.0342 & $-0.3606$ & 2.2876 & 1.2394 & 1.7364 & 2.6145 & 3.3123 & 1.9805 & $-0.0108$ & 68 & 32 & 0 & 0 & 0 \\
			Qwen3.8-27B  & I, II & 2.9370 & $-0.8904$ & 1.6522 & 2.7875 & 1.3119 & 1.9403 & 2.9890 & 1.8182 & $-0.1731$ & 28 & 40 & 0 & 27 & 5 \\
			& III   & 3.0376 & $-0.4503$ & 2.3080 & 1.1792 & 1.7669 & 2.7842 & 3.3672 & 1.9990 & $+0.0076$ & 97 & 0 & 0 & 3 & 0 \\
			\bottomrule
	\end{tabular}}
\end{table}

The results on the 3G families do not predict those outside them. Outside them, $\Delta$ of the nine policies lies between $-0.173$ and $+0.0076$, far below that of RobustMPC alone ($+0.541$) and of the family oracle ($+0.657$) (Table~\ref{tab:llm-res-ood}). The prompt describes the experts only on the 3G families, where RobustMPC and FDASH are the weakest, so a policy written from it has no basis for selecting RobustMPC where it is the best. The one-shot policies therefore do not carry over to networks outside the training distribution.

\subsection{A Learned Gate as a Baseline}
\label{app:llm-mlp-gate}

The LLM writes a policy from summaries of the expert rollouts (Tables~A and~B2). As a baseline we fit a small neural gate to the same rollouts, sample by sample. The gate sees the same training families and routes among the same five experts; at run time it observes five of the quantities that Form~I may use, and none of the router-side inputs of Forms~II and~III. The two thus differ in how the policy is obtained, written from summaries or fitted to every sample, and the gate additionally in having fewer inputs.

\paragraph{\textbf{Samples.}} The gate is fitted to the rollouts from which Tables~A and~B2 are computed: on each of the five training families, each expert plays 8192 episodes alone (4096 parallel environments, two episodes each, seed~123), with 47 decisions per episode. Decision $t$ of an episode played by expert $m$ yields one sample $(x_t, m, G_t)$: $x_t$ is the feature vector the router would observe before that decision, and $G_t$ is the QoE that $m$ obtains from the chunk chosen at decision $t$ to the end of the episode,
\begin{equation}
	G_t = \sum_{t'=t}^{I-1} r_{t'+1} ,
	\label{eq:gate-label}
\end{equation}
where $r_{t'+1}$ is the QoE of chunk $V_{t'+1}$, the chunk whose bitrate is chosen at decision $t'$ (Appendix~\ref{sec:system-model}). This gives $5 \times 5 \times 8192 \times 47 = 9{,}625{,}600$ samples, $1{,}925{,}120$ per family, with every expert and every family equally represented. The episodes of the environments whose index is congruent to $9 \bmod 10$ (409 of 4096, both episodes of each) are held out for model selection; no test split and no family outside the training distribution is used.

\paragraph{\textbf{Features.}} The features are the four quantities of Table~B2 (mean throughput over the last five chunks and its coefficient of variation, buffer level, buffer change), computed with the feature operators of Form~I (Appendix~\ref{app:llm-form1}), and the number of remaining chunks, an observation field that the LLM may also use. The number of remaining chunks is needed because $G_t$ scales with it. Each feature is standardised with its mean and standard deviation over the training samples.

\paragraph{\textbf{Why the return.}} Since the episode is played by $m$ alone, $G_t$ is a sample of $V_m(h_t)$, the value of the state $h_t$ (the history of Appendix~\ref{app:llm-policy}) when $m$ plays from decision $t$ on. Routing to the expert of highest value then has the following guarantee, a special case of policy improvement over several base policies.

\emph{Proposition.} Let $V_m$ be the value of expert $m \in \mathcal{M}$, $V_{\max}(h) = \max_m V_m(h)$, and let $\pi$ play, in every state $h$, the action of $\hat m(h) = \arg\max_m V_m(h)$. Then $V_\pi(h) \ge V_{\max}(h)$ for every $h$.

\emph{Proof.} Let $\hat m = \hat m(h)$ and let $\pi_{\hat m}$ denote the policy of expert $\hat m$. Since $\pi(h) = \pi_{\hat m}(h)$ and $V_{\hat m} \le V_{\max}$, we have $V_{\max}(h) = V_{\hat m}(h) = \mathbb{E}[r(h, \pi(h)) + V_{\hat m}(h')] \le \mathbb{E}[r(h, \pi(h)) + V_{\max}(h')]$, i.e.\ $V_{\max} \le \mathcal{B}_\pi V_{\max}$, where $\mathcal{B}_\pi$ is the Bellman operator of $\pi$. Since $\mathcal{B}_\pi$ is monotone and, over the finite horizon, reaches $V_\pi$ after $I - 1$ applications, $V_{\max} \le V_\pi$. $\square$

Taking the expectation over the initial state of the episodes of a family gives $\mathbb{E}[V_\pi(h_1)] \ge \max_m \mathbb{E}[V_m(h_1)]$, so $\pi$ is at least as good as the family oracle of Table~\ref{tab:llm-res-qoe} on every family.

\paragraph{\textbf{Model and training.}} The gate $f_\theta : \mathbb{R}^5 \to \mathbb{R}^5$ is a multilayer perceptron with one hidden layer of 32 tanh units, 357 parameters in total; output $f_\theta(x)_m$ estimates $V_m$ at the state described by $x$. Each sample supervises only the output of the expert that produced it:
\begin{equation}
	\mathcal{L}(\theta) = \frac{1}{|\mathcal{S}|} \sum_{(x, m, G) \in \mathcal{S}} \bigl( f_\theta(x)_m - G \bigr)^2 ,
	\label{eq:gate-loss}
\end{equation}
where $\mathcal{S}$ is the set of training samples. For optimisation, $G$ is standardised with its mean and standard deviation over the training samples. The map is the same for all experts, so it does not change the argmax, and it is folded into the output layer afterwards. At run time the gate selects $\arg\max_m f_\theta(x_t)_m$ at every decision, without warm-up or stickiness.

\begin{figure}[!htbp]
	\centering
	\includegraphics[width=0.5\linewidth]{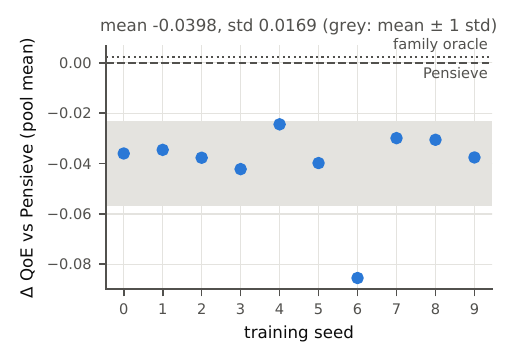}
	\caption{$\Delta$ of ten base gates that differ only in the training seed (200 epochs). Grey band: mean $\pm$ one standard deviation ($-0.040 \pm 0.017$).}
	\label{fig:gate-seed}
\end{figure}

\paragraph{\textbf{Evaluation.}} The gate is evaluated with the protocol of Appendix~\ref{app:llm-eval} On the training splits its $\Delta$ (Eq.~\eqref{eq:gate-delta}) is $-0.036$ (training seed~0, Figure~\ref{fig:gate-seed}); on the test splits and on four datasets outside the training distribution it is compared with the Form~I policy of Opus~4.7 in Appendix~\ref{app:llm-routing-worth}. At seed~123 the training splits replay exactly the episodes the gate was fitted on, so its training-split results are in-sample to a larger degree than those of the LLM policies, which saw only summaries of these episodes.

\paragraph{\textbf{Sensitivity to training.}} To see whether the result depends on how the gate is trained, we vary one setting at a time and keep the others at their base values (one hidden layer of 32 units, learning rate $10^{-3}$, training seed~0). All gates are fitted to the same samples and evaluated on the training splits of the five families with the protocol of Appendix~\ref{app:llm-eval}. We report
\begin{equation}
	\Delta = \frac{1}{5} \sum_{f} \bigl( \mathrm{QoE}_f(\text{gate}) - \mathrm{QoE}_f(\text{Pensieve}) \bigr),
	\label{eq:gate-delta}
\end{equation}
where $\mathrm{QoE}_f(\cdot)$ is the QoE per chunk averaged over the 8192 episodes of the training split of family $f$, and the sum runs over the five families with equal weight. $\mathrm{QoE}_f(\text{Pensieve})$ is the QoE of Pensieve playing alone, the best single expert (Table~A). Since all policies play the same episodes, $\Delta$ differs from zero only through the routing decisions; $\Delta < 0$ means that the gate is worse than Pensieve alone. On the same scale the family oracle has $\Delta = +0.0025$.

\begin{figure}[!htbp]
	\centering
	\includegraphics[width=\linewidth]{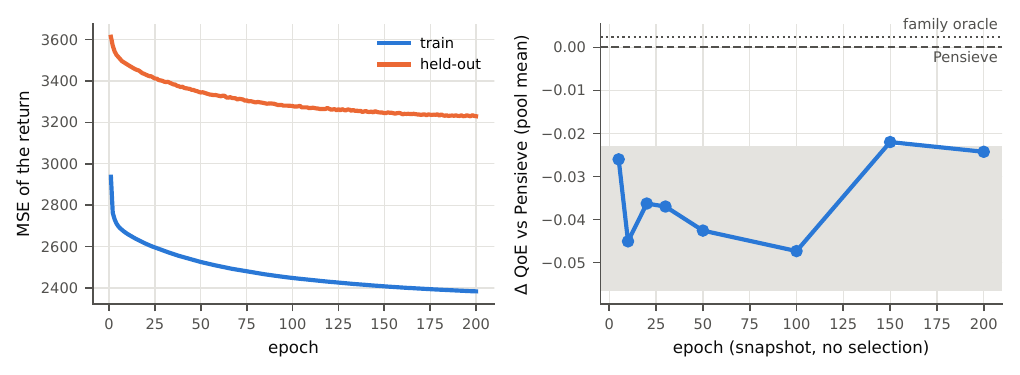}
	\caption{Base gate trained for 200 epochs. Left: training and held-out loss. Right: $\Delta$ of the parameters after the given number of epochs, without selection. Dashed: Pensieve; dotted: family oracle.}
	\label{fig:gate-epoch}
\end{figure}

\emph{Epochs.} In this experiment only, we train the base gate once for 200 epochs and evaluate its parameters as they are after 5, 10, 20, 30, 50, 100, 150 and 200 epochs, without selecting the epoch with the lowest held-out loss (Figure~\ref{fig:gate-epoch}). The held-out loss decreases monotonically, from 3516 after 5 epochs to 3230 after 200 epochs, and by only 0.5\% over the last 50. $\Delta$ does not follow it: it ranges from $-0.047$ to $-0.022$ with no trend, and is negative after every number of epochs. We therefore train every gate in the remaining experiments for 200 epochs and, as in the base setting, evaluate the parameters of the epoch with the lowest held-out loss. For example, the held-out loss of the gate with 32 hidden units is lowest after epoch 197, so this gate is evaluated with its parameters after epoch 197, not after epoch 200.

\begin{figure}[!htbp]
	\centering
	\includegraphics[width=\linewidth]{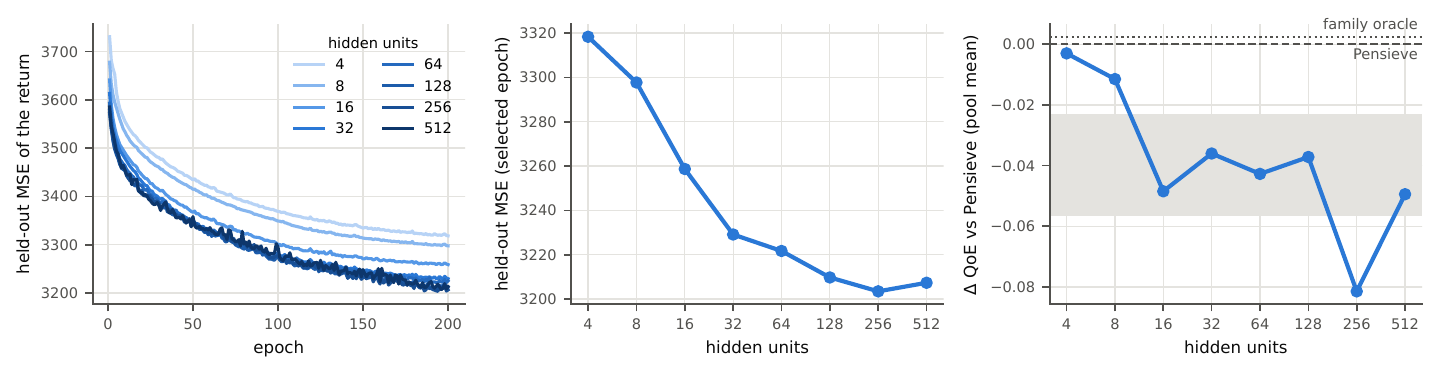}
	\caption{Width of the hidden layer (200 epochs, learning rate $10^{-3}$). Left: held-out loss per epoch. Middle: held-out loss of the selected epoch. Right: $\Delta$; grey band: mean $\pm$ one standard deviation over ten training seeds.}
	\label{fig:gate-width}
\end{figure}

\emph{Width.} With 4 to 512 hidden units (Figure~\ref{fig:gate-width}), the held-out loss decreases from 3318 at 4 units to 3203 at 256 units and does not decrease further at 512 units (3207). The two smallest gates come closest to Pensieve, with $\Delta = -0.003$ (4 units) and $-0.011$ (8 units). They do so by selecting Pensieve on 98\% (4 units) and 74\% (8 units) of the decisions, against 48\% for the base gate (32 units); the gate with 4 units thus behaves almost like Pensieve alone. From 16 to 512 units, except at 256 units, $\Delta$ lies between $-0.049$ and $-0.036$, within the band, i.e.\ within the variation caused by the training seed alone. At 256 units, $\Delta = -0.081$, although this width has the lowest held-out loss.

Beyond 8 units, a wider gate therefore fits the return better but does not route better: from 16 to 512 units, $\Delta$ shows no trend and, except at 256 units, stays within the variation caused by the training seed, while the held-out loss decreases. The higher QoE of the two smallest gates comes from selecting Pensieve on most decisions, not from better routing. 

\begin{figure}[!htbp]
	\centering
	\includegraphics[width=\linewidth]{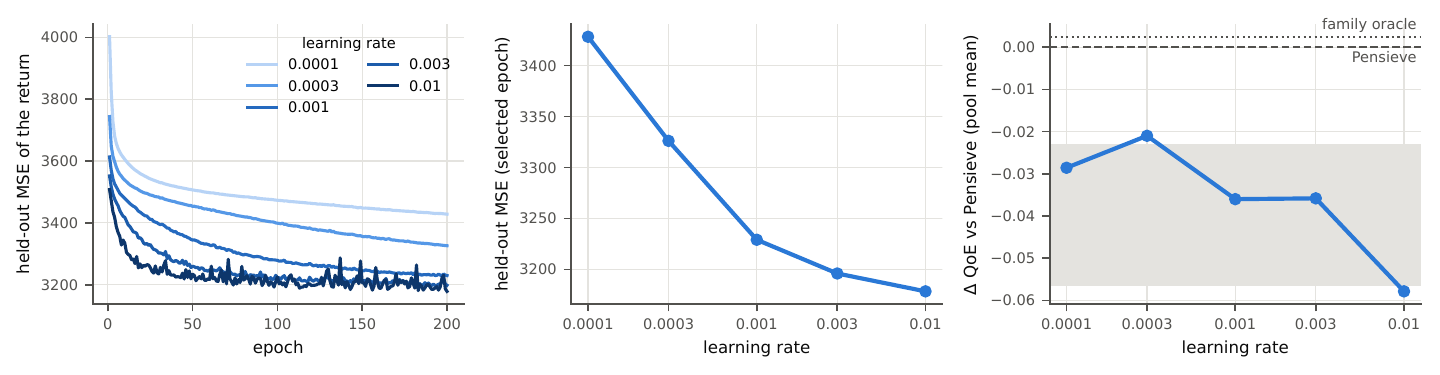}
	\caption{Learning rate (200 epochs, 32 hidden units); panels as in Figure~\ref{fig:gate-width}.}
	\label{fig:gate-lr}
\end{figure}

\emph{Learning rate.} We vary the learning rate from $10^{-4}$ to $10^{-2}$ (Figure~\ref{fig:gate-lr}). The smaller the learning rate, the more slowly the held-out loss decreases (left). With $10^{-3}$ and below, the held-out loss is still decreasing at epoch 200; with $3 \cdot 10^{-3}$ and $10^{-2}$ it levels off after about 100 epochs, and  with $10^{-2}$ it fluctuates from epoch to epoch. The held-out loss of the selected epoch decreases monotonically with the learning rate, from 3428 at $10^{-4}$ to 3179 at $10^{-2}$ (middle). $\Delta$ does not follow it (right): it is $-0.029$, $-0.021$, $-0.036$, $-0.036$ and $-0.058$. All five gates remain below Pensieve.

Adjusting the learning rate thus changes how well the gate fits the return, but not how well it routes. Over the range of rates, the held-out loss of the selected epoch decreases by 7\%, from 3428 to 3179, whereas $\Delta$ shows no trend with the learning rate. Even the rate that fits the return best ($10^{-2}$) does not give a higher QoE: its $\Delta$ ($-0.058$) is the lowest of the five.

\emph{Seed.} To measure how much $\Delta$ varies when only the randomness of training changes, we train ten gates with the base setting that differ only in the training seed, which sets the initial parameters and the order of the samples (Figure~\ref{fig:gate-seed}). All ten gates are below Pensieve. Their $\Delta$ ranges from $-0.086$ to $-0.024$, with mean $-0.040$ and standard deviation $0.017$. This spread comes mostly from one gate: without seed~6 ($-0.086$), the other nine lie between $-0.042$ and $-0.024$, with standard deviation $0.006$. A single training run can therefore differ from another by up to $0.06$ in $\Delta$ without any change of setting.

Across all settings, no gate reaches Pensieve, and a lower held-out loss does not come with a higher QoE. The training settings thus have little effect on the result, and we keep the base setting for all further experiments: one hidden layer of 32 tanh units, learning rate $10^{-3}$, batch size 4096, 200 epochs, and the parameters of the epoch with the lowest held-out loss.

\newpage

\subsection{Is Routing Worth It?}
\label{app:llm-routing-worth}

\begin{figure*}[!htbp]
	\centering
	\includegraphics[width=\textwidth]{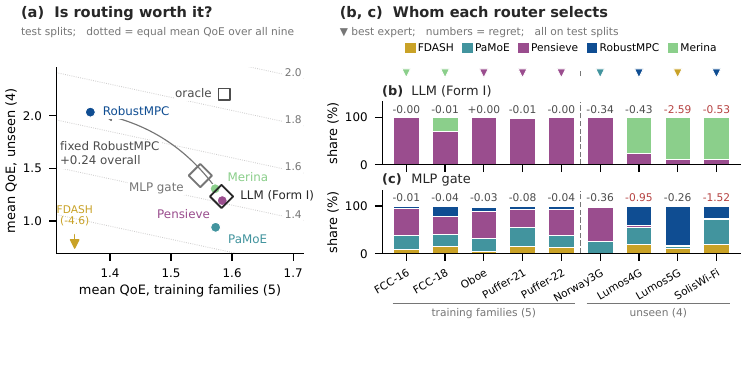}
	\caption{Routing on the training families and outside them, all on test splits (seed~123). (a)~Mean QoE over the five training families ($x$-axis) and over the four unseen datasets ($y$-axis); dotted: equal mean QoE over all nine datasets. Filled circles: experts alone; diamonds: routers; square: family oracle. (b, c)~Share of decisions on which each expert is selected by the Form~I policy of Opus~4.7 and by the gate; numbers: regret with respect to the best single expert on each dataset (red: below $-0.5$); triangles: the expert with the highest QoE when playing alone on that dataset.}
	\label{fig:routing-overview}
\end{figure*}

On the training families neither router reaches Pensieve, and even the family oracle exceeds Pensieve by only $0.0025$. On these families routing therefore has little to gain.  Routing can pay off only where different experts are best on different networks, and this is the case outside the training distribution. RobustMPC, for example, has the second-lowest mean QoE over the training families but the highest mean QoE over the unseen datasets (Figure~\ref{fig:routing-overview}(a)). We therefore ask two questions. First, does routing outperform a single fixed expert once the networks change? Second, which experts do two routers that have seen only the training families select on networks outside them?

\paragraph{\textbf{Setup.}} We compare the five experts playing alone, the family oracle, which plays on each dataset the expert that is best there, and two routers. The first is the policy that Opus~4.7 writes in Form~I. The three forms of Opus~4.7 give similar results (Table~\ref{tab:llm-res-qoe}), so we show only this one. The second is the MLP gate in the default setting of Appendix~\ref{app:llm-mlp-gate}. In the seed experiment of that appendix, ten gates with different training seed reach similar QoE on the training families. We therefore use the gate with training seed 0. We evaluate them on the test splits of the five training families and on the test splits of the four datasets outside the training distribution: Norway3G, Lumos4G, Lumos5G and SolisWi-Fi. All evaluations follow the protocol of Appendix~\ref{app:llm-eval} (seed~123, 4096 environments, two episodes each), so every method plays the same episodes on a given dataset, and no result is in-sample. For every method and dataset we record the mean QoE per chunk and the share of decisions on which each expert is selected. The regret of a router on a dataset is its QoE minus that of the best single expert there.

\paragraph{\textbf{Results.}}  In Figure~\ref{fig:routing-overview}(a), each method is a point. Its horizontal coordinate is the mean QoE over the five training families, and its vertical coordinate the mean QoE over the four unseen datasets; points further to the upper right are better. The dotted lines connect points with equal mean QoE over all nine datasets, $(5x + 4y)/9$. FDASH lies far below the axes on the unseen datasets ($-4.6$) and is marked by an arrow. Figures~\ref{fig:routing-overview}(b) and~(c) show, for each dataset, the share of decisions on which each expert is selected, with the regret above each bar. The triangle above each bar marks, by its colour, the expert that reaches the highest QoE when playing alone on that dataset. If this colour does not appear in the bar, the router never selects that expert on that dataset.

The ranking of the experts reverses between the two axes. Pensieve and PaMoE are among the best experts on the training families and among the worst on the unseen datasets; RobustMPC is the second worst on the training families (1.368) and the best on the unseen datasets (2.034). As a result, routing does not pay off. A fixed RobustMPC reaches a mean QoE of 1.664 over all nine datasets, $0.24$ above the LLM policy (1.427). The gate (1.496) is close to the LLM policy and also far below RobustMPC.

The two routers behave in opposite ways on the two groups of datasets. On the training families, the LLM policy selects Pensieve on almost every decision and has regrets between $0$ and $-0.01$. The gate mixes Pensieve, PaMoE and FDASH and is slightly worse on every family (regrets between $-0.01$ and $-0.08$). On the unseen datasets, the LLM policy keeps Pensieve on Norway3G and selects mostly Merina on the other three; its regret reaches $-2.59$ on Lumos5G. The gate selects RobustMPC on most decisions on Lumos5G (regret $-0.26$), but loses $0.95$ on Lumos4G and $1.52$ on SolisWi-Fi. Neither router tracks the best expert on the unseen datasets, which is PaMoE on Norway3G, RobustMPC on Lumos4G and SolisWi-Fi, and FDASH on Lumos5G.

\paragraph{\textbf{Conclusion.}} Both routers are limited by what they were built from, not by how the policy is expressed or obtained. The LLM policy is written from measurements on the training families, and the gate is fitted to rollouts on the same families. On these trace families RobustMPC is among the weakest experts, so neither router has any basis for preferring it on the networks where it is the best. That the gate selects RobustMPC on Lumos5G is an extrapolation beyond its training data rather than something it learned there. A router fixed at deployment thus inherits the ranking of the experts on the training families, and loses wherever that ranking changes. Closing this gap requires that the router continue to learn after deployment. It has to measure how the experts perform on the networks it actually meets and revise its policy accordingly, while keeping the near-zero regret it already has on the training families.

\newpage

\section{Continual Revision of the Routing Policy}
\label{app:continual}

The policies of Appendix~\ref{app:llm-routing} are written once, from measurements on the training splits of the five 3G trace families, and are then fixed. On every 3G family the best single expert is Pensieve or Merina, and the one-shot policies select mostly these two. Outside these families the best single expert is RobustMPC on Lumos4G and SolisWi-Fi, FDASH on Lumos5G and PaMoE on Norway3G (Appendix~\ref{app:llm-res-ood}). On the 3G families RobustMPC and FDASH are among the weakest experts, so a policy written from the 3G measurements has no basis for selecting them on these networks. A deployed router, however, keeps meeting new networks. We assume that when it meets one, only a few episodes can be recorded on it before the policy has to be revised, not the thousands used to design the original policy. This section describes a procedure that revises the deployed policy from a few recorded episodes per new network, and compares two ways of producing the revision under the same data and the same acceptance test: the LLM rewrites the policy from summaries of the episodes, as a human designer would, or the learned gate of Appendix~\ref{app:llm-mlp-gate} is refitted to them.

\subsection{Setting}
\label{app:continual-setting}

\paragraph{\textbf{Stages.}} The procedure runs in stages $s = 0, 1, \dots, S$ with $S = 4$. At stage $s = 0$ the deployed policy $\pi_0$ is the one-shot policy: for the LLM, the Form~I policy of Opus~4.7 (Appendix~\ref{app:llm-res-gen}); for the gate, the gate in the default setting of Appendix~\ref{app:llm-mlp-gate} with training seed~0. At each stage $s \ge 1$ one new network arrives, in the order Lumos4G, Lumos5G, SolisWi-Fi, Norway3G, i.e.\ two cellular datasets, a Wi-Fi dataset and a second 3G dataset. A network is represented by the training split $\mathcal{D}_s$ of its dataset; its test split is never shown to either reviser. It serves only to measure the policy after each stage: the policy deployed at the end of stage $s$ plays the test splits of all nine datasets, i.e.\ of the five 3G families and of the four arriving networks, including those that have not arrived yet, and we report its mean QoE on each.

\paragraph{\textbf{Entries.}} After stage $s$ the router has met the \emph{entries}
\begin{equation*}
	\mathcal{E}_s = \{\text{FCC-16}, \text{FCC-18}, \text{Oboe}, \text{Puffer-21}, \text{Puffer-22}\} \cup \{\text{Network-}1, \dots, \text{Network-}s\},
\end{equation*}
$|\mathcal{E}_s| = 5 + s$. Each 3G family is one entry, so that the five families of the original design keep the weight they had in it. The new networks are named Network-$1$ to Network-$4$ in everything the LLM sees; the dataset name is withheld, so that the revision has to rest on the measurements and not on what the LLM may know about a named dataset.

\paragraph{\textbf{What is available on a new network.}} When network $s$ arrives, the deployed policy plays $N_{\mathrm{c}} = 16$ episodes on it; an episode is one playback of the video, $I = 48$ chunks of $L = 4$\,s. These 16 episodes are all that is known about the network. For the 3G families, the measurements on their full training splits (Appendix~\ref{app:llm-routing}) were made when $\pi_0$ was designed, so they already exist when the router is deployed and are used as they are. The two kinds of entries are therefore measured on very different amounts of data (8192 episodes against 16), and the evidence makes this visible: Table~D (Appendix~\ref{app:continual-evidence}) shows, for each new network, how consistent the 16 episodes are.

\subsection{Recording and Replaying Episodes}
\label{app:continual-replay}

\paragraph{\textbf{What is recorded.}} An episode is determined by the trace $\mathcal{T}^{(n)} = \{(t_k, c_k)\}_{k=1}^{K}$ on which it is played, $n \in \{1, \dots, N\}$, and by the initial trace time $\tau_1$, the position in the trace at which the first chunk starts; the environment draws both when the episode starts (Appendix~\ref{sec:system-model}). For each of the $N_{\mathrm{c}}$ episodes on network $s$ we record the pair $(n_\ell, \tau_{1,\ell})$, $\ell = 1, \dots, N_{\mathrm{c}}$. The pairs are drawn by the environment and do not depend on the policy that plays the episodes, so the LLM and the gate receive exactly the same 16 episodes; they are drawn from the training split only.

\paragraph{\textbf{Replay.}} Any policy or expert can be played again on a recorded episode by placing the environment on trace $\mathcal{T}^{(n_\ell)}$ at time $\tau_{1,\ell}$; the capacity it meets is then the same as in the recording. During the revision, every measurement on a new network is made by replaying its 16 recorded episodes, which come from its training split: the numbers shown to the LLM, the samples the gate is fitted to, and the acceptance test. The test split is not involved; it is used only after each stage. The download time of a chunk includes a random delay noise, so two replays of the same episode give slightly different QoE. We therefore replay each recorded episode 512 times with different noise and average: the simulator runs 4096 environments in parallel, each plays two episodes, and environment $e$ replays recorded episode $\ell = e \bmod N_{\mathrm{c}}$, which gives $4096 \times 2 / 16 = 512$ replays per episode. The noise is drawn from a seed, and the seed is the only thing that differs between replays of the same episode. Replay assumes that the link capacity during a recorded episode is known. In our trace-driven setting it is, since it is given by the trace; on a real link it would have to be estimated from the throughput log of the episode, which reflects the chunks the deployed policy chose, so the replay of the other experts would only be approximate.

\paragraph{\textbf{Repetitions.}} The 16 episodes recorded on a network are a random draw: for each episode the environment picks a trace of the training split and an initial trace time at random. Another draw would give other episodes and possibly other revisions. To measure how much the results depend on the draw, we repeat the whole procedure three times, each time with a new draw of 16 episodes on every network, and change nothing else. We call the index of the repetition the \emph{collection seed} $c \in \{0, 1, 2\}$; technically, the draw on a network in repetition $c$ uses the random seed $100c$ plus the position of the network in the default order Lumos4G, Lumos5G, SolisWi-Fi, Norway3G (1 to 4; equal to $s$ unless the order is changed, Appendix~\ref{app:continual-order}), so that every network in every repetition gets its own draw, and a network gets the same draw under every order.

\paragraph{\textbf{Seeds.}} Five kinds of random seed appear in this section. The \emph{collection seed} $c \in \{0, 1, 2\}$ selects the 16 recorded episodes of every new network. The \emph{evidence seed} 123 is the seed of the simulation from which Tables~A, B2, C and~D and the samples of the gate are computed. The \emph{validation seed} 0 is the seed of the simulation used by the acceptance test. The \emph{test seed} 123 is the seed of the evaluation on the test splits after each stage; the test splits contain other traces than the training splits, so the evidence seed and the test seed share the value but not the episodes. The \emph{training seed} of the gate sets its initial parameters and the order of its samples.

\subsection{Evidence Given to the Reviser}
\label{app:continual-evidence}

In every round the LLM receives a text prompt with the material for its revision: measurements on every network met so far, i.e.\ the five 3G families and each new network that has arrived, together with the deployed policy and a log of the earlier rounds. We describe the items in turn. Tables~A, B2, C and~D record how the experts and the deployed policy play, and come from one simulation of the training data with random seed 123: on the 3G families it plays the episodes of the training split, on a new network it replays the 16 recorded episodes. The gate receives only part of this material: it is fitted to the expert rollouts of the same simulation, from which Tables~A and~B2 are computed.

\emph{Tables A, B and B2} are those of Appendix~\ref{app:llm-procedure}, extended by one row per new network. On the 3G families they are unchanged. On network $s$, Table~A gives the QoE, rebuffering and bitrate change of each expert playing alone on the replays of the 16 episodes. Table~B2 gives, for each of the four quantities the router computes ($\texttt{tp\_mean}$, $\texttt{tp\_cv}$, $\texttt{buffer}$, $\texttt{buffer\_diff}$), the distribution of its values over all decisions of these replays. Table~B describes the network itself, not how any policy plays, so it needs no simulation and is computed directly from the traces. On a new network only the parts of the traces that the 16 episodes cover are known, so Table~B uses only these: for episode $\ell$, the capacity samples $c_k$ of trace $\mathcal{T}^{(n_\ell)}$ in the $I L = 48 \times 4\,\mathrm{s} = 192\,\mathrm{s}$ after its start $\tau_{1,\ell}$, i.e.\ whose interval ends in $(\tau_{1,\ell}, \tau_{1,\ell} + I L]$; the samples of the 16 episodes are pooled.

\emph{Table C: the deployed policy on every entry.} For each entry, the QoE of the deployed policy, the best expert alone and its QoE, the \emph{regret} (QoE of the policy minus that of the best expert), and the fraction of decisions on which the policy selects each expert. Table~C shows where the deployed policy loses and which expert it uses there. For example, at stage~1 it showed that the one-shot policy obtained $+2.003$ on Network-1 (evidence seed), against $+2.596$ for RobustMPC alone, while selecting Merina on 77\% of the decisions.

\emph{Table D: per-episode winners on the new networks.} On a new network, Table~A ranks the experts by their mean QoE over only 16 episodes. A mean over so few episodes can be dominated by a few of them: an expert that does much better than the others on two or three episodes can be best on average although it is worse on most of the rest. Such a lead is unlikely to hold on other episodes of the network. Table~D shows whether this is the case: for each new network, it gives the number of the 16 episodes on which each expert obtains the highest QoE. If the expert that is best on average is also best on most of the 16 episodes, its lead does not rest on a few of them and is more likely to hold on other episodes; it is not guaranteed to, since the 16 episodes may not be representative of the network. On Network-1, for instance, RobustMPC is best on average ($+2.60$) and on 9 of the 16 episodes. FDASH is best on 4 episodes, yet its mean QoE over the 16 is $-5.07$: it wins a few episodes and loses heavily on the others.

\emph{Revision log.} An entry gives three things: whether the candidate of that round was accepted by the acceptance test (Appendix~\ref{app:continual-revise}) and, if not, why; the notes the LLM wrote with the candidate; and the QoE that the acceptance test measured on each network, both for the candidate and for the policy deployed at that time. For example, the entry of stage~1, round~2 reads ``Network-1 +2.497 (was +2.444)'': the candidate obtained $+2.497$ on Network-1, the deployed policy $+2.444$. From the log the LLM can see which changes have already been tried and what they achieved.

\subsection{Revision and Acceptance}
\label{app:continual-revise}

\paragraph{One round.} Each stage consists of at most $N_{\mathrm{r}} = 5$ rounds. In a round the reviser produces a candidate $\pi'$ from the evidence on the deployed policy $\pi$:
\begin{itemize}
	\item \emph{LLM.} Opus~4.7 writes a complete Form~I policy, which replaces $\pi$ as a whole. The answer is checked with $\textsc{Check}_{\mathrm{I}}$ (Appendix~\ref{app:llm-check}) and queried again with the rejection message if it fails, up to three attempts.
	\item \emph{Gate.} The gate is refitted from scratch, in the default setting, to the samples of all entries: the 3G samples of Appendix~\ref{app:llm-mlp-gate} and, for each new network, the samples of the experts' replays of its 16 episodes at the evidence seed (each new network adds $5 \times 8192 \times 47 = 1{,}925{,}120$ samples, as many as one 3G family). Round $k$ uses training seed $k - 1$. The samples come from the 4096 environments that the simulator runs in parallel. Before the fit, the samples of 409 of these environments (those whose index is $9, 19, 29, \dots$) are set aside; they are not trained on and serve only to choose the epoch: the fit keeps the parameters of the epoch with the lowest loss on them. On a 3G family these 410 environments play other episodes than the remaining ones, so a low loss on them indicates that the gate generalises to unseen episodes. 
\end{itemize}

\paragraph{Acceptance test.} A candidate is not deployed directly; it must first pass an acceptance test. The test rejects two kinds of candidates. The first only appears to be better: its gain is no larger than the noise of the simulation. The second improves the new network but performs worse on networks met earlier, i.e.\ it forgets. The test scores policies in a simulation of the training data with random seed 0, which we call the \emph{validation} seed, rather than with the evidence seed 123. The reason is that the candidate was designed from the numbers of the evidence seed, so it may beat the deployed policy only on the particular episodes and delay noise drawn there; a different seed checks whether the gain holds on other random draws.

Let $\mathrm{QoE}(\pi, u)$ be the mean QoE of $\pi$ on entry $u$ in the validation simulation, and $\mathrm{QoE}^{\mathrm{best}}_u$ the highest $\mathrm{QoE}(\cdot, u)$ of any policy deployed since entry $u$ arrived; for a 3G family this starts at $\mathrm{QoE}(\pi_0, u)$, for network $s$ at the QoE of the policy deployed when it arrived. A candidate $\pi'$ replaces $\pi$ if and only if
\begin{align}
	\frac{1}{|\mathcal{E}_s|} \sum_{u \in \mathcal{E}_s} \mathrm{QoE}(\pi', u) &\ \ge\ \frac{1}{|\mathcal{E}_s|} \sum_{u \in \mathcal{E}_s} \mathrm{QoE}(\pi, u) + \varepsilon, \label{eq:cont-gain}\\
	\mathrm{QoE}(\pi', u) &\ \ge\ \mathrm{QoE}^{\mathrm{best}}_u - \varepsilon \quad \text{for every } u \in \mathcal{E}_s, \label{eq:cont-floor}
\end{align}
with $\varepsilon = 0.01$. Condition~\eqref{eq:cont-gain} requires a gain of at least $\varepsilon$. The evaluation noise of Table~\ref{tab:eval_noise} ($0.026$) measures how the QoE of one policy changes from one set of episodes to another. Both policies here are scored on the same episodes with the same delay noise, so this variation affects them alike and largely cancels in their difference; $\varepsilon$ can therefore be smaller than $0.026$. Condition~\eqref{eq:cont-floor} forbids forgetting: no entry may fall more than $\varepsilon$ below the best it has had, so a gain on a new network cannot be bought with a loss on an old one. Because the gain is averaged over all entries, a candidate that improves only network $s$ must improve it by $(5 + s)\,\varepsilon$, i.e.\ by $0.06$ at stage~1 and $0.09$ at stage~4. When $\pi'$ is accepted, $\mathrm{QoE}^{\mathrm{best}}_u$ is updated on every entry.

\paragraph{A limit of the acceptance test.} The acceptance test cannot tell whether a revision works on a new network as a whole or only on its 16 recorded episodes. There are two reasons. First, the test uses the same 16 episodes from which the revision was derived. Second, the LLM sees the acceptance scores of its earlier candidates in the revision log, so its later candidates can be tuned to these scores; the gate does not see them. Whether a revision works on the whole network is therefore known only from the test split, which contains other episodes of the same network and is evaluated after each stage.

\paragraph{End of a stage.} A stage ends after $N_{\mathrm{r}}$ rounds or after two consecutive rejections, whichever comes first; the policy deployed at that point is $\pi_s$. Algorithm~\ref{alg:cont} summarises the procedure.

\begin{algorithm}[!htbp]
	\caption{Continual revision of the routing policy}
	\label{alg:cont}
	\small
	\SetKwComment{Comment}{$\triangleright$\ }{}
	\SetCommentSty{textit}
	\DontPrintSemicolon
	\KwIn{One-shot policy $\pi_0$; training splits $\mathcal{D}_1, \dots, \mathcal{D}_S$ of the arriving networks; $N_{\mathrm{c}} = 16$, $N_{\mathrm{r}} = 5$, $\varepsilon = 0.01$.}
	\KwOut{Policies $\pi_1, \dots, \pi_S$.}
	$\mathcal{E}_0 \gets$ the five 3G families;\; $\pi \gets \pi_0$;\; $\mathrm{QoE}^{\mathrm{best}}_u \gets \mathrm{QoE}(\pi_0, u)$ for $u \in \mathcal{E}_0$\;
	\For{$s \gets 1$ \KwTo $S$}{
		$\pi$ plays $N_{\mathrm{c}}$ episodes on $\mathcal{D}_s$; record $(n_\ell, \tau_{1,\ell})_{\ell=1}^{N_{\mathrm{c}}}$\;
		$\mathcal{E}_s \gets \mathcal{E}_{s-1} \cup \{\text{Network-}s\}$;\; replay every expert on the recorded episodes (evidence seed)\;
		$\mathrm{QoE}^{\mathrm{best}}_{\text{Network-}s} \gets \mathrm{QoE}(\pi, \text{Network-}s)$;\; $z \gets 0$\;
		\For{$k \gets 1$ \KwTo $N_{\mathrm{r}}$}{
			evidence $\gets$ Tables A, B, B2, C, D on $\mathcal{E}_s$ $\Vert$ $\pi$ $\Vert$ revision log\;
			$\pi' \gets$ \textsc{Revise}(evidence) \Comment*[r]{LLM rewrite or gate refit}
			\eIf{$\pi'$ satisfies \eqref{eq:cont-gain} and \eqref{eq:cont-floor}}{
				$\pi \gets \pi'$;\; $\mathrm{QoE}^{\mathrm{best}}_u \gets \max(\mathrm{QoE}^{\mathrm{best}}_u, \mathrm{QoE}(\pi', u))$ for $u \in \mathcal{E}_s$;\; $z \gets 0$\;
			}{
				$z \gets z + 1$;\; \lIf{$z = 2$}{\textbf{break}}
			}
			append the round to the revision log\;
		}
		$\pi_s \gets \pi$\;
	}
\end{algorithm}

\subsection{Evaluation}
\label{app:continual-eval}

After every stage, $\pi_s$ is evaluated on the test splits of all nine datasets, the five 3G families and the four arriving networks, with the protocol of Appendix~\ref{app:llm-eval} (seed 123, 4096 environments $\times$ 2 episodes, entire test split). No test split is seen by either reviser. On each test split this gives the mean QoE of $\pi_s$, and its \emph{regret}: the QoE of $\pi_s$ minus that of the best expert alone on this test split. The best expert of each test split is fixed in advance (Table~\ref{tab:cont-res}, row ``Best expert per dataset''); a regret of 0 means that $\pi_s$ is as good as this expert there, and a negative regret how much it falls short. From these values we compute three measurements per stage: \emph{\textbf{Forgetting}}: the mean QoE of $\pi_s$ over the five 3G test splits. The 3G families are known from the start, so this value should not fall as new networks arrive. \emph{\textbf{Learning}}: the mean regret over the test splits of the networks that have arrived, i.e.\ networks $1, \dots, s$. It measures how well the policy serves the networks it has met, of which it has seen only 16 episodes each. It has no value at stage~0, when no network has arrived. \emph{\textbf{Forward transfer}}: the mean regret over the test splits of the networks that have not yet arrived, i.e.\ networks $s + 1, \dots, 4$. It measures whether a revision made for the networks met so far also helps networks the policy has never seen. It has no value at stage~4, when all networks have arrived.

The whole procedure is repeated three times, once for each collection seed $c = 0, 1, 2$, i.e.\ with a different set of 16 recorded episodes on every new network. In each of the three repetitions, the acceptance test rejected every candidate of the gate: each candidate lowered at least one 3G family by more than $\varepsilon$ below its best, which violates condition~\eqref{eq:cont-floor}. The deployed gate therefore never changes. To separate the effect of the acceptance test from that of the reviser, we also evaluate the gate \emph{without acceptance}: the candidate refitted at stage $s$ is deployed regardless of the test. Since the refit does not depend on the previously deployed policy, the gate without acceptance at stage $s$ is exactly the stage-$s$ candidate of round~1 (training seed~0).

\paragraph{Offline optimum.} As an upper reference we compute, for every test episode, the best QoE that any bitrate sequence could have obtained if the whole episode had been known in advance. This is possible in our simulator because nothing random in an episode depends on the bitrates chosen: the trace $\mathcal{T}^{(n)}$, the initial trace time $\tau_1$ and the delay noise of every chunk are drawn from the seed, the index of the environment, the index of the episode and, for the noise, the index of the chunk. For each of the $4096 \times 2$ episodes of a test split we therefore know everything the episode will meet, and the result is paired episode by episode with the experts and the policies of Table~\ref{tab:cont-res}. The first chunk is downloaded at the lowest bitrate before the first decision, so the bitrates of the remaining $I - 1 = 47$ chunks are chosen. We choose them by dynamic programming over the chunks. The state after a chunk is the buffer level, rounded down to 100\,ms, and the bitrate of that chunk, i.e.\ at most $601 \times 6$ states, since the buffer never exceeds 60\,s. From every state each of the six bitrates is tried for the next chunk, with the download time, stall and buffer computed exactly as in the simulator; when several paths reach the same state, the one with the highest QoE so far is kept, together with its exact buffer level and trace time. The bitrate sequence of the best final state is then played in the simulator, and the QoE of this playback is reported.

This upper reference is an approximation. What the next chunks meet depends not only on the buffer level and the last bitrate but also on the trace time at which the next download starts, and two paths with the same state can be at different trace times, e.g.\ because one of them stalled longer. Keeping only the path with the higher QoE so far can thus discard a path that would have met higher capacity later and ended with a higher QoE. The result is therefore the QoE of a feasible bitrate sequence, and could be slightly below the true optimum. To check how much this matters, we compared it with each expert episode by episode: in 3 of the $9 \times 8192 = 73{,}728$ test episodes an expert obtained a higher QoE, by at most $0.003$. The offline optimum is not a policy: it uses the future capacity and noise, which no router can observe, so it is not part of the expert pool and only shows how much QoE is left.

\subsection{Results}
\label{app:continual-res}

\begin{figure*}[!htbp]
	\centering
	\includegraphics[width=\textwidth]{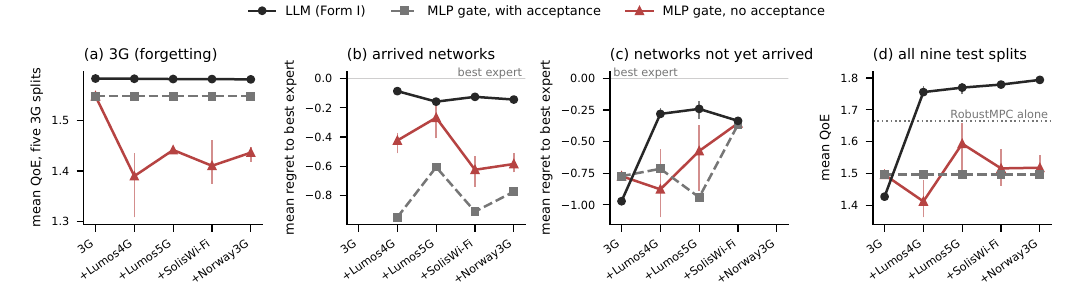}
	\caption{Continual revision over the four stages, on the test splits (seed 123); lines: mean over the three collection seeds, whiskers: their range. (a)~Mean QoE over the five 3G families. (b)~Mean regret over the networks that have arrived. (c)~Mean regret over the networks that have not yet arrived. Panel~(b) has no value at stage~0, when no network has arrived, and (c) none at stage~4, when all four have arrived; Norway3G is part of the average in (c) at stages~0--3 and enters (b) at stage~4. (d)~Mean QoE over all nine datasets; dotted: RobustMPC alone, the best single expert over the nine. Regret is the QoE minus that of the best expert on each dataset. The set of networks averaged in (b) and (c) changes from stage to stage, so a change in these panels can come from the set and not from the policy; for instance, the drop of the LLM in (b) at stage~2 is mostly due to Lumos5G joining the average.}
	\label{fig:cont}
\end{figure*}

The whole procedure was repeated three times, each time with a different set of 16 recorded episodes on every new network; the three repetitions are numbered by the collection seed $c = 0, 1, 2$. Table~\ref{tab:cont-res} and Figure~\ref{fig:cont} show the mean over the three repetitions and, for some values, the range over the repetitions (smallest to largest value).

\begin{table}[!htbp]
	\centering
	\caption{Test QoE (seed 123) after selected stages, mean over the three collection seeds; for all nine datasets also the range over the seeds. 3G: mean over the five families. Gate with acceptance: every candidate was rejected in every repetition, so the values are those of the one-shot gate at every stage. Gate without acceptance: the candidate refitted at the stage, training seed~0. Offline optimum: best bitrate sequence of each test episode with the episode known in advance (Appendix~\ref{app:continual-eval}); it does not depend on the repetition.}
	\label{tab:cont-res}
	\small
	\setlength{\tabcolsep}{3.5pt}
	\begin{tabular}{llcccccc}
		\toprule
		Method & Stage & 3G & Lumos4G & Lumos5G & SolisWi-Fi & Norway3G & All nine \\
		\midrule
		LLM                       & 0 & 1.583 & 2.336 & 1.433 & 1.626 & $-0.465$ & 1.427 \\
		& 1 & 1.582 & 2.680 & 3.760 & 1.916 & $-0.465$ & 1.756 [1.743, 1.775] \\
		& 4 & 1.581 & 2.701 & 3.829 & 2.117 & $-0.405$ & \textbf{1.794} [1.791, 1.798] \\
		Gate, with acceptance     & 0--4 & 1.548 & 1.818 & 3.763 & 0.638 & $-0.495$ & 1.496 \\
		Gate, without acceptance  & 1 & 1.390 & 2.342 & 2.833 & 1.263 & $-0.678$ & 1.412 [1.362, 1.479] \\
		& 4 & 1.436 & 2.134 & 3.815 & 0.942 & $-0.410$ & 1.518 [1.483, 1.557] \\
		\midrule
		Best expert per dataset   & -- & 1.587 & 2.766 & 4.026 & 2.156 & $-0.130$ & 1.861 \\
		RobustMPC alone           & -- & & & & & & 1.664 \\
		Offline optimum & -- & 1.816 & 3.295 & 4.209 & 2.371 & 0.151 & 2.123 \\
		\bottomrule
	\end{tabular}
\end{table}

\paragraph{Starting points.} The two revisers (LLM and gate) start from different one-shot policies, and the gate starts higher: $1.496$ against $1.427$ over the nine test splits (Table~\ref{tab:cont-res}, stage~0). Its advantage lies outside the 3G families, on Lumos5G ($3.763$ against $1.433$), where it selects RobustMPC on 84\% of the decisions and the LLM policy selects Merina; on the 3G families the LLM policy is higher ($1.583$ against $1.548$).

\paragraph{Overall.} In each of the three repetitions the LLM revisions raised the mean QoE over the nine test splits from $1.427$ to between $1.791$ and $1.798$ (mean $1.794$), $0.13$ above RobustMPC alone and $0.067$ below the mean of the best expert of each dataset. The gate stayed at $1.496$ with the acceptance test; without it, it ended at $1.483$ to $1.557$, and its best value at any stage of any repetition was $1.659$, below RobustMPC alone.

\paragraph{Distance to the offline optimum.} The last row of Table~\ref{tab:cont-res} shows how much QoE could be obtained on each test split at all, with every episode known in advance: $2.123$ over all nine. The LLM revisions raised the mean over all nine from $1.427$ at stage~0 to $1.794$ at stage~4, so the distance to the optimum fell from $2.123 - 1.427 = 0.696$ to $2.123 - 1.794 = 0.329$, i.e.\ by about half. The best expert per dataset obtains $1.861$, which is itself $0.262$ below the optimum: even a router that always selected the best expert of each dataset would leave most of the remaining distance. Per dataset, the final LLM policy is closest to the optimum on the 3G families ($1.581$ against $1.816$) and on SolisWi-Fi ($2.117$ against $2.371$), and furthest on Lumos4G ($2.701$ against $3.295$) and Norway3G ($-0.405$ against $0.151$).

\paragraph{No forgetting.} The LLM policy kept its QoE on the 3G families: their mean was $1.583$ at stage~0 and $1.581$ at stage~4 (Table~\ref{tab:cont-res}), and in no repetition did this mean move change by more than $0.004$ from its stage-0 value (Fig.~\ref{fig:cont}a). Condition~\eqref{eq:cont-floor} of the acceptance test contributed to this: it rejected four LLM candidates in total because they lowered an earlier network below its floor.

\paragraph{Forward transfer.} The stage-1 revisions, made from 16 Lumos4G episodes, also improved two networks that had not arrived yet (Table~\ref{tab:cont-res}, LLM, stages~0 and~1): Lumos5G rose from $1.433$ to $3.760$ and SolisWi-Fi from $1.626$ to $1.916$, while Norway3G stayed at $-0.465$. The mean regret over these three networks rose from $-1.15$ at stage 0 to between $-0.31$ and $-0.23$ at stage 1, depending on the repetition (Fig.~\ref{fig:cont}c). As a consequence, little was left to gain when Lumos5G and SolisWi-Fi arrived. 

\paragraph{The gate forgets.} Each of the 24 gate candidates raised the network that had just arrived but lowered at least one 3G family below its floor, so none was accepted. The strict test is not why the gate falls behind: deployed without it, the gate at stage~4 is worse than the LLM on the 3G families ($1.436$ against $1.581$), on Lumos4G ($2.134$ against $2.701$) and on SolisWi-Fi ($0.942$ against $2.117$), and about equal on Lumos5G and Norway3G (Table~\ref{tab:cont-res}).

\newpage
\subsection{One Repetition Round by Round}
\label{app:continual-run}

\begin{figure*}[!htbp]
	\centering
	\includegraphics[width=\textwidth]{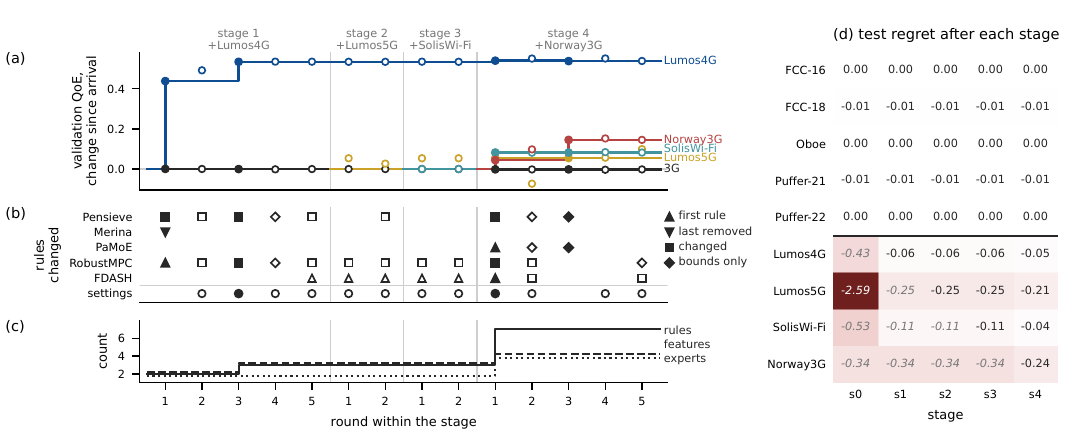}
	\caption{The 14 rounds of the LLM in the repetition with collection seed $c = 0$. (a)~QoE of the deployed policy in the acceptance test (validation seed) after each round, as the change since the network arrived; 3G is the mean over the five families. Filled markers: accepted candidates; hollow markers: the scores of rejected candidates. (b)~The experts whose rules a candidate changed, compared with the deployed policy. Upward triangle: the expert receives its first rule; downward triangle: its last rule is removed; square: it has rules both before and after, but they differ, because one of its rules has other conditions, or it gains a rule, or it loses a rule but keeps at least one; diamond: its rules are unchanged, but an interval used in one of them has other bounds. Row ``settings'': warm-up, stickiness or the fence changed. Filled: accepted, hollow: rejected. (c)~Size of the deployed policy: number of rules, of features and of experts selected by some rule. (d)~Regret on the test splits (seed 123) after each stage.}
	\label{fig:cont-run}
\end{figure*}

Figure~\ref{fig:cont-run} follows the LLM through the repetition with $c = 0$: 14 rounds, of which four were accepted. Every answer is first checked by $\textsc{Check}_{\mathrm{I}}$, which tests only whether it is a valid policy, before its candidate enters the acceptance test. In 13 of the 14 rounds the first answer passed this check; in round~1 of stage~3 the first answer lacked the stickiness settings, and the second answer, sent after the rejection message, passed, so every round produced a valid candidate. In this subsection, a condition such as ``throughput above 12--18\,Mbps'' denotes an interval whose membership rises from 0 at 12\,Mbps to 1 at 18\,Mbps.

\paragraph{Stage 1: Lumos4G} The one-shot policy sent high throughput to Merina, which on Lumos4G is $0.67$ below RobustMPC (Table~A). The first candidate gave this range to RobustMPC instead (panel~b, round~1: Merina loses its rule, RobustMPC receives its first, the rule of Pensieve is adjusted) and raised Lumos4G by $0.44$ while the 3G families stayed flat (panel~a). The second accepted candidate (round~3) let RobustMPC also take over at lower throughput and keep control through short dips (two filled squares), and shortened warm-up and stickiness (filled circle); Lumos4G rose by another $0.10$.

\paragraph{Stages 2 and 3: no revision.} In every round of these two stages the LLM tried to give FDASH its first rule, for the very high throughput of Lumos5G, on which FDASH is the best expert (hollow upward triangles in the FDASH row). These candidates raised Lumos5G by at most $0.054$ and left the other networks unchanged (panel~a). Condition~\eqref{eq:cont-gain} averages the gain over all entries, so a gain on one network must reach $0.07$ at stage~2 and $0.08$ at stage~3; all four candidates were rejected. Both networks arrived with small regret, because the stage-1 revisions had already improved them (panel~d: Lumos5G from $-2.59$ to $-0.25$, SolisWi-Fi from $-0.53$ to $-0.11$).

\paragraph{Stage 4: Norway3G} At stages~2 and~3 the FDASH rule alone was rejected: it raised only Lumos5G, by at most $0.054$, whereas at stage~4 a gain on a single network must reach $0.09$. In round~1 the LLM therefore put three changes into one candidate, each aimed at a different network: (i)~the FDASH rule, for Lumos5G; (ii)~a lower throughput threshold for RobustMPC, for SolisWi-Fi; (iii)~a first rule for PaMoE at low buffer and low throughput, for Norway3G, on which Pensieve stalls more than PaMoE (Table~A). With the gains on three networks combined, the candidate was accepted: all three networks rose (panel~a), and the policy grew from three to seven rules and from two to four experts (panel~c). Round~2 was rejected by condition~\eqref{eq:cont-floor}: it raised Norway3G but lowered Lumos5G far below its best (panel~a, the hollow marker below zero). Round~3 changed no rule, only the bounds of intervals used by the rules of Pensieve and PaMoE (filled diamonds), which changed the buffer and throughput levels at which PaMoE takes over; it raised Norway3G to $+0.145$ since its arrival.

\paragraph{Outcome.} On the test splits (panel~d), the regret on Lumos4G went from $-0.43$ at stage~0 to $-0.06$ at stage~4 and on Norway3G from $-0.34$ to $-0.24$. The final policy selects Pensieve, PaMoE, RobustMPC and FDASH; Merina, which the one-shot policy selected at high throughput, is no longer selected.

\begin{figure*}[!htbp]
	\centering
	\includegraphics[width=\textwidth]{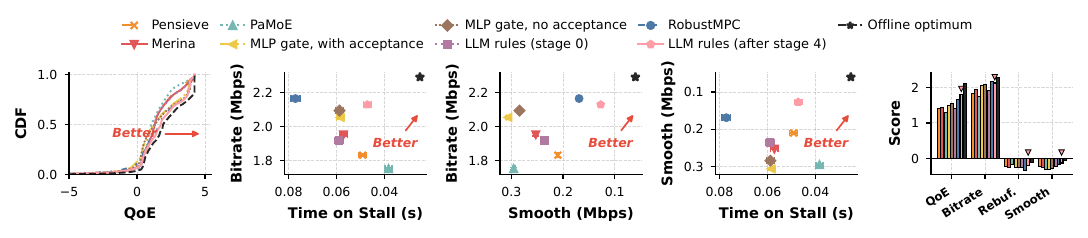}
	\caption{The policies at the end of the repetition with $c = 0$ on the nine test splits (seed 123), all episode pooled. (a)~Distribution of episode QoE. (b)--(d)~Mean bitrate, stall time per chunk and bitrate change (``Smooth'') per chunk, with 95\% confidence intervals; better is up and to the right. (e)~Mean QoE and its three terms.}
	\label{fig:cont-views}
\end{figure*}

\paragraph{Where the gain comes from.} Figure~\ref{fig:cont-views} pools the test episodes of all nine splits for the policies of this repetition after stage~4. FDASH is left out, as its stall time on the new networks lies far outside the axes, and the offline optimum has no confidence intervals, as only its means are stored. The LLM has the highest mean QoE of the policies ($1.798$, against $1.664$ for RobustMPC, the best expert). Its bitrate is close to that of RobustMPC, but it stalls less and changes the bitrate less (panels~b--d). The remaining gap to the offline optimum lies in all three terms (panel~e).

\newpage
\subsection{Choice of LLM}
\label{app:continual-models}

To see how much the revisions depend on the LLM, we repeated the procedure with two smaller models of the same family, Sonnet and Haiku, in place of Opus~4.7. Nothing else was changed: all three models start from the same one-shot policy of Opus~4.7 and face the same acceptance test, and each model is run with the same three collection seeds $c = 0, 1, 2$, i.e.\ on the same recorded episodes as Opus~4.7 in Appendix~\ref{app:continual-res}. Figure~\ref{fig:cont-models} shows the four measurements of Appendix~\ref{app:continual-eval} after each stage. Table~\ref{tab:cont-models} show the test QoE after stage~4.

\begin{figure*}[!htbp]
	\centering
	\includegraphics[width=\textwidth]{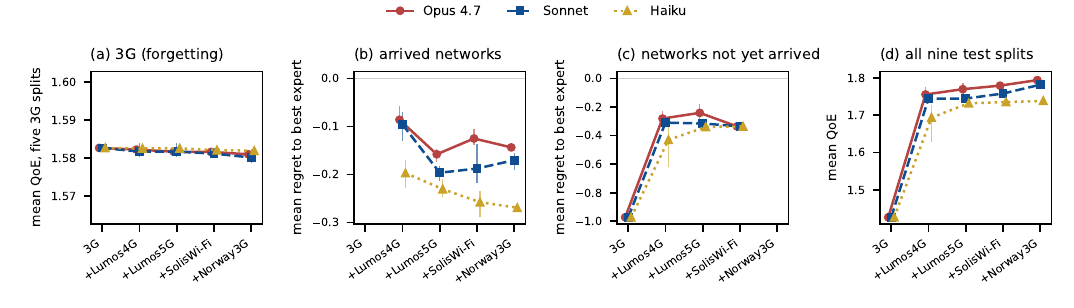}
	\caption{Choice of LLM, on the test splits (seed 123); all three models start from the one-shot policy of Opus~4.7. Lines: mean over the three collection seeds, whiskers: their range. Panels as in Fig.~\ref{fig:cont}: (a)~mean QoE over the five 3G families, on an axis of $\pm 0.02$ around the stage-0 value; (b)~mean regret over the networks that have arrived; (c)~mean regret over the networks that have not yet arrived; (d)~mean QoE over all nine datasets.}
	\label{fig:cont-models}
\end{figure*}

\begin{table}[!htbp]
	\centering
	\caption{Choice of LLM: test QoE (seed 123) after stage~4, mean over the three collection seeds; for all nine datasets also the range over the seeds. Accepted: accepted candidates and rounds, summed over the three seeds. 3G: mean over the five families. The row of Opus~4.7 is that of the LLM at stage~4 in Table~\ref{tab:cont-res}.}
	\label{tab:cont-models}
	\small
	\setlength{\tabcolsep}{3.0pt}
	\begin{tabular}{lccccccc}
		\toprule
		Model & 3G & Lumos4G & Lumos5G & SolisWi-Fi & Norway3G & All nine & Accepted \\
		\midrule
		Opus~4.7 & 1.581 & 2.701 & 3.829 & 2.117 & $-0.405$ & \textbf{1.794} [1.791, 1.798] & 12 of 40 \\
		Sonnet   & 1.580 & 2.687 & 3.795 & 2.010 & $-0.356$ & 1.782 [1.772, 1.789] & 13 of 41 \\
		Haiku    & 1.582 & 2.620 & 3.752 & 1.826 & $-0.459$ & 1.739 [1.736, 1.743] & 7 of 31 \\
		\midrule
		Best expert per dataset & 1.587 & 2.766 & 4.026 & 2.156 & $-0.130$ & 1.861 & -- \\
		\bottomrule
	\end{tabular}
\end{table}

\paragraph{No model forgets.} With every model and every seed, the mean over the 3G families stays within $0.005$ of its stage-0 value at every stage (Fig.~\ref{fig:cont-models}a). Condition~\eqref{eq:cont-floor} does not depend on the model: a candidate that lowers an earlier network below its floor is rejected, whichever model wrote it.

\paragraph{Smaller models gain less on the new networks.} After stage~4 the mean over all nine test splits is $1.794$ for Opus~4.7, $1.782$ for Sonnet and $1.739$ for Haiku (Table~\ref{tab:cont-models}). The 3G families are almost equal, so the differences lie on the four new networks. Sonnet is close to Opus~4.7 on Lumos4G and Lumos5G, higher on Norway3G and lower on SolisWi-Fi ($2.010$ against $2.117$). Haiku is the lowest on all four. Its stage-1 revisions gained less than those of the other two models (Fig.~\ref{fig:cont-models}d, $1.694$ against $1.756$ for Opus~4.7), and later stages added little: from stage~2 to stage~4 its mean over all nine rose by $0.007$, that of Opus~4.7 by $0.024$. Haiku also had fewer candidates accepted. On Norway3G, for example, its revisions raised the QoE only from $-0.465$, the value of the one-shot policy, to $-0.459$, whereas those of Opus~4.7 raised it to $-0.405$ and those of Sonnet to $-0.356$.

\newpage
\subsection{Arrival Order}
\label{app:continual-order}

\begin{figure*}[!htbp]
	\centering
	\includegraphics[width=\textwidth]{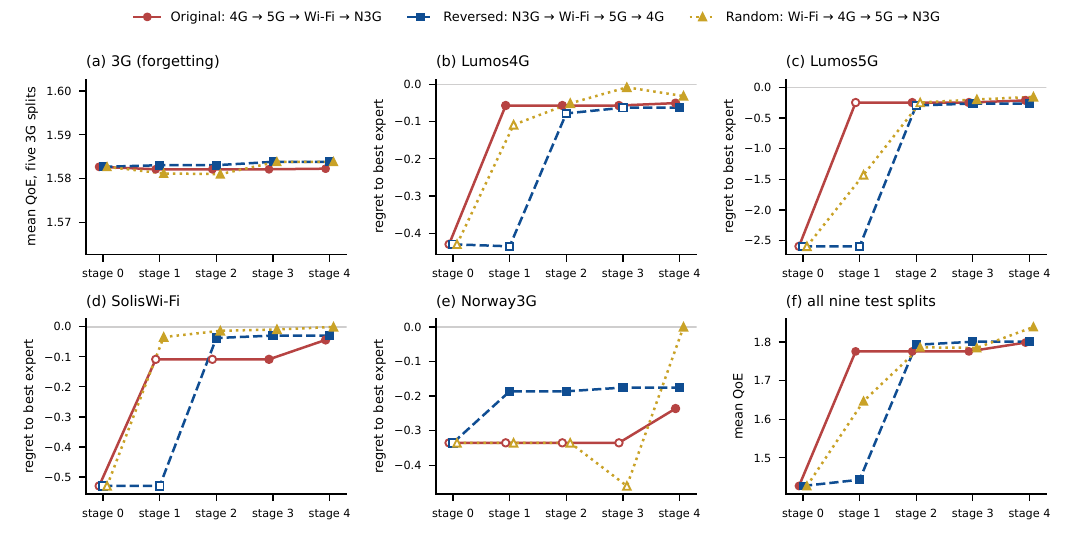}
	\caption{Arrival order, Opus~4.7, repetition with $c = 0$, on the test splits (seed 123). Legend: the order in which the networks arrive at stages~1--4 (4G: Lumos4G, 5G: Lumos5G, Wi-Fi: SolisWi-Fi, N3G: Norway3G). (a)~Mean QoE over the five 3G families, on an axis of $\pm 0.02$ around the stage-0 value. (b)--(e)~Regret on each new network; hollow markers: the network has not yet arrived under that order, filled: it has. (f)~Mean QoE over all nine datasets.}
	\label{fig:cont-order}
\end{figure*}

All runs above meet the new networks in the same order. To see how much the result depends on it, we repeated the run of Opus~4.7 with $c = 0$ under two other orders: the reversed order, and an order drawn at random among the remaining permutations (Fig.~\ref{fig:cont-order}, legend). Nothing else was changed; in particular, the recorded episodes of each network depend only on the network and $c$, not on its position, so every network is represented by the same 16 episodes under all three orders. The averages over arrived and not-yet-arrived networks (Fig.~\ref{fig:cont}b,c) would average different networks under different orders, so Figure~\ref{fig:cont-order} shows the regret on each new network separately. Table~\ref{tab:cont-order} gives the test QoE after stage~4, when all four networks have arrived under every order.

\begin{table}[!htbp]
	\centering
	\caption{Arrival order, Opus~4.7, repetition with $c = 0$: test QoE (seed 123) after stage~4, and the number of accepted candidates and of rounds over the four stages. 3G: mean over the five families.}
	\label{tab:cont-order}
	\small
	\setlength{\tabcolsep}{3.5pt}
	\begin{tabular}{lccccccc}
		\toprule
		Order & 3G & Lumos4G & Lumos5G & SolisWi-Fi & Norway3G & All nine & Accepted \\
		\midrule
		Original & 1.582 & 2.716 & 3.813 & 2.112 & $-0.366$ & 1.798 & 4 of 14 \\
		Reversed & 1.584 & 2.703 & 3.762 & 2.126 & $-0.305$ & 1.801 & 3 of 11 \\
		Random   & 1.584 & 2.735 & 3.870 & 2.155 & $-0.130$ & \textbf{1.839} & 6 of 15 \\
		\midrule
		Best expert per dataset & 1.587 & 2.766 & 4.026 & 2.156 & $-0.130$ & 1.861 & -- \\
		\bottomrule
	\end{tabular}
\end{table}

Under every order the mean over the 3G families stays between $1.581$ and $1.584$ at every stage (Fig.~\ref{fig:cont-order}a). On Lumos4G, Lumos5G and SolisWi-Fi the three orders end within $0.11$ of each other (Table~\ref{tab:cont-order}). Whichever of Lumos4G and SolisWi-Fi arrives first, its revision also raises the other two networks before they arrive (Fig.~\ref{fig:cont-order}b--d): under the reversed order, the stage-2 revision for SolisWi-Fi raised Lumos4G from $2.330$ to $2.688$. The mean over all nine test splits is $1.798$, $1.801$ and $1.839$, and most of this difference lies on Norway3G ($-0.366$, $-0.305$ and $-0.130$, the last equal to its best expert). Under the random order, Norway3G also shows that a revision can lower a network that has not yet arrived: the stage-3 revision for Lumos5G lowered it (Fig.~\ref{fig:cont-order}e), which the acceptance test cannot detect, as it replays only the networks that have arrived. The loss was undone at stage~4, when Norway3G arrived. As repeated runs of the LLM end within $0.007$ of each other (Table~\ref{tab:cont-res}). Under every order, the revisions raised the mean over all nine from $1.427$ to between $1.798$ and $1.839$: the three orders end within $0.041$ of each other, which is small compared with the gain of at least $0.371$.

\newpage

\subsection{Policy Form}
\label{app:continual-forms}

All runs above revise Form~I policies. To see whether a richer form helps, we repeated the procedure of Opus~4.7 in Form~II and in Form~III (Appendix~\ref{app:llm-forms}), with the same three collection seeds $c = 0, 1, 2$. In each form the run starts from the one-shot policy of Opus~4.7 in that form, the LLM writes the revised policy in that form, and the answer is checked with the check of that form (Appendix~\ref{app:llm-check}). The three one-shot policies differ slightly, so Table~\ref{tab:cont-forms} also gives the mean over all nine test splits at stage~0.

\begin{figure*}[!htbp]
	\centering
	\includegraphics[width=\textwidth]{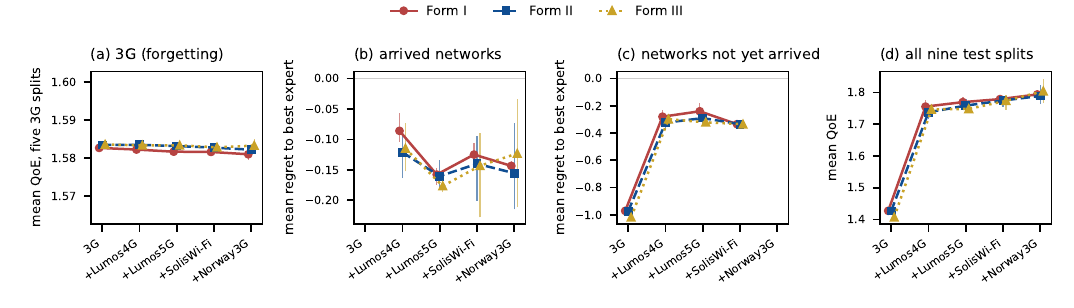}
	\caption{Policy form, Opus~4.7, on the test splits (seed 123); each form starts from the one-shot policy of Opus~4.7 in that form. Lines: mean over the three collection seeds, whiskers: their range. Panels as in Fig.~\ref{fig:cont-models}.}
	\label{fig:cont-forms}
\end{figure*}

\begin{table*}[!htbp]
	\centering
	\caption{Policy form, Opus~4.7: test QoE (seed 123) after stage~4, mean over the three collection seeds; for all nine datasets also the range over the seeds. Stage~0: mean over all nine datasets of the one-shot policy. Accepted: accepted candidates and rounds, summed over the three seeds. 3G: mean over the five families. The row of Form~I is that of Opus~4.7 in Table~\ref{tab:cont-models}.}
	\label{tab:cont-forms}
	\small
	\setlength{\tabcolsep}{4pt}
	\begin{tabular}{lcccccccc}
		\toprule
		Form & Stage~0 & 3G & Lumos4G & Lumos5G & SolisWi-Fi & Norway3G & All nine & Accepted \\
		\midrule
		Form~I   & 1.427 & 1.581 & 2.701 & 3.829 & 2.117 & $-0.405$ & 1.794 [1.791, 1.798] & 12 of 40 \\
		Form~II  & 1.427 & 1.582 & 2.682 & 3.813 & 2.037 & $-0.339$ & 1.789 [1.764, 1.825] & 13 of 41 \\
		Form~III & 1.407 & 1.583 & 2.703 & 3.838 & 2.036 & $-0.253$ & \textbf{1.804} [1.766, 1.844] & 15 of 43 \\
		\midrule
		Best expert per dataset & -- & 1.587 & 2.766 & 4.026 & 2.156 & $-0.130$ & 1.861 & -- \\
		\bottomrule
	\end{tabular}
\end{table*}

With every form and every seed, the mean over the 3G families stays within $0.004$ of its stage-0 value at every stage (Fig.~\ref{fig:cont-forms}a). After stage~4 the mean over all nine test splits is $1.794$ in Form~I, $1.789$ in Form~II and $1.804$ in Form~III (Table~\ref{tab:cont-forms}). These means differ by at most $0.015$, less than the range over the seeds in Form~II ($0.061$) and Form~III ($0.078$), so three seeds cannot tell the forms apart; in Form~I the range is $0.007$. Form~III starts lowest and gains most ($0.397$, against $0.367$ and $0.362$). The richer forms end higher on Norway3G ($-0.339$ and $-0.253$, against $-0.405$) and lower on SolisWi-Fi ($2.037$ and $2.036$, against $2.117$). Of the 125 answers of all nine runs, one failed the check, in Form~I; none of the 43 answers in Form~III did. No round was rejected for lack of a valid answer: every rejection came from condition~\eqref{eq:cont-gain} or~\eqref{eq:cont-floor}.

\newpage
\subsection{Revisers without an LLM}
\label{app:continual-baselines}

The gate is the only reviser in Appendix~\ref{app:continual-res} that does not use an LLM, and it differs from the LLM in two respects at once: its policy is a neural network, and it is refitted to all entries at every stage. To see what the LLM itself contributes, we replaced it by three revisers (Network Lookup, Value Tree, Evolutionary Search) that do not use an LLM, each of which removes a different part of what the LLM does. Everything else was unchanged: the recorded episodes, the acceptance test, the number of rounds and the three collection seeds $c = 0, 1, 2$. None of the three receives more data than the LLM: they use the expert rollouts of the evidence seed, i.e.\ the samples of the gate (Appendix~\ref{app:continual-revise}), from which Tables~A and~B2 are computed.

\paragraph{Episode features.} The network lookup routes an episode by recognising the network it is played on, and the value tree splits the samples by the network conditions under which they were collected. Both therefore need features that describe the network. Table~B describes a network by two properties, the level of its capacity and its variability, which is how the 3G families differ most. The router does not observe the capacity, only the throughput of the chunks it downloads, so we measure the two properties on the throughput: its mean and its coefficient of variation over all chunks downloaded so far in the episode. The features \texttt{tp\_mean} and \texttt{tp\_cv} measure the same two quantities, but over the last five chunks only, so they follow short-term changes within an episode, and an episode would move from one entry of the lookup to another as its throughput fluctuates. Over all chunks downloaded so far, the two values settle as the episode proceeds and describe the network rather than its current state. We chose these two features for the two revisers. The LLM is not given them. A Form~I policy declares its own features from a fixed list of operators (Appendix~\ref{app:llm-form1}), so the LLM could declare the same two features, but none of its policies does. The one-shot policy uses only the throughput mean over the last five chunks. In the repetitions with $c = 0$ and $c = 2$, the policy accepted at stage~1 adds a throughput mean over the last 20 chunks.

\begin{figure*}[!htbp]
	\centering
	\includegraphics[width=\textwidth]{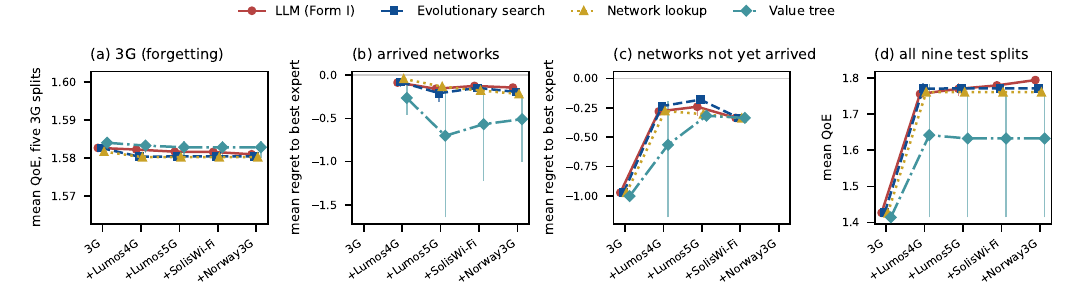}
	\caption{Revisers without an LLM, on the test splits (seed 123). Lines: mean over the three collection seeds, whiskers: their range. Panels as in Fig.~\ref{fig:cont-models}.}
	\label{fig:cont-baselines}
\end{figure*}

\paragraph{Network Lookup.} A table with one entry per entry of $\mathcal{E}_s$. An entry holds a centroid, the mean of the two episode features over the expert rollouts on that entry (the mean throughput on a log scale, both standardised over all entries), and the expert with the highest QoE on that entry (Table~A). At every chunk the router plays the expert of the entry whose centroid is nearest to the episode features. When network $s$ arrives the table gains one entry and keeps the others. The lookup tests whether recognising the network is enough: it can select only the expert that is best on a whole entry.

 \paragraph{Value tree.} The policy is a decision tree of depth at most four. Each internal node compares one of \texttt{tp\_mean}, \texttt{tp\_cv}, \texttt{ep\_mean} and \texttt{ep\_cv} with a threshold, and each leaf names an expert. The tree is fitted to the rollouts of all entries: each split is chosen so that, on each side, the expert with the highest mean QoE in the rollouts is as good as possible, and each leaf names that expert. The tree uses no buffer feature, because the buffer depends on which expert played. Like the gate, the tree is refitted from scratch at every stage. It tests whether a readable policy fitted to the rollouts suffices.

\paragraph{Evolutionary search.} The search keeps everything of the LLM reviser except the LLM: it starts from the same one-shot policy, writes Form~I policies, and checks them with $\textsc{Check}_{\mathrm{I}}$. In every round it changes the deployed policy at random, e.g.\ it moves a threshold, changes the expert of a rule, or adds or removes a rule or a feature, and keeps the changed policies that score best, for eight generations of eight policies each. A policy scores its mean gain on the entries at the evidence seed, with a penalty if it falls more than $\varepsilon$ below the deployed policy on any entry. The best policy found is submitted to the acceptance test. The search reads no table; instead it plays up to 64 policies per round on every entry, whereas the LLM plays none.

\begin{table*}[!htbp]
	\centering
	\caption{Revisers without an LLM: test QoE (seed 123) after stage~4, mean over the three collection seeds; for all nine datasets also the range over the seeds. Stage~0 and Stage~1: mean over all nine datasets after these stages. Accepted: accepted candidates and rounds, summed over the three seeds. 3G: mean over the five families. The row of the LLM is that of Form~I in Table~\ref{tab:cont-forms}.}
	\label{tab:cont-baselines}
	\small
	\setlength{\tabcolsep}{3.5pt}
	\begin{tabular}{lccccccccc}
		\toprule
		Reviser & Stage~0 & Stage~1 & 3G & Lumos4G & Lumos5G & SolisWi-Fi & Norway3G & All nine & Accepted \\
		\midrule
		LLM (Form~I)        & 1.427 & 1.756 & 1.581 & 2.701 & 3.829 & 2.117 & $-0.405$ & \textbf{1.794} [1.791, 1.798] & 12 of 40 \\
		Evolutionary search & 1.427 & 1.770 & 1.580 & 2.561 & 3.812 & 2.131 & $-0.465$ & 1.771 [1.750, 1.783] & 7 of 32 \\
		Network lookup      & 1.426 & 1.761 & 1.580 & 2.722 & 3.803 & 1.887 & $-0.468$ & 1.761 [1.759, 1.762] & 3 of 27 \\
		Value tree          & 1.415 & 1.642 & 1.583 & 2.463 & 2.924 & 1.857 & $-0.465$ & 1.633 [1.415, 1.754] & 3 of 27 \\
		\midrule
		Best expert per dataset & -- & -- & 1.587 & 2.766 & 4.026 & 2.156 & $-0.130$ & 1.861 & -- \\
		\bottomrule
	\end{tabular}
\end{table*}

\paragraph{Overall.} After stage~4 the mean over all nine test splits is $1.794$ for the LLM, $1.771$ for the evolutionary search, $1.761$ for the lookup and $1.633$ for the value tree (Table~\ref{tab:cont-baselines}); the lowest of the three LLM runs is higher than the highest run of any other reviser. The differences lie on the new networks, where each reviser stops improving for a different reason.

\paragraph{Evolutionary search stops after stage~2.} After stage~1 the search is ahead of the LLM ($1.770$ against $1.756$). In stages~3 and~4, however, none of its candidates was accepted in any repetition: the largest mean gain was $0.007$, below $\varepsilon$, and Norway3G stayed at $-0.465$. The LLM had a candidate accepted at stage~4 in every repetition and raised Norway3G to $-0.405$. The search can also fit the 16 recorded episodes rather than the network.

\paragraph{The lookup stops after stage~1.} At stage~1 the lookup added a row for Network-1, which plays RobustMPC, and this candidate was accepted in every repetition. At stage~2 it added a row for Network-2, which plays FDASH, the best expert on Lumos5G. Some episodes of Network-1 are closer to this new row than to their own, so they are now played by FDASH, the worst expert on Network-1 (Table~A). Network-1 therefore fell below its floor, e.g.\ from $2.564$ to $2.104$ with $c = 0$, and the candidate was rejected. The same happened to every later candidate.

\paragraph{The value tree lowers the 3G families.} At every stage the tree is refitted from scratch to the rollouts of all entries, so each new tree also routes the 3G families differently.  Each refit also changes how the 3G families are routed, and almost every candidate was rejected because it lowered one of them below its floor. With $c = 1$ no candidate was accepted at all, so the tree never improved on its stage-0 value.

\paragraph{Summary.} Each of the three revisers stops improving for its own reason: the value tree lowers earlier networks, the lookup confuses overlapping networks, and the evolutionary search finds no change that passes the acceptance test once the remaining gains are small. Only the LLM kept improving the policy until the last stage, although its final lead is small.

\newpage
\section{Validation in a Real Player}
\label{app:realworld}

All results so far come from the simulator. This appendix tests whether they hold when the rules control a real player that downloads real video segments over HTTP, keeps its own buffer and reports its own stalls. Section~\ref{sec:eval-real} summarises the outcome.

\subsection{Test Bed}
\label{app:realworld-testbed}

\paragraph{\textbf{Player and video.}} The player is hls.js~1.7.4 in headless Chromium, unmodified except for how the bitrate is chosen. The video is a clip of 48 segments of 4\,s, encoded with H.264 at six levels from 300 to 4300\,kbps with one key frame per segment. Its mean segment size at every level is within 1.4\% of that of the Envivio video of the simulator. The player buffers at most 30\,s of video, the default of hls.js, while the simulator buffers up to 60\,s.

\paragraph{\textbf{Link.}} Every request of the player passes through a local proxy that limits the download rate to the bandwidth of a trace in real time. The proxy reads the current bandwidth of the trace at least every 50\,ms during a download, so the rate follows the trace while a segment is loading, as in the simulator. While the trace is at zero, nothing is sent until it resumes. The trace starts when the player is opened and repeats if the playback outlasts it. The link is emulated on one machine, so there is no additional latency or packet loss.

\paragraph{\textbf{Policy.}} A policy is plugged into hls.js as its ABR controller, the interface through which a custom ABR algorithm is added to hls.js in a deployment. When hls.js chooses the level of the next segment, it asks the controller and receives the latest decision of the policy. By default, hls.js aborts a download that is too slow and switches to a lower level. We turn this off, so that every level is chosen by the policy. The policy itself runs unchanged in a separate process. After every downloaded segment, it receives the same observation as in the simulator. The download time and size of the segment are measured at the proxy, the buffer level is reported by the player, and the sizes of the next segment come from the segment list. QoE has the form and coefficients of the simulator. 

\subsection{Policies and Traces}
\label{app:realworld-setup}

We compare four policies. \method{} uses the final rules of the repetition with collection seed $c = 0$ (Appendix~\ref{app:continual-run}). The one-shot rules are those deployed before the first arrival. RobustMPC is the best single expert, and Pensieve is the best expert on the 3G datasets. Every policy plays three traces of the test split of each of the nine datasets. The traces are the ones at the 31st, 55th and 93rd percentiles of the mean bandwidth of the split, and playback starts at the beginning of the trace. Four policies, nine datasets and three traces per dataset give 108 episodes, each played once in real time. The video is the same on all nine datasets, unlike in the simulation of Table~\ref{tab:main}, where the four arriving networks use Big Buck Bunny. The values in this appendix can therefore differ from those of Table~\ref{tab:main} and are compared only with each other.

\begin{table*}[!htbp]
	\centering
	\caption{Mean QoE in the real player and in the simulator, three traces per dataset. Sim plays the same trace from its beginning with the segment sizes of the real video, averaged over 20 seeds of the delay noise. 3G is the mean over the five 3G datasets, and All the mean over all 27 traces.}
	\label{tab:realworld}
	\small
	\setlength{\tabcolsep}{4pt}
	\begin{tabular}{l*{6}{cc}}
		\toprule
		& \multicolumn{2}{c}{3G} & \multicolumn{2}{c}{Lumos4G} & \multicolumn{2}{c}{Lumos5G} & \multicolumn{2}{c}{SolisWi-Fi} & \multicolumn{2}{c}{Norway3G} & \multicolumn{2}{c}{All} \\
		\cmidrule(lr){2-3}\cmidrule(lr){4-5}\cmidrule(lr){6-7}\cmidrule(lr){8-9}\cmidrule(lr){10-11}\cmidrule(lr){12-13}
		Policy & Real & Sim & Real & Sim & Real & Sim & Real & Sim & Real & Sim & Real & Sim \\
		\midrule
		\method{}  & \textbf{1.886} & 1.949 & 4.033 & 4.087 & 3.948 & 3.923 & 4.074 & 4.155 & 1.186 & 1.307 & \textbf{2.519} & 2.580 \\
		RobustMPC  & 1.471 & 1.767 & \textbf{4.131} & 4.215 & \textbf{4.062} & 4.172 & \textbf{4.103} & 4.215 & 1.129 & 1.023 & 2.309 & 2.496 \\
		One-shot rules & 1.852 & 1.939 & 3.930 & 3.951 & 2.342 & 1.414 & 3.698 & 4.134 & 1.148 & 1.314 & 2.264 & 2.279 \\
		Pensieve   & 1.806 & 1.939 & 2.182 & 2.605 & 1.039 & 0.954 & 3.119 & 3.577 & \textbf{1.338} & 1.314 & 1.856 & 2.016 \\
		\bottomrule
	\end{tabular}
\end{table*}

\begin{table}[!htbp]
	\centering
	\caption{QoE terms in the real player, mean over all 27 traces. The last two columns compare \method{} with each policy on the same traces, as the mean difference with its standard error and the number of traces on which \method{} is higher.}
	\label{tab:realworld-terms}
	\footnotesize
	\setlength{\tabcolsep}{2.5pt}
	\begin{tabular}{@{}lrrrrrrr@{}}
		\toprule
		Policy & QoE & Bitrate & \shortstack{Stall\\pen.} & \shortstack{Switch\\pen.} & \shortstack{Stall\\(s)} & \shortstack{\method{}\\minus} & \shortstack{\method{}\\higher} \\
		\midrule
		\method{}      & \textbf{2.519} & \textbf{2.743} & \textbf{0.120} & \textbf{0.104} & \textbf{0.74} & -- & -- \\
		RobustMPC      & 2.309 & 2.728 & 0.234 & 0.185 & 1.95 & 0.210 $\pm$ 0.102 & 15 of 27 \\
		One-shot rules & 2.264 & 2.552 & 0.149 & 0.139 & 1.06 & 0.255 $\pm$ 0.099 & 18 of 27 \\
		Pensieve       & 1.856 & 2.134 & 0.155 & 0.122 & 1.09 & 0.663 $\pm$ 0.216 & 20 of 27 \\
		\bottomrule
	\end{tabular}
\end{table}

\paragraph{\textbf{Comparison of the policies.}} Table~\ref{tab:realworld-terms} compares \method{} with each other policy on the same 27 traces. Its advantage over RobustMPC comes from the 3G datasets (Table~\ref{tab:realworld}). On Lumos4G, Lumos5G and SolisWi-Fi, RobustMPC plays the highest level for 96.5\% of the segments and is slightly higher. Over all traces the two policies play about the same bitrate, but \method{} stalls less and switches less. The revisions also hold in the player. On Lumos5G the one-shot rules obtain 2.342 and \method{} 3.948, and on SolisWi-Fi 3.698 and 4.074. On the 3G datasets, served from the start, \method{} is not lower than the one-shot rules, 1.886 against 1.852.

\newpage

\section{The Stage-1 Prompt}
\label{app:prompt}

The prompt of the one-shot generation of Opus~4.7 in Form~I (Appendix~\ref{app:llm-res-gen}), verbatim, and its answer, which is the policy deployed at stage~0 of every continual run (Appendix~\ref{app:continual-res}).
The system prompt states the setting, the decision procedure, the operators and the answer format.
The user prompt gives only measurements: the expert profiles (Table~A), the trace statistics (Table~B) and the observed ranges of candidate quantities (Table~B2).
The answer passed $\textsc{Check}_{\mathrm{I}}$ at the first attempt.

\subsection{System prompt}
\begin{lstlisting}
You are designing, from scratch, the controller of a router that picks, once per video chunk, which of several frozen pre-trained adaptive-bitrate (ABR) experts acts. The router only delegates: the quality played is the chosen expert's own decision for this chunk; the router never chooses a bitrate itself. Every expert sees every chunk, so an expert that is not chosen still keeps its state up to date, and it is told the quality actually played.

You are given ONLY measurements: how each expert scores on each training trace family, the bandwidth statistics of those families, and the observed ranges of quantities the router can compute. You are NOT given any routing heuristic. Infer what to observe and how to route from the measurements.

At run time the router does not know which trace family it is on; it only sees what it observes. A regime is only actionable if the observations can tell it apart from the others.

== How the deployed router runs your controller, once per chunk ==

1. Evaluate the `features` section in declaration order to get one number per feature.
2. For every feature that appears in `membership`, each label's trapezoid gives a degree in [0,1]:
trapezoid(x; a,b,c,d) = 0 for x<=a or x>=d; 1 for b<=x<=c; linear on (a,b) and (c,d).
Shoulders: a == b means degree 1 for every x <= b; c == d means degree 1 for every x >= c.
Use shoulders on the outermost labels so that the whole axis is covered.
3. Rule firing strength = PRODUCT of its antecedent degrees.
4. Per expert, aggregate across rules by MAX.
5. Experts blocked by the safety fence this chunk are dropped.
6. The highest remaining weight wins, subject to stickiness: the current expert is kept until it
has been used for meta.stickiness.min_dwell chunks AND another expert's weight exceeds its
weight by more than meta.stickiness.margin (or the fence blocks it).
7. During the first meta.warmup.steps chunks, and whenever no rule fires, the router uses
meta.warmup.default. Avoid leaving holes.
Every video session (episode) starts afresh: windows are empty, diff is 0 on its first chunk,
and warmup and stickiness restart.

== The feature vocabulary ==

A feature is one node. `of` / `num` / `den` may only reference a feature declared EARLIER
in the object (no forward references, no cycles). Operators:

{"op":"obs", "field":<name>, "scale":<float>, "index":<int>}
a raw observation field, multiplied by `scale` (default 1.0).
`index` (0..5) is required only for next_video_chunk_sizes.
{"op":"throughput"}
goodput of the chunk just downloaded, kbps (= chunk_bytes*8/delay_ms).
{"op":"last_bitrate_kbps"}
the bitrate actually played on the previous chunk, kbps.
{"op":"diff", "of":<feature>}                          value now minus value last chunk
{"op":"window_mean",  "of":<feature>, "window":<int>}  over the last `window` chunks (1..64)
{"op":"window_std",   "of":<feature>, "window":<int>}
{"op":"window_cv",    "of":<feature>, "window":<int>}  std/mean, dimensionless
{"op":"window_slope", "of":<feature>, "window":<int>}  least-squares slope per chunk
{"op":"ema",   "of":<feature>, "alpha":<(0,1]>}        alpha on the new sample
{"op":"ratio", "num":<feature>, "den":<feature>}

Observation fields available to `obs`:
delay_ms                          download time of the chunk just fetched, milliseconds
sleep_time_ms                     time the player slept because the buffer was full, milliseconds
buffer_size_ms                    current playout buffer occupancy, milliseconds
rebuffer_ms                       stall time caused by the chunk just fetched, milliseconds
selected_video_chunk_size_bytes   bytes actually downloaded for that chunk
remain_chunk                      chunks left in the video
next_video_chunk_sizes            bytes of the NEXT chunk at each of the 6 bitrate rungs (6-vector, needs index 0..5)

You may declare intermediate features without membership functions (e.g. a raw throughput
series feeding a windowed statistic). Only features in `membership` can appear in rules.

== The safety fence ==

A hard filter, evaluated before aggregation, so that an expert cannot be selected in states
where you judge it catastrophic. You decide which experts it guards and the thresholds,
over your own declared features:

"fence": [{"expert": "<name>",
"require": [{"feature": "<declared feature>", "min": <float>},
{"feature": "<declared feature>", "max": <float>}]}]

An expert is blocked on any chunk where any of its `require` conditions fails.
Use [] if no expert needs guarding.

== Output ==

A SINGLE JSON object, nothing else — no prose, no markdown fence:

{
"features":   {"<name>": {"op": ..., ...}, ...},
"membership": {"<feature>": {"<Label>": [a,b,c,d], ...}, ...},
"rules":      [{"antecedents": {"<feature>": "<Label>", ...}, "consequent": "<expert>"}, ...],
"meta": {
"fence":      [...],
"stickiness": {"margin": <float>, "min_dwell": <int>},
"warmup":     {"steps": <int>, "default": "<expert>"}
},
"notes": ["<why you chose these features>", "<what regime each rule group targets>", ...]
}

Hard constraints:
- Breakpoints satisfy a <= b <= c <= d, in that feature's own units.
- Every label used in a rule is declared in `membership` for that feature.
- Every consequent is one of: pensieve, merina, pamoe, robustmpc, fdash.
- Every rule has at least one antecedent; it ignores the features it omits.
- No two rules have the same antecedents (their scores would always be equal).
- At most 24 rules. Prefer the smallest controller that covers the measured
regimes; every rule is one a human has to audit.

In `notes`, say why each feature earns its place. A feature that no measurement justifies
enlarges the rule space a human must read without buying discrimination.
\end{lstlisting}

\subsection{User prompt}
\begin{lstlisting}
## Expert pool

- pensieve: deep RL policy trained with PPO, using Pensieve's 6x8 history-matrix feature extractor
- merina: meta-RL policy: a VAE encodes the recent bandwidth trace into a latent, which conditions a PPO policy
- pamoe: sparse mixture-of-experts policy: a noisy top-k router over several expert heads, trained with PPO
- robustmpc: model-predictive control: harmonic-mean throughput prediction discounted by the worst recent prediction error, then a search over the bitrates of the next 3 chunks (no learned parameters)
- fdash: fuzzy-logic controller on buffer level and buffer change (no learned parameters)

## A. Expert profiles: each expert alone on each training trace family
QoE is the mean reward per chunk (bitrate reward - rebuffer penalty - smoothness penalty); higher is better.
stall_s: mean stall time per chunk (s). smooth: mean |delta bitrate| per chunk (Mbps).

scenario                            pensieve                    merina                     pamoe                 robustmpc                     fdash
QoE / stall_s / smooth    QoE / stall_s / smooth    QoE / stall_s / smooth    QoE / stall_s / smooth    QoE / stall_s / smooth
FCC-16-Train         +0.866 /  0.047 / 0.083   +0.861 /  0.047 / 0.086   +0.860 /  0.046 / 0.083   +0.717 /  0.074 / 0.158   +0.752 /  0.038 / 0.064
FCC-18-Train         +3.005 /  0.008 / 0.085   +3.018 /  0.009 / 0.087   +3.006 /  0.008 / 0.093   +2.934 /  0.018 / 0.148   +2.658 /  0.008 / 0.078
Oboe-Train           +2.269 /  0.005 / 0.105   +2.263 /  0.003 / 0.113   +2.262 /  0.003 / 0.118   +2.177 /  0.013 / 0.206   +1.910 /  0.001 / 0.100
Puffer-21-Train      +1.151 /  0.065 / 0.180   +1.128 /  0.068 / 0.185   +1.131 /  0.068 / 0.188   +0.749 /  0.142 / 0.312   +0.951 /  0.066 / 0.263
Puffer-22-Train      +0.849 /  0.108 / 0.099   +0.832 /  0.110 / 0.101   +0.835 /  0.108 / 0.106   +0.615 /  0.147 / 0.197   +0.707 /  0.090 / 0.107

Best expert per family:
FCC-16-Train       best=pensieve   QoE=+0.866  (spread across the pool 0.148)
FCC-18-Train       best=merina     QoE=+3.018  (spread across the pool 0.360)
Oboe-Train         best=pensieve   QoE=+2.269  (spread across the pool 0.359)
Puffer-21-Train    best=pensieve   QoE=+1.151  (spread across the pool 0.402)
Puffer-22-Train    best=pensieve   QoE=+0.849  (spread across the pool 0.233)

## B. Raw trace statistics: link capacity in the trace files (kbps)
scenario           traces     mean      p10      p25   median      p75      p90  cv_med  cv_p90
FCC-16-Train          151     1553      525      822     1267     2056     2871   0.110   0.324
FCC-18-Train          300     6400     1529     2163     4272     8355    13361   0.046   0.257
Oboe-Train            328     2806      850     1492     2758     3885     4701   0.042   0.214
Puffer-21-Train       500     2022       42      293     1188     3213     5211   0.643   1.230
Puffer-22-Train       500     1713      161      536     1222     2491     3938   0.369   0.917
cv = std/mean over a sliding window of 5 trace samples.
These are capacities, NOT what a router observes: the observed goodput (chunk_bytes*8/delay_ms) is
limited by the chunk sizes and download timing and can differ by a large factor. Set thresholds in
the observed space below.

## B2. Observed ranges of candidate quantities, per chunk, pooled over all experts playing
tp_mean_kbps: mean goodput (chunk_bytes*8/delay_ms) of the last 5 chunks; tp_cv: std/mean of those values;
buffer_s: playout buffer (s); buffer_diff_s: buffer change since the previous chunk (s). Windows restart every session.
They are ranges, not a prescribed set of inputs; you choose what to observe.

tp_mean_kbps:
scenario                 min        p5       p25       p50       p75       p95       max
FCC-16-Train              42       398       697      1014      1676      2942      5925
FCC-18-Train               4      1109      2045      3872      7457     16244     34723
Oboe-Train               127       573      1411      2571      3573      4702      9567
Puffer-21-Train           37       429      1078      2054      3231      5302     28589
Puffer-22-Train           14       360       754      1349      2276      3692     16368
tp_cv:
scenario                 min        p5       p25       p50       p75       p95       max
FCC-16-Train           0.000     0.031     0.069     0.127     0.213     0.372     1.183
FCC-18-Train           0.000     0.024     0.047     0.064     0.097     0.249     1.710
Oboe-Train             0.000     0.026     0.055     0.084     0.152     0.322     1.356
Puffer-21-Train        0.000     0.054     0.207     0.345     0.488     0.733     1.752
Puffer-22-Train        0.000     0.031     0.072     0.162     0.295     0.517     1.862
buffer_s:
scenario                 min        p5       p25       p50       p75       p95       max
FCC-16-Train            4.00      4.64     11.01     17.76     28.28     42.28     60.00
FCC-18-Train            4.00      6.01     13.26     23.02     42.35     59.83     60.00
Oboe-Train              4.00      5.57     10.37     16.79     25.39     41.27     60.00
Puffer-21-Train         4.00      4.91     13.09     20.60     29.04     39.97     60.00
Puffer-22-Train         4.00      4.80     10.91     17.77     27.46     41.24     60.00
buffer_diff_s:
scenario                 min        p5       p25       p50       p75       p95       max
FCC-16-Train          -32.75     -3.05     -0.44      0.54      1.61      2.80      3.77
FCC-18-Train          -55.96     -1.36     -0.08      0.53      1.55      2.96      3.90
Oboe-Train            -38.93     -2.16     -0.44      0.35      1.24      2.75      3.81
Puffer-21-Train       -55.99     -5.76     -0.52      1.14      2.54      3.49      3.88
Puffer-22-Train       -45.75     -3.36     -0.50      0.61      1.67      2.96      3.86

## Task

Write the routing controller in the format described in the system prompt. Route each measured regime to the expert the measurements say is strongest there.
\end{lstlisting}

\subsection{Model output}
\begin{lstlisting}
{
"features": {
"tp_kbps": {"op": "throughput"},
"tp_mean_kbps": {"op": "window_mean", "of": "tp_kbps", "window": 5}
},
"membership": {
"tp_mean_kbps": {
"NotVeryHigh": [0, 0, 5500, 7500],
"VeryHigh": [5500, 7500, 1000000, 1000000]
}
},
"rules": [
{"antecedents": {"tp_mean_kbps": "NotVeryHigh"}, "consequent": "pensieve"},
{"antecedents": {"tp_mean_kbps": "VeryHigh"}, "consequent": "merina"}
],
"meta": {
"fence": [],
"stickiness": {"margin": 0.2, "min_dwell": 4},
"warmup": {"steps": 5, "default": "pensieve"}
},
"notes": [
"Table A: pensieve has the best QoE on 4 of 5 families (FCC-16, Oboe, Puffer-21, Puffer-22). On FCC-18 it is 0.013 behind merina. pamoe and merina are never more than 0.023 behind pensieve. robustmpc and fdash are clearly worse everywhere: 0.07 to 0.40 below the best. So pensieve is the default and warmup expert, and robustmpc and fdash are never routed to.",
"The only measured regime where a different expert leads is FCC-18-Train (merina, +3.018 vs +3.006). Table B2 shows how to recognise it: FCC-18's tp_mean_kbps has p75 7457 and p95 16244. The p95 of every other family is lower: Oboe 4702, Puffer-21 5302, Puffer-22 3692, FCC-16 2942. So mean goodput over the last 5 chunks above about 7.5 Mbps is almost only FCC-18. The overlap ramp from 5.5 to 7.5 Mbps sits above the other families' p95. That is the only feature I included. Its window of 5 matches the window used in the B2 measurements.",
"Rule group 1 (tp_mean NotVeryHigh -> pensieve) covers FCC-16, Oboe, Puffer-21, Puffer-22 and the lower half of FCC-18. Rule group 2 (tp_mean VeryHigh -> merina) covers the high-capacity part of FCC-18. The labels use shoulders on both outer ends, so the whole axis is covered and no input is left without a firing rule.",
"Features I left out on purpose: tp_cv does separate Puffer-21 (median 0.345) from the others. But pensieve is already best on Puffer-21, so routing on it would change nothing. Buffer level and buffer change overlap heavily across families, and no expert other than pensieve wins in any buffer range that the measurements show. fdash's slightly lower stall on FCC-16 and Puffer-22 does not make up for its QoE loss (0.11 to 0.14). Adding any of these would make the rules bigger with no measured benefit.",
"Fence is empty. The two weak experts can never be chosen because no rule routes to them. The measurements show no state where pensieve or merina fails badly.",
"Stickiness (min_dwell 4, margin 0.2) prevents rapid switching near the 5.5 to 7.5 Mbps boundary. The merina gain on FCC-18 is only about 0.4% of QoE, so frequent switching could easily cost more than it gains. Warmup of 5 chunks on pensieve lets the 5-chunk throughput window fill before routing begins."
]
}
\end{lstlisting}

\newpage
\section{The Stage-2 Continual Round Prompt}
\label{app:roundprompt}

One full round of the second stage.
The system half is fixed across all 48 rounds.
The user half is rebuilt every round from the current page, the probe scores, and the evidence channels, and the one shown here is phase~2, round~3 (one validator message, logged in Chinese, is shown in English).
The model's answer to it is the single edit that moved the no-rule default from PA-MoE to MPC.
The final rule base, as it stands after phase~3, closes the appendix.

\subsection{System prompt}
\begin{lstlisting}
You maintain the controller of a router that picks, once per video chunk, which of several frozen pre-trained adaptive-bitrate (ABR) experts acts. The router only delegates: the quality played is the chosen expert's own decision for this chunk; the router never chooses a bitrate itself. Every expert sees every chunk, so an expert that is not chosen still keeps its state up to date, and it is told the quality actually played.

The controller is deployed. It was designed on the training trace families named below; since then the router has met new networks (Network-1, Network-2, ...). Of each new network only a few episodes were recorded. Every expert and the deployed controller were replayed on exactly these episodes, so their numbers are paired, but they describe a small sample of that network.

You are given ONLY measurements, the deployed controller, and the log of earlier revisions. You are NOT given any routing heuristic. At run time the router does not know which network it is on; it only sees what it observes. A regime is only actionable if the observations can tell it apart from the others.

Your revision is accepted only if, replayed on every network seen so far (each training family counts once, each new network once), it raises the mean QoE over them by at least 0.01 and leaves no network more than 0.01 below the best QoE it has had so far.

== How the deployed router runs your controller, once per chunk ==

1. Evaluate the `features` section in declaration order to get one number per feature.
2. For every feature that appears in `membership`, each label's trapezoid gives a degree in [0,1]:
trapezoid(x; a,b,c,d) = 0 for x<=a or x>=d; 1 for b<=x<=c; linear on (a,b) and (c,d).
Shoulders: a == b means degree 1 for every x <= b; c == d means degree 1 for every x >= c.
Use shoulders on the outermost labels so that the whole axis is covered.
3. Rule firing strength = PRODUCT of its antecedent degrees.
4. Per expert, aggregate across rules by MAX.
5. Experts blocked by the safety fence this chunk are dropped.
6. The highest remaining weight wins, subject to stickiness: the current expert is kept until it
has been used for meta.stickiness.min_dwell chunks AND another expert's weight exceeds its
weight by more than meta.stickiness.margin (or the fence blocks it).
7. During the first meta.warmup.steps chunks, and whenever no rule fires, the router uses
meta.warmup.default. Avoid leaving holes.
Every video session (episode) starts afresh: windows are empty, diff is 0 on its first chunk,
and warmup and stickiness restart.

== The feature vocabulary ==

A feature is one node. `of` / `num` / `den` may only reference a feature declared EARLIER
in the object (no forward references, no cycles). Operators:

{"op":"obs", "field":<name>, "scale":<float>, "index":<int>}
a raw observation field, multiplied by `scale` (default 1.0).
`index` (0..5) is required only for next_video_chunk_sizes.
{"op":"throughput"}
goodput of the chunk just downloaded, kbps (= chunk_bytes*8/delay_ms).
{"op":"last_bitrate_kbps"}
the bitrate actually played on the previous chunk, kbps.
{"op":"diff", "of":<feature>}                          value now minus value last chunk
{"op":"window_mean",  "of":<feature>, "window":<int>}  over the last `window` chunks (1..64)
{"op":"window_std",   "of":<feature>, "window":<int>}
{"op":"window_cv",    "of":<feature>, "window":<int>}  std/mean, dimensionless
{"op":"window_slope", "of":<feature>, "window":<int>}  least-squares slope per chunk
{"op":"ema",   "of":<feature>, "alpha":<(0,1]>}        alpha on the new sample
{"op":"ratio", "num":<feature>, "den":<feature>}

Observation fields available to `obs`:
delay_ms                          download time of the chunk just fetched, milliseconds
sleep_time_ms                     time the player slept because the buffer was full, milliseconds
buffer_size_ms                    current playout buffer occupancy, milliseconds
rebuffer_ms                       stall time caused by the chunk just fetched, milliseconds
selected_video_chunk_size_bytes   bytes actually downloaded for that chunk
remain_chunk                      chunks left in the video
next_video_chunk_sizes            bytes of the NEXT chunk at each of the 6 bitrate rungs (6-vector, needs index 0..5)

You may declare intermediate features without membership functions (e.g. a raw throughput
series feeding a windowed statistic). Only features in `membership` can appear in rules.

== The safety fence ==

A hard filter, evaluated before aggregation, so that an expert cannot be selected in states
where you judge it catastrophic. You decide which experts it guards and the thresholds,
over your own declared features:

"fence": [{"expert": "<name>",
"require": [{"feature": "<declared feature>", "min": <float>},
{"feature": "<declared feature>", "max": <float>}]}]

An expert is blocked on any chunk where any of its `require` conditions fails.
Use [] if no expert needs guarding.

== Output ==

A SINGLE JSON object, nothing else — no prose, no markdown fence:

{
"features":   {"<name>": {"op": ..., ...}, ...},
"membership": {"<feature>": {"<Label>": [a,b,c,d], ...}, ...},
"rules":      [{"antecedents": {"<feature>": "<Label>", ...}, "consequent": "<expert>"}, ...],
"meta": {
"fence":      [...],
"stickiness": {"margin": <float>, "min_dwell": <int>},
"warmup":     {"steps": <int>, "default": "<expert>"}
},
"notes": ["<why you chose these features>", "<what regime each rule group targets>", ...]
}

Hard constraints:
- Breakpoints satisfy a <= b <= c <= d, in that feature's own units.
- Every label used in a rule is declared in `membership` for that feature.
- Every consequent is one of: pensieve, merina, pamoe, robustmpc, fdash.
- Every rule has at least one antecedent; it ignores the features it omits.
- No two rules have the same antecedents (their scores would always be equal).
- At most 24 rules. Prefer the smallest controller that covers the measured
regimes; every rule is one a human has to audit.

In `notes`, say why each feature earns its place. A feature that no measurement justifies
enlarges the rule space a human must read without buying discrimination.
\end{lstlisting}

\subsection{User prompt (phase 2, round 3)}
\begin{lstlisting}
## Expert pool

- pensieve: deep RL policy trained with PPO, using Pensieve's 6x8 history-matrix feature extractor
- merina: meta-RL policy: a VAE encodes the recent bandwidth trace into a latent, which conditions a PPO policy
- pamoe: sparse mixture-of-experts policy: a noisy top-k router over several expert heads, trained with PPO
- robustmpc: model-predictive control: harmonic-mean throughput prediction discounted by the worst recent prediction error, then a search over the bitrates of the next 3 chunks (no learned parameters)
- fdash: fuzzy-logic controller on buffer level and buffer change (no learned parameters)

## A. Expert profiles: each expert alone on each training trace family

QoE is the mean reward per chunk (bitrate reward - rebuffer penalty - smoothness penalty); higher is better.
stall_s: mean stall time per chunk (s). smooth: mean |delta bitrate| per chunk (Mbps).

scenario                            pensieve                    merina                     pamoe                 robustmpc                     fdash
QoE / stall_s / smooth    QoE / stall_s / smooth    QoE / stall_s / smooth    QoE / stall_s / smooth    QoE / stall_s / smooth
FCC-16-Train         +0.866 /  0.047 / 0.083   +0.861 /  0.047 / 0.086   +0.860 /  0.046 / 0.083   +0.717 /  0.074 / 0.158   +0.752 /  0.038 / 0.064
FCC-18-Train         +3.005 /  0.008 / 0.085   +3.018 /  0.009 / 0.087   +3.006 /  0.008 / 0.093   +2.934 /  0.018 / 0.148   +2.658 /  0.008 / 0.078
Oboe-Train           +2.269 /  0.005 / 0.105   +2.263 /  0.003 / 0.113   +2.262 /  0.003 / 0.118   +2.177 /  0.013 / 0.206   +1.910 /  0.001 / 0.100
Puffer-21-Train      +1.151 /  0.065 / 0.180   +1.128 /  0.068 / 0.185   +1.131 /  0.068 / 0.188   +0.749 /  0.142 / 0.312   +0.951 /  0.066 / 0.263
Puffer-22-Train      +0.849 /  0.108 / 0.099   +0.832 /  0.110 / 0.101   +0.835 /  0.108 / 0.106   +0.615 /  0.147 / 0.197   +0.707 /  0.090 / 0.107
Network-1            +2.113 /  0.008 / 0.521   +1.923 /  0.159 / 0.395   +1.547 /  0.001 / 0.729   +2.596 /  0.049 / 0.198   -5.072 /  2.075 / 0.106

Best expert per family:
FCC-16-Train       best=pensieve   QoE=+0.866  (spread across the pool 0.148)
FCC-18-Train       best=merina     QoE=+3.018  (spread across the pool 0.360)
Oboe-Train         best=pensieve   QoE=+2.269  (spread across the pool 0.359)
Puffer-21-Train    best=pensieve   QoE=+1.151  (spread across the pool 0.402)
Puffer-22-Train    best=pensieve   QoE=+0.849  (spread across the pool 0.233)
Network-1          best=robustmpc  QoE=+2.596  (spread across the pool 7.668)

## B. Raw trace statistics: link capacity in the trace files (kbps)

scenario           traces     mean      p10      p25   median      p75      p90  cv_med  cv_p90
FCC-16-Train          151     1553      525      822     1267     2056     2871   0.110   0.324
FCC-18-Train          300     6400     1529     2163     4272     8355    13361   0.046   0.257
Oboe-Train            328     2806      850     1492     2758     3885     4701   0.042   0.214
Puffer-21-Train       500     2022       42      293     1188     3213     5211   0.643   1.230
Puffer-22-Train       500     1713      161      536     1222     2491     3938   0.369   0.917
Network-1              16    39324     4000    11000    27000    56000    91000   0.178   0.474
cv = std/mean over a sliding window of 5 trace samples.
These are capacities, NOT what a router observes: the observed goodput (chunk_bytes*8/delay_ms) is
limited by the chunk sizes and download timing and can differ by a large factor. Set thresholds in
the observed space below.

## B2. Observed ranges of candidate quantities, per chunk, pooled over all experts playing

tp_mean_kbps: mean goodput (chunk_bytes*8/delay_ms) of the last 5 chunks; tp_cv: std/mean of those values;
buffer_s: playout buffer (s); buffer_diff_s: buffer change since the previous chunk (s). Windows restart every session.
They are ranges, not a prescribed set of inputs; you choose what to observe.

tp_mean_kbps:
scenario                 min        p5       p25       p50       p75       p95       max
FCC-16-Train              42       398       697      1014      1676      2942      5925
FCC-18-Train               4      1109      2045      3872      7457     16244     34723
Oboe-Train               127       573      1411      2571      3573      4702      9567
Puffer-21-Train           37       429      1078      2054      3231      5302     28589
Puffer-22-Train           14       360       754      1349      2276      3692     16368
Network-1               1649      3780     14055     28672     55328    110545    208107
tp_cv:
scenario                 min        p5       p25       p50       p75       p95       max
FCC-16-Train           0.000     0.031     0.069     0.127     0.213     0.372     1.183
FCC-18-Train           0.000     0.024     0.047     0.064     0.097     0.249     1.710
Oboe-Train             0.000     0.026     0.055     0.084     0.152     0.322     1.356
Puffer-21-Train        0.000     0.054     0.207     0.345     0.488     0.733     1.752
Puffer-22-Train        0.000     0.031     0.072     0.162     0.295     0.517     1.862
Network-1              0.000     0.057     0.130     0.224     0.358     0.606     1.584
buffer_s:
scenario                 min        p5       p25       p50       p75       p95       max
FCC-16-Train            4.00      4.64     11.01     17.76     28.28     42.28     60.00
FCC-18-Train            4.00      6.01     13.26     23.02     42.35     59.83     60.00
Oboe-Train              4.00      5.57     10.37     16.79     25.39     41.27     60.00
Puffer-21-Train         4.00      4.91     13.09     20.60     29.04     39.97     60.00
Puffer-22-Train         4.00      4.80     10.91     17.77     27.46     41.24     60.00
Network-1               4.00      4.00     13.03     26.66     45.81     59.79     60.00
buffer_diff_s:
scenario                 min        p5       p25       p50       p75       p95       max
FCC-16-Train          -32.75     -3.05     -0.44      0.54      1.61      2.80      3.77
FCC-18-Train          -55.96     -1.36     -0.08      0.53      1.55      2.96      3.90
Oboe-Train            -38.93     -2.16     -0.44      0.35      1.24      2.75      3.81
Puffer-21-Train       -55.99     -5.76     -0.52      1.14      2.54      3.49      3.88
Puffer-22-Train       -45.75     -3.36     -0.50      0.61      1.67      2.96      3.86
Network-1             -48.04     -3.52     -0.05      0.93      2.43      3.66      3.91
## C. The deployed controller on every network (same replays as Table A)

regret = its QoE minus that of the best single expert on the network; share = fraction of decisions given to each expert.

network                QoE   best expert  best QoE   regret   share
FCC-16-Train        +0.866      pensieve    +0.866   +0.000   pensieve 1.00  merina 0.00  pamoe 0.00  robustmpc 0.00  fdash 0.00
FCC-18-Train        +3.006        merina    +3.018   -0.012   pensieve 0.71  merina 0.29  pamoe 0.00  robustmpc 0.00  fdash 0.00
Oboe-Train          +2.269      pensieve    +2.269   +0.000   pensieve 1.00  merina 0.00  pamoe 0.00  robustmpc 0.00  fdash 0.00
Puffer-21-Train     +1.150      pensieve    +1.151   -0.001   pensieve 0.98  merina 0.02  pamoe 0.00  robustmpc 0.00  fdash 0.00
Puffer-22-Train     +0.849      pensieve    +0.849   +0.000   pensieve 1.00  merina 0.00  pamoe 0.00  robustmpc 0.00  fdash 0.00
Network-1           +2.003     robustmpc    +2.596   -0.593   pensieve 0.23  merina 0.77  pamoe 0.00  robustmpc 0.00  fdash 0.00

## D. Best expert of each recorded episode

For each new network: which expert has the highest QoE on each recorded episode, and how consistently.
A difference that holds on most episodes is a property of the network; one that flips from episode to episode may not be.
Network-1 (16 episodes): best per episode robustmpc 9, fdash 4, pensieve 2, merina 1; robustmpc (best on average) beats every other expert on 9 of 16 episodes

## The deployed controller

{
"features": {
"tp_kbps": {
"op": "throughput"
},
"tp_mean_kbps": {
"op": "window_mean",
"of": "tp_kbps",
"window": 5
}
},
"membership": {
"tp_mean_kbps": {
"NotVeryHigh": [
0,
0,
5500,
7500
],
"VeryHigh": [
5500,
7500,
1000000,
1000000
]
}
},
"rules": [
{
"antecedents": {
"tp_mean_kbps": "NotVeryHigh"
},
"consequent": "pensieve"
},
{
"antecedents": {
"tp_mean_kbps": "VeryHigh"
},
"consequent": "merina"
}
],
"meta": {
"fence": [],
"stickiness": {
"margin": 0.2,
"min_dwell": 4
},
"warmup": {
"steps": 5,
"default": "pensieve"
}
},
"notes": [
"Table A: pensieve has the best QoE on 4 of 5 families (FCC-16, Oboe, Puffer-21, Puffer-22). On FCC-18 it is 0.013 behind merina. pamoe and merina are never more than 0.023 behind pensieve. robustmpc and fdash are clearly worse everywhere: 0.07 to 0.40 below the best. So pensieve is the default and warmup expert, and robustmpc and fdash are never routed to.",
"The only measured regime where a different expert leads is FCC-18-Train (merina, +3.018 vs +3.006). Table B2 shows how to recognise it: FCC-18's tp_mean_kbps has p75 7457 and p95 16244. The p95 of every other family is lower: Oboe 4702, Puffer-21 5302, Puffer-22 3692, FCC-16 2942. So mean goodput over the last 5 chunks above about 7.5 Mbps is almost only FCC-18. The overlap ramp from 5.5 to 7.5 Mbps sits above the other families' p95. That is the only feature I included. Its window of 5 matches the window used in the B2 measurements.",
"Rule group 1 (tp_mean NotVeryHigh -> pensieve) covers FCC-16, Oboe, Puffer-21, Puffer-22 and the lower half of FCC-18. Rule group 2 (tp_mean VeryHigh -> merina) covers the high-capacity part of FCC-18. The labels use shoulders on both outer ends, so the whole axis is covered and no input is left without a firing rule.",
"Features I left out on purpose: tp_cv does separate Puffer-21 (median 0.345) from the others. But pensieve is already best on Puffer-21, so routing on it would change nothing. Buffer level and buffer change overlap heavily across families, and no expert other than pensieve wins in any buffer range that the measurements show. fdash's slightly lower stall on FCC-16 and Puffer-22 does not make up for its QoE loss (0.11 to 0.14). Adding any of these would make the rules bigger with no measured benefit.",
"Fence is empty. The two weak experts can never be chosen because no rule routes to them. The measurements show no state where pensieve or merina fails badly.",
"Stickiness (min_dwell 4, margin 0.2) prevents rapid switching near the 5.5 to 7.5 Mbps boundary. The merina gain on FCC-18 is only about 0.4% of QoE, so frequent switching could easily cost more than it gains. Warmup of 5 chunks on pensieve lets the 5-chunk throughput window fill before routing begins."
]
}

## Revision log

(no earlier revisions)

## Task

Write the revised routing controller in full, in the format described in the system prompt. It replaces the deployed one. Keep what the measurements still support, change what they contradict, and say in `notes` what you changed and why.
\end{lstlisting}

\subsection{Model output}
\begin{lstlisting}
{
"features": {
"tp_kbps": {"op": "throughput"},
"tp_mean_kbps": {"op": "window_mean", "of": "tp_kbps", "window": 5}
},
"membership": {
"tp_mean_kbps": {
"NotExtreme": [0, 0, 12000, 18000],
"Extreme": [12000, 18000, 1000000, 1000000]
}
},
"rules": [
{
"antecedents": {"tp_mean_kbps": "NotExtreme"},
"consequent": "pensieve"
},
{
"antecedents": {"tp_mean_kbps": "Extreme"},
"consequent": "robustmpc"
}
],
"meta": {
"fence": [],
"stickiness": {"margin": 0.2, "min_dwell": 4},
"warmup": {"steps": 5, "default": "pensieve"}
},
"notes": [
"Kept: there is still one feature, tp_mean_kbps (mean goodput over the last 5 chunks, the same window as Table B2). It is the only measured quantity that tells Network-1 apart from the training families. Network-1 tp_mean has p25 14055, p50 28672 and p75 55328. FCC-18 has p95 16244, and every other training family has p95 at or below 5302. tp_cv, buffer and buffer_diff overlap heavily between Network-1 and the training families, so I did not add them.",
"Changed 1: very high goodput now goes to robustmpc instead of merina. Table C shows this rule sent 77% of Network-1 decisions to merina. On Network-1 merina is the second-worst expert (+1.923, stall 0.159 s), so the deployed controller got +2.003 against robustmpc's +2.596 (regret -0.593). robustmpc is best on average on Network-1 and best on 9 of 16 episodes (Table D), so the result holds across episodes and is not one outlier. The ramp now runs from 12 to 18 Mbps, so roughly 70% or more of Network-1 chunks fall in the Extreme label. Only about the top 5-8% of FCC-18 chunks reach it, and no chunks from the other families except rare outliers. At goodputs that high, every expert should be at the top bitrate rung, so robustmpc's FCC-18 deficit (0.07, mostly stalls and switching at lower rates) should barely show on those chunks.",
"Changed 2: the middle band (7.5 to 12-18 Mbps), which used to go to merina, now goes to pensieve. On FCC-18, merina routing gained only +0.001 over pensieve alone (deployed 3.006 vs pensieve-only 3.005). So dropping it costs at most about 0.001 there, well inside the 0.01 tolerance. On Network-1, pensieve stalls far less than merina (0.008 vs 0.159 s per chunk) and scores higher (+2.113 vs +1.923). So the Network-1 chunks below the robustmpc ramp are safer with pensieve.",
"Rule groups: (1) NotExtreme -> pensieve covers FCC-16, Oboe, Puffer-21, Puffer-22, almost all of FCC-18, and the lower-goodput stretches of Network-1. These keep their deployed QoE, which was within 0.001 of pensieve alone. (2) Extreme -> robustmpc covers the multi-ten-Mbps regime that is typical of Network-1. Both outer labels use shoulders, so no input is left without a firing rule.",
"Not routed to: fdash. It wins 4 Network-1 episodes, but its mean there is -5.072 with 2.075 s of stall per chunk, and nothing we observe predicts which episodes it would win. pamoe is never best anywhere. The fence stays empty because no rule routes to fdash or pamoe.",
"Stickiness (margin 0.2, min_dwell 4) and a 5-chunk pensieve warmup are unchanged. The warmup fills the 5-chunk window, and the stickiness damps switching near the 12 to 18 Mbps ramp. Every expert tracks every chunk, so robustmpc's error history is up to date when it takes over.",
"Caveat: Network-1 has only 16 recorded episodes. The 12/18 Mbps thresholds are chosen to sit above FCC-18's p95, so that the training families stay within their 0.01 tolerance even if the Network-1 gain turns out smaller than these numbers suggest."
]
}
\end{lstlisting}

\subsection{The final rule base}
\begin{lstlisting}
{
"features": {
"tp_kbps": {"op": "throughput"},
"tp_mean_kbps": {"op": "window_mean", "of": "tp_kbps", "window": 5},
"tp_mean20_kbps": {"op": "window_mean", "of": "tp_kbps", "window": 20},
"buffer_s": {"op": "obs", "field": "buffer_size_ms", "scale": 0.001}
},
"membership": {
"tp_mean_kbps": {
"Low": [0, 0, 2500, 3500],
"Mid": [2500, 3500, 7000, 10000],
"Extreme": [7000, 10000, 2000000, 2000000]
},
"tp_mean20_kbps": {
"NotExtreme": [0, 0, 5000, 8000],
"Extreme": [5000, 8000, 40000, 60000],
"Huge": [40000, 60000, 2000000, 2000000],
"NotHuge": [0, 0, 40000, 60000]
},
"buffer_s": {
"Low": [0, 0, 10, 14],
"Ok": [10, 14, 60, 60],
"NotFull": [0, 0, 40, 50],
"Full": [40, 50, 60, 60]
}
},
"rules": [
{
"antecedents": {"tp_mean_kbps": "Low", "buffer_s": "Ok", "tp_mean20_kbps": "NotExtreme"},
"consequent": "pensieve"
},
{
"antecedents": {"tp_mean_kbps": "Mid", "tp_mean20_kbps": "NotExtreme"},
"consequent": "pensieve"
},
{
"antecedents": {"tp_mean_kbps": "Low", "buffer_s": "Low", "tp_mean20_kbps": "NotExtreme"},
"consequent": "pamoe"
},
{
"antecedents": {"tp_mean_kbps": "Extreme", "tp_mean20_kbps": "NotHuge"},
"consequent": "robustmpc"
},
{
"antecedents": {"tp_mean20_kbps": "Extreme"},
"consequent": "robustmpc"
},
{
"antecedents": {"tp_mean20_kbps": "Huge", "buffer_s": "Full"},
"consequent": "fdash"
},
{
"antecedents": {"tp_mean20_kbps": "Huge", "buffer_s": "NotFull"},
"consequent": "robustmpc"
}
],
"meta": {
"fence": [
{
"expert": "fdash",
"require": [
{"feature": "buffer_s", "min": 30},
{"feature": "tp_mean20_kbps", "min": 40000}
]
}
],
"stickiness": {"margin": 0.2, "min_dwell": 3},
"warmup": {"steps": 2, "default": "pensieve"}
},
"notes": [
"Floors (best so far minus 0.01): FCC-16 0.865, FCC-18 3.006, Oboe 2.233, Puffer-21 1.093, Puffer-22 0.886, Network-1 2.535, Network-2 3.812, Network-3 2.162, Network-4 -0.333. The QoE sum has to rise by about 0.09.",
"What stage 4 round 2 showed. It made two changes. (a) fdash during startup on Network-2: rule 'Huge and NotFull and Rising -> fdash', with the fence floor lowered to 6 s. Network-2 lost 0.128, so the measurement contradicts that change. I revert it completely. The Network-2 region is again exactly the accepted design: Huge and Full -> fdash, Huge and NotFull -> robustmpc, fence at buffer >= 30 s and 20-chunk mean >= 40 Mbps. The bufdiff_s feature is dropped. With it gone, Network-2 should return to its accepted 3.822. (b) Widening the pamoe low-buffer corner from a 6-9 s ramp to 8-11 s. Network-4 gained +0.054 (-0.323 to -0.269). The training families moved by at most -0.002 (Oboe), and Puffer-21 and FCC-16 moved by +0.001. The measurements support this lever, and it still has headroom: pamoe alone scores -0.223 on Network-4, and pamoe and robustmpc are each best on 5 of 16 episodes there.",
"Change 1 (Network-4): the buffer Low/Ok ramp moves further, from 8-11 s to 10-14 s. Network-4's buffer median is 11.9 s and its p75 is 21 s, so roughly half its chunks are now fully Low. Every widening so far has gained on Network-4 at almost no training cost (6-9 s: +0.044; 8-11 s: +0.054 more). That fits the stall mechanism: pamoe stalls 0.138 s per chunk on Network-4 against pensieve's 0.182, and stalls start from a shallow buffer. Expected gain is a further +0.02 to +0.04, which is a projection, not a measurement.",
"Change 2 (to protect training families): the 5-chunk Low label, which gates the pamoe rule, moves from 3-4 Mbps to 2.5-3.5 Mbps. Mid moves to match, so Low, Mid and Extreme still sum to 1. Oboe is the training family closest to its floor (2.238 after round 2, floor 2.233), and it was the only one that lost from the wider corner. Its goodput median is 2571 kbps, so a lower gate takes a large part of Oboe out of the pamoe corner. Network-4 stays mostly inside: goodput p50 1843 and p75 2377. FCC-16 (p75 1676) and Puffer-22 (p75 2276) stay exposed, but neither lost anything from earlier widenings. Puffer-21 is where pamoe is weakest (-0.020 for the whole family), yet it moved +0.001 under the round-2 widening, so low-buffer Puffer-21 chunks evidently do not hurt pamoe.",
"Rollback order if a floor breaks. If Oboe or Puffer-21 breaks, put the buffer ramp back to 8-11 s first. If Network-4 gains less than expected, put the 5-chunk Low ramp back to 3-4 Mbps.",
"Unchanged, with reasons. tp_mean_kbps (5 chunks) is still the only observation that separates fast Network-1 and Network-3 stretches from FCC-16, Oboe and the Puffer families, and its Low label now only gates the pamoe corner. tp_mean20_kbps (20 chunks) is session memory. Its Extreme ramp (5-8 Mbps) gave Network-3 +0.082 in stage 4 and was kept. I do not lower it further because FCC-18 is only 0.006 above its floor. Its Huge label marks Network-2's sustained sessions. buffer_s marks the shallow-buffer state where Network-4 loses QoE to stalls, and the Full state that bounds fdash's stall risk. tp_cv and buffer_diff are left out: tp_cv overlaps across Network-4, FCC-16 and Puffer-22, and buffer_diff gating is what broke Network-2.",
"Rule groups. Pensieve (rules 1-2) covers the training families and slow stretches outside the corner. Pamoe (rule 3) is the complement of rule 1 on buffer, so pensieve's weight there falls to 0 and the margin does not block the switch. Robustmpc (rules 4, 5, 7) covers fast links, sustained fast sessions, and Huge sessions with a buffer not yet full. Fdash (rule 6) covers Huge sessions with a Full buffer. All label pairs are complementary, so some rule always fires. robustmpc is never fenced, so an expert is always available. Stickiness (margin 0.2, min_dwell 3) and warmup (2 chunks, pensieve) are unchanged; earlier rounds found no gain in tuning them. merina gets no rule: pamoe matches it on Network-4 and stalls less on Network-1 and Network-3."
]
}
\end{lstlisting}

\end{document}

%% file: parallel_env.tex
\subsection{Parallel Environment Simulation on GPU}
\label{sec:gpu_sim}

The system model of Appendix~\ref{sec:system-model} advances one episode chunk by chunk. A direct implementation, as in the reference simulator, computes the transfer time $T_i^{\mathrm{tr}}$ by walking through the trace interval by interval until $d_i(b_i)$ bytes are accumulated, and advances the trace time through the waiting period $\Delta t_i^{w}$ in the same way. The number of iterations of these loops depends on the chunk size, the bandwidth and the current position in the trace, and therefore differs across episodes. Such data-dependent loops cannot be executed efficiently in a single-instruction-multiple-data fashion. We therefore reformulate one decision step as a fixed sequence of element-wise operations, so that $E$ independent environments can be simulated simultaneously on a GPU while producing exactly the same dynamics.

\paragraph{\textbf{Batched state:}}
Each environment $e \in \{1, \dots, E\}$ is described by the state $\bigl(n^{(e)}, \tau^{(e)}, B^{(e)}, i^{(e)}, b^{(e)}_{i-1}\bigr)$, i.e., the index of its trace in $\mathcal{D}$, the current trace time, the buffer occupancy, the chunk index and the previous bitrate. Each component is stored as a vector of length $E$, and all operations below are applied element-wise; we drop the superscript $(e)$ for brevity. Since traces have different lengths, all traces in $\mathcal{D}$ are concatenated into flat arrays, together with the offset and length of each trace, so that every environment can index its own trace within the same batched operation.

\paragraph{\textbf{Cumulative byte curve:}}
For each trace $\mathcal{T}^{(n)}$, we precompute the number of bytes that can be downloaded from the beginning of the trace up to time $\tau$:
\begin{equation}
  \Phi_n(\tau) = \int_{0}^{\tau} S(t)\, dt
  = \Phi_n(t_{k-1}) + S_k \,(\tau - t_{k-1}),
  \quad \tau \in (t_{k-1}, t_k],
\end{equation}
where $\Phi_n(t_k) = \sum_{k'=1}^{k} S_{k'}\,(t_{k'} - t_{k'-1})$ is stored at every breakpoint. Since $S_k \geq 0$, $\Phi_n$ is piecewise linear and non-decreasing. Let $\Phi_n^{\mathrm{tot}} = \Phi_n(T)$ denote the bytes available in one pass of the trace. Under cyclic replay, the curve extends as $\bar{\Phi}_n(\tau + \nu T) = \Phi_n(\tau) + \nu\,\Phi_n^{\mathrm{tot}}$ for $\nu \in \mathbb{N}$.

\paragraph{\textbf{Closed-form transfer time:}}
With $\bar{\Phi}_n$, the transfer condition of the system model becomes $\bar{\Phi}_n(\tau_i + T_i^{\mathrm{tr}}) - \bar{\Phi}_n(\tau_i) = d_i(b_i)$. Instead of walking through the intervals, we solve it directly. Let $\beta_i = \Phi_n(\tau_i) + d_i(b_i)$ be the target cumulative bytes. The number of complete passes over the trace and the end position within the last pass are
\begin{equation}
	\nu_i = \left\lfloor \frac{\beta_i}{\Phi_n^{\mathrm{tot}}} \right\rfloor,
	\qquad
	\tau_i' = \Phi_n^{-1}\!\left( \beta_i - \nu_i\, \Phi_n^{\mathrm{tot}} \right),
\end{equation}
and the transfer time follows as
\begin{equation}
	T_i^{\mathrm{tr}} = \nu_i T + \tau_i' - \tau_i .
\end{equation}
The inverse of the piecewise-linear curve is
\begin{equation}
	\Phi_n^{-1}(\beta) = t_{k-1} + \frac{\beta - \Phi_n(t_{k-1})}{S_k},
	\qquad
	k = \min\{\, k : \Phi_n(t_k) > \beta \,\},
\end{equation}
where the strict inequality skips intervals with zero capacity, so the download ends in the first interval that actually delivers the remaining bytes.

\paragraph{\textbf{Fixed-cost interval lookup:}}
Both $\Phi_n(\tau_i)$ and $\Phi_n^{-1}(\cdot)$ require locating the interval that contains a given time or byte count. Because the breakpoints $\{t_k\}$ and $\{\Phi_n(t_k)\}$ are sorted, the interval is found by binary search. We run the search for a fixed number of $\lceil \log_2 (K_{\max}+1) \rceil + 1$ iterations for all environments, where $K_{\max}$ is the length of the longest trace in $\mathcal{D}$. Every environment thus executes the same instructions, regardless of how many intervals its download spans, and the cost of a step drops from $O(\text{number of traversed intervals})$ to $O(\log K_{\max})$.

\paragraph{\textbf{Remaining updates:}}
The rest of the step consists of element-wise operations only. A noise factor $\eta_i$ is sampled independently for each environment, and the delay, rebuffering time, buffer and waiting time are computed exactly as in the system model, with the conditional cases implemented by masked selection rather than branching. The trace time after waiting is obtained without a loop:
\begin{equation}
  \tau_{i+1} = \left( \tau_i' + \Delta t_i^{w} \right) \bmod T .
\end{equation}

\paragraph{\textbf{In-place reset:}}
An episode ends after its $I$-th chunk is downloaded. The environment then starts a new episode within the same step, following exactly the same procedure as at the beginning of training: a new trace is sampled from $\mathcal{D}$, a new initial trace time $\tau_1$ is drawn from its breakpoints, the buffer, the chunk index and the previous bitrate are cleared, and the warm-up chunk $V_1$ is downloaded at the lowest bitrate $b^{(1)}$ without reward. Every episode therefore consists of the warm-up chunk followed by $I-1$ decision steps. The reset is applied through a mask, so it requires no separate reset call and only a single check of whether any episode has ended; all other data stay on the GPU. Since all episodes have the same length $I$, the environments currently end their episodes at the same step; the masked formulation also supports episodes of different lengths.

\paragraph{\textbf{Pseudocode of the batched GPU simulator:}}
Algorithm~\ref{alg:gpu_env} gives the complete procedure. Its structure follows the sequential simulator closely; what changes is how a download is computed. Instead of walking through the trace interval by interval, \texttt{Download} adds the chunk size to the cumulative byte count at the current position and maps the result back to a time through $\Phi^{-1}$ (lines~10--13), where $\boldsymbol{\nu}$ accounts for downloads that run past the end of the trace. Both lookups use a binary search whose depth $H$ is set by the longest trace, so environments with short and long traces execute the same instructions. The waiting period is handled in the same spirit, by a modulo on the trace time rather than a loop (line~17).

Resets are applied through masks as well. \texttt{StartEpisode} performs the warm-up download for every environment and then restores $\boldsymbol{\tau}$, $\mathbf{B}$ and $\mathbf{i}$ wherever $\mathbf{M}$ is false. This wastes one download for the environments that are not reset, but it keeps the batch intact. The host only needs to read GPU data once per step, for the test $\operatorname{any}(\mathbf{done})$ in \texttt{Step}.

\begin{algorithm}[!htbp]
	\caption{Batched simulation of $E$ environments on GPU}
	\label{alg:gpu_env}
	\small
	\SetKwFunction{FLookup}{Lookup}
	\SetKwFunction{FDownload}{Download}
	\SetKwFunction{FStart}{StartEpisode}
	\SetKwFunction{FStep}{Step}
	\SetKwProg{Fn}{Function}{:}{}
	\SetKwComment{Comment}{$\triangleright$\ }{}
	\SetCommentSty{textit}
	\DontPrintSemicolon
	\KwIn{Trace dataset $\mathcal{D}$, chunk sizes $d_i(b)$. Bold symbols are
		vectors in $\mathbb{R}^{E}$ (one entry per environment), and all operations
		on them are element-wise and run in parallel;
		$\operatorname{where}(\mathbf{M}, \mathbf{x}, \boldsymbol{\beta})$ selects $x_e$ if
		$M_e$ holds and $y_e$ otherwise.}
	\KwData{Trace index $\mathbf{n}$, trace time $\boldsymbol{\tau}$, buffer
		$\mathbf{B}$, next chunk index $\mathbf{i}$, previous bitrate
		$\mathbf{b}^{\mathrm{prev}}$.}
	Precompute $T_n$, $\{S_k\}$, $\{\Phi_n(t_k)\}$, $\Phi_n^{\mathrm{tot}}$ for every
	trace and store them as flat arrays on the GPU\;
	$H \gets \lceil \log_2 (K_{\max}+1) \rceil + 1$ \Comment*[r]{fixed search depth}
	\BlankLine
	\Fn(\Comment*[f]{first $k$ with $a_{\mathbf{n}}[k] > \mathbf{x}$}){\FLookup{$a$, $\mathbf{x}$}}{
		$\mathbf{lo} \gets \mathbf{0}$;\quad $\mathbf{hi} \gets \mathbf{K}_{\mathbf{n}} + 1$\;
		\For(\Comment*[f]{same $H$ for all envs}){$h = 1$ \KwTo $H$}{
			$\mathbf{mid} \gets \lfloor (\mathbf{lo} + \mathbf{hi}) / 2 \rfloor$;\quad
			$\mathbf{go} \gets (\mathbf{lo} < \mathbf{hi}) \wedge (a_{\mathbf{n}}[\mathbf{mid}] \le \mathbf{x})$\;
			$\mathbf{lo} \gets \operatorname{where}(\mathbf{go}, \mathbf{mid}+1, \mathbf{lo})$;\quad
			$\mathbf{hi} \gets \operatorname{where}(\neg\mathbf{go} \wedge (\mathbf{lo}<\mathbf{hi}), \mathbf{mid}, \mathbf{hi})$\;
		}
		\KwRet{$\mathbf{lo}$}\;
	}
	\BlankLine
	\Fn{\FDownload{$\mathbf{b}$}}{
		$\mathbf{k} \gets$ \FLookup{$t$, $\boldsymbol{\tau}$};\quad
		$\boldsymbol{\beta} \gets \Phi_{\mathbf{n}}(t_{\mathbf{k}-1}) + S_{\mathbf{k}} (\boldsymbol{\tau} - t_{\mathbf{k}-1}) + d_{\mathbf{i}}(\mathbf{b})$\;
		$\boldsymbol{\nu} \gets \lfloor \boldsymbol{\beta} / \Phi_{\mathbf{n}}^{\mathrm{tot}} \rfloor$;\quad
		$\boldsymbol{\beta}' \gets \boldsymbol{\beta} - \boldsymbol{\nu}\, \Phi_{\mathbf{n}}^{\mathrm{tot}}$
		\Comment*[r]{passes over trace end}
		$\mathbf{k}' \gets$ \FLookup{$\Phi$, $\boldsymbol{\beta}'$};\quad
		$\boldsymbol{\tau}' \gets t_{\mathbf{k}'-1} + (\boldsymbol{\beta}' - \Phi_{\mathbf{n}}(t_{\mathbf{k}'-1})) / S_{\mathbf{k}'}$\;
		$\mathbf{T}^{\mathrm{tr}} \gets \boldsymbol{\nu}\, T_{\mathbf{n}} + \boldsymbol{\tau}' - \boldsymbol{\tau}$
		\Comment*[r]{no interval walk}
		$\boldsymbol{\eta} \sim \mathcal{U}[\eta_{\min}, \eta_{\max}]^{E}$;\quad
		$\mathbf{t}^{\mathrm{delay}} \gets (\mathbf{T}^{\mathrm{tr}} + t_{\mathrm{RTT}}) \odot \boldsymbol{\eta}$\;
		$\mathbf{R} \gets (\mathbf{t}^{\mathrm{delay}} - \mathbf{B})_+$;\quad
		$\tilde{\mathbf{B}} \gets (\mathbf{B} - \mathbf{t}^{\mathrm{delay}})_+ + L$\;
		$\Delta\mathbf{t}^{w} \gets \delta \lceil (\tilde{\mathbf{B}} - B^{\max})_+ / \delta \rceil$;\quad
		$\mathbf{B} \gets \tilde{\mathbf{B}} - \Delta\mathbf{t}^{w}$\;
		$\boldsymbol{\tau} \gets (\boldsymbol{\tau}' + \Delta\mathbf{t}^{w}) \bmod T_{\mathbf{n}}$;\quad
		$\mathbf{i} \gets \mathbf{i} + 1$ \Comment*[r]{no waiting loop}
		\KwRet{observation $\mathbf{o}$}\;
	}
	\BlankLine
	\Fn(\Comment*[f]{reset envs where $\mathbf{M}$ holds}){\FStart{$\mathbf{M}$}}{
		$(\boldsymbol{\tau}_0, \mathbf{B}_0, \mathbf{i}_0) \gets (\boldsymbol{\tau}, \mathbf{B}, \mathbf{i})$\;
		$\mathbf{n} \gets \operatorname{where}(\mathbf{M}, \mathcal{U}\{1..N\}^{E}, \mathbf{n})$;\quad
		$\boldsymbol{\tau} \gets \operatorname{where}(\mathbf{M}, t_{\mathbf{p}}, \boldsymbol{\tau})$
		with $\mathbf{p} \sim \mathcal{U}\{0..K_{\mathbf{n}}-2\}$\;
		$(\mathbf{B}, \mathbf{i}, \mathbf{b}^{\mathrm{prev}}) \gets \operatorname{where}(\mathbf{M}, (0, 1, b^{(1)}), (\mathbf{B}, \mathbf{i}, \mathbf{b}^{\mathrm{prev}}))$\;
		$\mathbf{o} \gets$ \FDownload{$b^{(1)} \cdot \mathbf{1}$} \Comment*[r]{warm-up $V_1$, no reward}
		$(\boldsymbol{\tau}, \mathbf{B}, \mathbf{i}) \gets \operatorname{where}(\mathbf{M}, (\boldsymbol{\tau}, \mathbf{B}, \mathbf{i}), (\boldsymbol{\tau}_0, \mathbf{B}_0, \mathbf{i}_0))$\;
		\KwRet{$\mathbf{o}$}\;
	}
	\BlankLine
	\Fn(\Comment*[f]{one decision for all $E$ envs}){\FStep{$\mathbf{b}$}}{
		$\mathbf{o} \gets$ \FDownload{$\mathbf{b}$}\;
		$\mathbf{r} \gets \mu_1 \mathbf{b} - \mu_3 \mathbf{R} - \mu_2 \lvert \mathbf{b} - \mathbf{b}^{\mathrm{prev}} \rvert$;\quad
		$\mathbf{b}^{\mathrm{prev}} \gets \mathbf{b}$;\quad
		$\mathbf{done} \gets (\mathbf{i} > I)$\;
		\If(\Comment*[f]{only GPU-to-host transfer}){$\operatorname{any}(\mathbf{done})$}{
			$\mathbf{o} \gets \operatorname{where}(\mathbf{done}, \FStart{$\mathbf{done}$}, \mathbf{o})$\;
		}
		\KwRet{$\mathbf{o}, \mathbf{r}, \mathbf{done}$}\;
	}
\end{algorithm}

For a given state, action and noise realization $\eta_i$, the batched step produces the same $T_i^{\mathrm{tr}}$, $t_i^{\mathrm{delay}}$, $R_i$, $\Delta t_i^{w}$, $B_{i+1}$ and $\tau_{i+1}$ as the sequential simulator. The interval walk accumulates $S_k \Delta_{i,k}$ until the accumulated bytes exceed $d_i(b_i)$, which is exactly the root of $\bar{\Phi}_n(\tau_i + x) - \bar{\Phi}_n(\tau_i) = d_i(b_i)$ selected by the strict inequality above; since $\bar{\Phi}_n$ is non-decreasing, this root is unique. Cyclic replay is handled identically, as traces are shifted so that $t_0 = 0$. All remaining updates are the same element-wise formulas. The only sources of randomness, $\eta_i$ and the sampling at reset, are drawn from the same distributions in both simulators, so the two produce identically distributed trajectories, and identical trajectories when the noise and the reset samples are fixed.

\paragraph{\textbf{Scenarios:}}
The simulator does not fix the network or the video: the trace dataset $\mathcal{D}$ and the chunk sizes $d_i(b)$ are given when an environment is built. A scenario is therefore a choice of traces, namely one or more trace families with their train or test split, together with a video. We stream two standard test videos commonly used in ABR evaluation. For the 3G families (FCC-16, FCC-18, Oboe, Puffer-21, Puffer-22 and HSR), we use the \emph{Envivio-Dash3} video adopted by Pensieve, which is encoded into 4-second chunks at six bitrates from 300 to 4300\,kbps. For the families from other datasets (Lumos4G, Lumos5G, Norway3G, SolisWi-Fi, Ghent and Lab), most of which have a higher bandwidth, we use the open-source animated film \emph{Big Buck Bunny (BBB)}, encoded at higher bitrates and thus with considerably larger chunks~\cite{he2026nsma}. Each family provides a train and a test scenario, or a single one if it has no split, and the five 3G families are additionally pooled into one training and one test scenario. To build the GPU environment of a scenario, the selected traces are pooled into $\mathcal{D}$ and converted once into the flat arrays of Appendix~\ref{sec:gpu_sim}, after which $E$ environments sample their episodes from this pool in parallel. All random draws, i.e., the trace, the initial trace time and the delay noise, are computed by hashing the seed, the environment index, the episode and the chunk index rather than by advancing a sequential generator. The same data also builds a sequential CPU version of the scenario, which we keep as a reference: given the same seed, a CPU environment with index $e$ plays exactly the same episodes as the $e$-th GPU environment.

\begin{table}[!htbp]
\centering
\caption{Scenarios: trace families, the video used with each, and the number of traces per split. The 3G pool scenarios combine the first five families.}
\label{tab:scenarios}
\small
\begin{tabular}{llrrr}
\toprule
Family & Video & Train & Test & Mean BW (Mbps) \\
\midrule
FCC-16     & Envivio & 151 & 50  & 1.4 \\
FCC-18     & Envivio & 300 & 100 & 6.5 \\
Oboe       & Envivio & 328 & 100 & 2.8 \\
Puffer-21  & Envivio & 500 & 100 & 1.9 \\
Puffer-22  & Envivio & 500 & 100 & 1.6 \\
HSR        & Envivio & --  & 34$^{*}$ & 7.6 \\
\midrule
3G pool    & Envivio & 1779 & 450 & 2.7 \\
\midrule
Lumos4G    & Big Buck Bunny & 122 & 53  & 40.6 \\
Lumos5G    & Big Buck Bunny & 84  & 37  & 476.8 \\
Norway3G   & Big Buck Bunny & 93  & 41  & 2.1 \\
SolisWi-Fi & Big Buck Bunny & 56  & 24  & 24.2 \\
Ghent      & Big Buck Bunny & --  & 40$^{*}$ & 31.5 \\
Lab        & Big Buck Bunny & --  & 61$^{*}$ & 59.8 \\
\bottomrule
\multicolumn{5}{l}{\footnotesize $^{*}$No train/test split; the whole family forms a single scenario.}
\end{tabular}
\end{table}

\paragraph{\textbf{Consistency between the CPU and GPU simulators:}}
To check that the GPU simulator can replace the sequential one, we evaluate four rule-based agents (BBA, BOLA, RobustMPC and FDASH) on both simulators with the same seed and compare every episode. These agents have no trainable parameters and choose their actions deterministically from the observation, so any difference in the results can only come from the simulators.

\begin{table*}[!htbp]
\centering
\caption{Mean per-chunk QoE and its components over 200 episodes on the CPU and
GPU simulators with the same seed. For every episode, all four values are
identical on the two simulators.}
\label{tab:cpu_gpu_qoe}
\footnotesize
\begin{tabular}{lllrrrr}
\toprule
Scenario & Agent & Simulator & QoE & Bitrate & Rebuffering & Smoothness \\
\midrule
FCC-16-Train & BBA & CPU & 0.63260355 & 1.06820747 & $-0.09097625$ & $-0.34462765$ \\
 &  & GPU & 0.63260355 & 1.06820747 & $-0.09097625$ & $-0.34462765$ \\
 & BOLA & CPU & $-1.48957165$ & 1.17953195 & $-1.35711421$ & $-1.31198936$ \\
 &  & GPU & $-1.48957165$ & 1.17953195 & $-1.35711421$ & $-1.31198936$ \\
 & RobustMPC & CPU & 0.78934009 & 1.17832981 & $-0.23560142$ & $-0.15338830$ \\
 &  & GPU & 0.78934009 & 1.17832981 & $-0.23560142$ & $-0.15338830$ \\
 & FDASH & CPU & 0.83310116 & 0.97379789 & $-0.07784035$ & $-0.06285638$ \\
 &  & GPU & 0.83310116 & 0.97379789 & $-0.07784035$ & $-0.06285638$ \\
\midrule
FCC-18-Train & BBA & CPU & 2.58657474 & 2.87049477 & $-0.00070194$ & $-0.28321808$ \\
 &  & GPU & 2.58657474 & 2.87049477 & $-0.00070194$ & $-0.28321808$ \\
 & BOLA & CPU & 1.91624428 & 3.01338309 & $-0.06393666$ & $-1.03320213$ \\
 &  & GPU & 1.91624428 & 3.01338309 & $-0.06393666$ & $-1.03320213$ \\
 & RobustMPC & CPU & 3.03212850 & 3.21009052 & $-0.03672266$ & $-0.14123936$ \\
 &  & GPU & 3.03212850 & 3.21009052 & $-0.03672266$ & $-0.14123936$ \\
 & FDASH & CPU & 2.76335144 & 2.84373944 & $-0.00045715$ & $-0.07993085$ \\
 &  & GPU & 2.76335144 & 2.84373944 & $-0.00045715$ & $-0.07993085$ \\
\midrule
Oboe-Train & BBA & CPU & 1.79409039 & 2.26301067 & $-0.00157984$ & $-0.46734043$ \\
 &  & GPU & 1.79409039 & 2.26301067 & $-0.00157984$ & $-0.46734043$ \\
 & BOLA & CPU & 0.62099099 & 2.40147880 & $-0.27324308$ & $-1.50724468$ \\
 &  & GPU & 0.62099099 & 2.40147880 & $-0.27324308$ & $-1.50724468$ \\
 & RobustMPC & CPU & 2.26628522 & 2.52273408 & $-0.04597545$ & $-0.21047340$ \\
 &  & GPU & 2.26628522 & 2.52273408 & $-0.04597545$ & $-0.21047340$ \\
 & FDASH & CPU & 1.98008904 & 2.08543619 & $-0.00334715$ & $-0.10200000$ \\
 &  & GPU & 1.98008904 & 2.08543619 & $-0.00334715$ & $-0.10200000$ \\
\bottomrule
\end{tabular}
\end{table*}

The essential point is that both simulators see the same randomness. The GPU simulator is created with a single seed $\omega$ and runs $E = 50$ environments; on the CPU side we create 50 separate environments, all with the same seed $\omega$, and give the $e$-th one the index $e$. Every random draw, i.e., the trace and the initial trace time of an episode and the delay noise of each chunk, is obtained by hashing $(\omega, e, \text{episode}, \text{chunk})$, so the pair $(\omega, e)$ acts as the seed of environment $e$ on both sides. Since the draws do not depend on the order in which they are made, running the 50 environments in parallel on the GPU or one after another on the CPU produces the same episodes. With a conventional sequential generator this would not hold, as a batched step consumes random numbers in a different order than a single environment does. Each environment plays four episodes, which gives 200 episodes per agent and scenario. Table~\ref{tab:cpu_gpu_qoe} reports the results on three training scenarios: in all 12 cases, the QoE and each of its components are identical on the two simulators for every episode.

\paragraph{\textbf{Simulation speed:}}
We measure the throughput of the simulators alone, with random actions drawn in advance, on an AMD Ryzen Threadripper 7960X CPU (24 cores, 48 hardware threads) and an NVIDIA RTX A6000 GPU. Throughput counts environment steps, i.e., chunk downloads, per second summed over all environments, and speedups are relative to a single CPU environment on one core (Table~\ref{tab:cpu_gpu_speed}). 

On a single core, adding more environments does not help, since they simply share the same core. Running environments in parallel processes, one per process, speeds things up almost linearly up to 16 processes. With all 48 hardware threads in use, the CPU reaches $4.2\times10^{6}$ steps/s, 24.6 times the single-core rate. A step of the GPU simulator takes about 3.4~ms for up to 65,536 environments, because it consists of a fixed sequence of small kernels whose launch overhead dominates. The GPU simulator therefore needs about a thousand environments to match a single core, matches the full CPU at 16,384 environments, and exceeds it by a factor of 4.5 and 11.7 with 65,536 and 262,144 environments.

\begin{table}[!htbp]
\centering
\caption{Simulation throughput on FCC-16-Train (environment steps per second
summed over all environments, median of three runs). Speedup is relative to
one CPU environment on one core.}
\label{tab:cpu_gpu_speed}
\small
\begin{tabular}{lrrrr}
\toprule
Simulator & Processes & Environments & Steps/s & Speedup \\
\midrule
CPU & 1  & 1   & $1.72\times10^{5}$ & 1.0 \\
CPU & 1  & 256 & $1.66\times10^{5}$ & 1.0 \\
CPU & 8  & 8   & $1.33\times10^{6}$ & 7.8 \\
CPU & 16 & 16  & $2.63\times10^{6}$ & 15.3 \\
CPU & 24 & 24  & $2.73\times10^{6}$ & 15.9 \\
CPU & 48 & 48  & $4.23\times10^{6}$ & 24.6 \\
\midrule
GPU & -- & 1       & $2.85\times10^{2}$ & 0.002 \\
GPU & -- & 1,024   & $2.92\times10^{5}$ & 1.7 \\
GPU & -- & 4,096   & $1.17\times10^{6}$ & 6.8 \\
GPU & -- & 16,384  & $4.72\times10^{6}$ & 27.5 \\
GPU & -- & 65,536  & $1.89\times10^{7}$ & 110.1 \\
GPU & -- & 262,144 & $4.93\times10^{7}$ & 286.9 \\
\bottomrule
\end{tabular}
\end{table}

\newpage

%% file: parallel_env_train.tex
\subsection{Training and Evaluating Policies with the GPU Simulator}
\paragraph{\textbf{Training speed:}}
The throughput of the simulator alone does not tell how much faster training becomes, since each training step also includes the forward pass of the policy and the PPO update. We therefore train Pensieve on 3G-Train-Pool with the PPO hyperparameters in Table~\ref{tab:ppo_params} and change only the environment backend: the batched GPU simulator, the CPU simulator stepped by one process, and the CPU simulator split over two worker processes.

\begin{table}[!htbp]
\centering
\caption{PPO hyperparameters used in the training-speed experiments on Pensieve. $E$ is the number of parallel environments.}
\label{tab:ppo_params}
\small
\begin{tabular}{lr}
\toprule
Hyperparameter & Value \\
\midrule
Rollout length per environment & 96 \\
Samples per PPO update & $96E$ \\
Minibatches per epoch & 16 \\
Epochs per update & 3 \\
Learning rate (Adam) & $8 \times 10^{-4}$ \\
Adam $\epsilon$ & $10^{-8}$ \\
Discount factor $\gamma$ & 0.99 \\
GAE parameter $\lambda$ & 0.95 \\
Clipping coefficient & 0.2 \\
Value loss coefficient & 0.5 \\
Entropy coefficient & 0.05 \\
Gradient norm clipping & 1.0 \\
\bottomrule
\end{tabular}
\end{table}

The policy and the PPO update run on the GPU (NVIDIA RTX A6000) in all cases, and the CPU backends copy their observations to the GPU in the same format as the GPU simulator. Each run performs six PPO updates of $96 \times E$ environment steps, where $E$ is the number of parallel environments; the first update is a warm-up and the remaining five are timed. To attribute the time to the environment step, the policy forward pass, the PPO update and the remaining bookkeeping, every phase is followed by a CUDA synchronization, which slightly lowers the throughput of the GPU backend.

\begin{table}[!htbp]
\centering
\caption{Training throughput of Pensieve with different environment backends (environment steps per second over five timed PPO updates). Speedup is relative to the CPU simulator with one process at the same number of environments. The last four columns give the share of the training time spent in each phase.}
\label{tab:train_speed}
\footnotesize
\setlength{\tabcolsep}{4pt}
\begin{tabular}{rlrrrrrr}
\toprule
Envs & Backend & Steps/s & Speedup & Env & Forward & PPO update & Other \\
\midrule
64    & GPU          & 8,345   & 0.40 & 59.5\% & 14.0\% & 23.9\% & 2.6\% \\
      & CPU, 1 proc. & 20,724  & 1.00 & 22.9\% & 27.4\% & 46.6\% & 3.2\% \\
\midrule
256   & GPU          & 29,815  & 0.65 & 56.4\% & 15.8\% & 24.7\% & 3.1\% \\
      & CPU, 1 proc. & 45,693  & 1.00 & 52.2\% & 18.3\% & 26.7\% & 2.8\% \\
\midrule
1,024 & GPU          & 125,546 & 1.62 & 57.5\% & 16.3\% & 23.1\% & 3.0\% \\
      & CPU, 1 proc. & 77,508  & 1.00 & 77.4\% &  8.0\% & 13.3\% & 1.2\% \\
      & CPU, 2 proc. & 100,535 & 1.30 & 67.3\% & 13.5\% & 16.9\% & 2.4\% \\
\midrule
4,096 & GPU          & 401,060 & 4.53 & 44.5\% & 10.4\% & 43.7\% & 1.4\% \\
      & CPU, 1 proc. & 88,586  & 1.00 & 87.4\% &  2.8\% &  9.3\% & 0.5\% \\
      & CPU, 2 proc. & 138,797 & 1.57 & 78.1\% &  6.6\% & 14.3\% & 1.0\% \\
\bottomrule
\end{tabular}
\end{table}

Table~\ref{tab:train_speed} shows that the GPU simulator only pays off with enough environments. We convert the shares in the table into the time per sample, i.e., per environment step, by dividing the share of a phase by the throughput. Each step of the GPU simulator has a fixed cost of about 4.5~ms regardless of the number of environments ($64 \times 59.5\% / 8{,}345 \approx 4{,}096 \times 44.5\% / 401{,}060 \approx 4.5$~ms), dominated by kernel launches. With 64 environments, this cost is shared by few samples: the environment takes 71~$\mu$s per sample on the GPU ($59.5\% / 8{,}345$) against 11~$\mu$s on the CPU ($22.9\% / 20{,}724$), and training is 2.5 times slower ($1 / 0.40$). The two backends break even between 256 and 1,024 environments (speedup 0.65 and 1.62). With 4,096 environments, the environment takes 1.1~$\mu$s per sample on the GPU ($44.5\% / 401{,}060$) against 9.9~$\mu$s on the CPU ($87.4\% / 88{,}586$), and training is 4.5 times faster than with one CPU process and 2.9 times faster than with two ($401{,}060 / 138{,}797$). At this point the bottleneck has moved from the environment to the PPO update: the share of the environment drops from 87.4\% to 44.5\%, while that of the PPO update grows from 9.3\% to 43.7\%. In absolute terms the PPO update costs the same on both backends, about 1.1~$\mu$s per sample ($43.7\% / 401{,}060$ and $9.3\% / 88{,}586$), since it runs on the same GPU. The phases other than the environment, i.e., the forward pass, the PPO update and the bookkeeping, take the same time on both backends, 1.4~$\mu$s per sample ($(10.4\% + 43.7\% + 1.4\%) / 401{,}060$) on the GPU. Even if the environment were free, a sample would still cost 1.4~$\mu$s, whereas one CPU process needs 11.3~$\mu$s per sample in total ($1 / 88{,}586$). So training can be at most about eight times faster than with one CPU process ($11.3 / 1.4$).

\paragraph{\textbf{Evaluation protocol:}}
The QoE of a policy varies strongly from one episode to another, since the episodes differ in their trace, initial trace time and delay noise. The mean QoE reported for a policy is therefore itself a random quantity, and its precision depends on how many episodes it averages. We determine how many episodes are needed for a stable result, how to distribute them over parallel environments, and how long the evaluation takes on the GPU and on the CPU.

We evaluate a fully trained Pensieve policy with greedy actions on 3G-Test-Pool. The QoE of an episode is its mean per-chunk reward, and the evaluation result is the mean QoE over $N$ episodes. For each $N \in \{512, 1024, \dots, 16384\}$, we repeat the evaluation with 10 different evaluation seeds, each of which draws a different set of episodes, and take the standard deviation of the 10 results as the evaluation noise at that $N$. If the episodes are independent, this noise should equal $\sigma / \sqrt{N}$, where $\sigma$ is the standard deviation of the QoE of a
single episode. On the GPU, the $N$ episodes are run with $\min(N, 4096)$ parallel environments. In addition, we fix $N = 8192$ and distribute the episodes over 8192, 4096, 1024 and 256 environments to test whether the split matters. For the speed comparison, the CPU baseline is the conventional evaluation that runs one environment at a time with the policy on the CPU; its throughput is measured on 64 episodes and its time for $N$ episodes is extrapolated linearly, since the episodes are evaluated one after another. GPU times are measured on an NVIDIA RTX A6000.

\begin{table}[!htbp]
\centering
\caption{Evaluation noise and time for $N$ evaluation episodes. Std and range are taken over the mean QoE of 10 evaluation seeds; $\sigma/\sqrt{N}$ uses the per-episode QoE standard deviation $\sigma = 1.70$. CPU times are extrapolated from 64 measured episodes.}
\label{tab:eval_noise}
\footnotesize
\setlength{\tabcolsep}{4pt}
\begin{tabular}{rrrrrrrr}
\toprule
$N$ & Envs $\times$ episodes & Std & $\sigma/\sqrt{N}$ & Range & GPU time (s) & CPU time (s) & Speedup \\
\midrule
512    & $512 \times 1$  & 0.065 & 0.075 & 0.188 & 0.40 & 13.8  & 35 \\
1,024  & $1024 \times 1$ & 0.058 & 0.053 & 0.181 & 0.38 & 27.6  & 73 \\
2,048  & $2048 \times 1$ & 0.041 & 0.038 & 0.158 & 0.46 & 55.1  & 120 \\
4,096  & $4096 \times 1$ & 0.033 & 0.027 & 0.116 & 0.41 & 110.2 & 267 \\
8,192  & $4096 \times 2$ & 0.026 & 0.019 & 0.067 & 0.53 & 220.5 & 415 \\
16,384 & $4096 \times 4$ & 0.016 & 0.013 & 0.057 & 1.44 & 441.0 & 307 \\
\bottomrule
\end{tabular}
\end{table}

\begin{figure}[!htbp]
\centering
\includegraphics[width=0.5\linewidth]{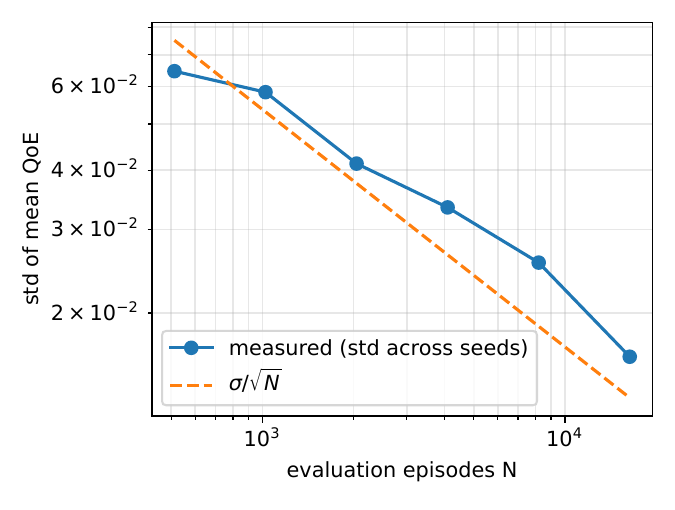}
\caption{Standard deviation of the mean QoE over 10 evaluation seeds versus the number of evaluation episodes $N$ (log--log), together with $\sigma/\sqrt{N}$.}
\label{fig:eval_noise}
\end{figure}

A single episode is a very noisy measurement: its QoE has a standard deviation of $\sigma = 1.70$, while the mean QoE of the policy is about 1.67. The noise of the evaluation result decreases with the number of episodes as expected from averaging independent episodes (Table~\ref{tab:eval_noise} and Figure~\ref{fig:eval_noise}): from 0.065 with 512 episodes to 0.016 with 16,384, roughly halving whenever $N$ is quadrupled, in line with $\sigma/\sqrt{N}$. With few episodes, the result is unreliable: with 512 to 2,048 episodes, the results of the 10 seeds spread over 0.158 to 0.188.

With 8,192 episodes, the noise is 0.026, about 1.5\% of the mean, and the 10 seeds lie within 0.067. How the episodes are distributed over environments does not matter (Table~\ref{tab:eval_split}): for $N = 8192$, the noise ranges from 0.015 to 0.027 without any trend in the split, which is within the precision of a 10-seed estimate, and the mean QoE agrees within 0.013. This is expected, since every episode draws its trace, initial trace time, and noise independently, and the policy clears its history at the start of each episode.

The number of environments therefore only affects the speed. Up to 4,096 episodes, all episodes run in parallel and the GPU evaluation takes 0.38--0.46~s regardless of $N$ (Table~\ref{tab:eval_noise}); beyond that, each environment plays several episodes in sequence, and the time increases to 0.53~s for 8,192 and 1.44~s for 16,384 episodes. The CPU baseline evaluates 37 episodes per second (64 episodes measured), so the same evaluations take from 14~s ($512 / 37$) to 441~s ($16{,}384 / 37$), and the GPU evaluation is 35 to 415 times faster.

\begin{table}[!htbp]
\centering
\caption{Evaluation noise for $N = 8192$ episodes distributed differently over
parallel environments (10 evaluation seeds).}
\label{tab:eval_split}
\small
\begin{tabular}{rrrr}
\toprule
Envs & Episodes per env & Mean QoE & Std \\
\midrule
8,192 & 1  & 1.659 & 0.027 \\
4,096 & 2  & 1.664 & 0.026 \\
1,024 & 8  & 1.664 & 0.015 \\
256   & 32 & 1.651 & 0.019 \\
\bottomrule
\end{tabular}
\end{table}

We therefore evaluate every policy with 8,192 episodes, run as 2 episodes in each of 4,096 parallel GPU environments, with the same fixed evaluation seed for all policies. This gives an evaluation noise of about 0.026, i.e., 1.5\% of the mean QoE, in about 0.5~s; going further would require four times as many episodes to halve the noise. Using the same seed evaluates all policies on the same set of episodes, so their comparison is not affected by which traces happen to be drawn.

\paragraph{\textbf{Training efficiency:}}
The throughput measurements above do not show directly how much the GPU simulator shortens training, i.e., how good a policy becomes within a given
time. We therefore train Pensieve on 3G-Train-Pool with the PPO hyperparameters in Table~\ref{tab:ppo_params} and 256, 1,024 or 4,096 parallel environments, once with the GPU simulator for 2 minutes and once with the CPU simulator for 12 minutes, to also measure how long the CPU simulator needs to catch up. The CPU simulator runs in a single process on one core; as before, the policy and the PPO update run on the GPU in both cases, so the two settings differ only in the environment. Every configuration is trained with 5 seeds. Every 15~s of training time, the current policy is evaluated with the protocol established above (8,192 episodes of 3G-Test-Pool with a fixed evaluation seed), and the evaluation time is not counted as training time.

\begin{figure*}[!htbp]
\centering
\includegraphics[width=\linewidth]{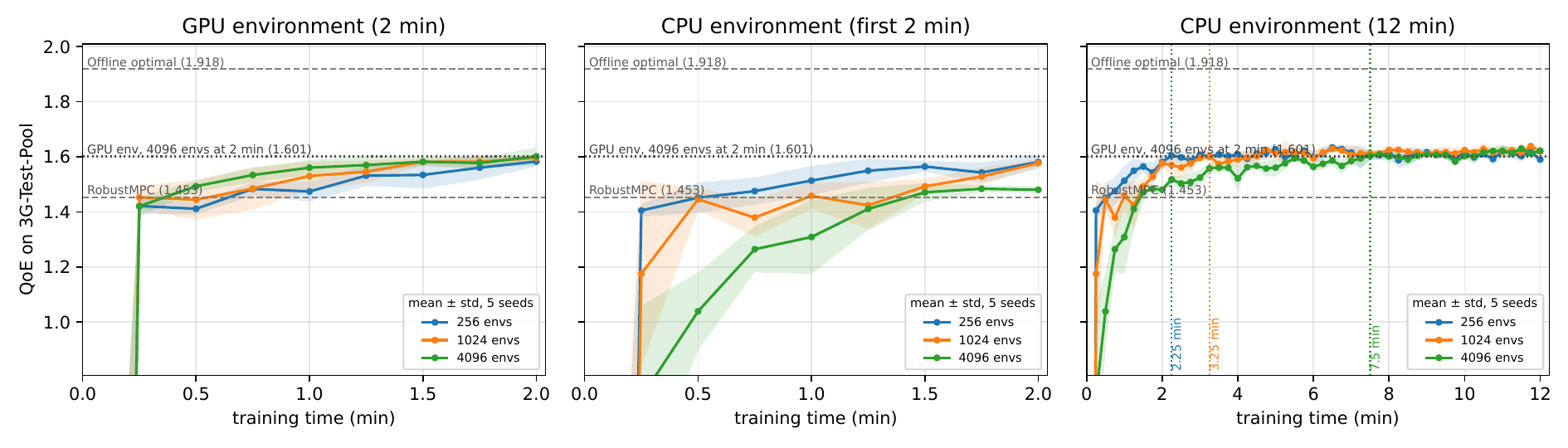}
\caption{QoE on 3G-Test-Pool versus training time for Pensieve with 256, 1,024 and 4,096 parallel environments: GPU simulator over 2 minutes (left), CPU simulator over the same first 2 minutes (middle) and over 12 minutes (right). Lines show the mean over 5 training seeds and shaded bands the mean $\pm$ standard deviation across the seeds at each evaluation. Dashed lines mark the offline optimum and RobustMPC; the dotted line marks the QoE reached by the GPU simulator with 4,096 environments after 2 minutes (1.601), and the vertical lines in the right panel mark when the mean CPU curves first reach it.}
\label{fig:train_scaling}
\end{figure*}

Figure~\ref{fig:train_scaling} also shows two reference levels on the same
episodes. RobustMPC reaches a QoE of 1.453. The offline optimum, 1.918, is an
upper bound that no deployable policy can reach. Every random quantity of an
episode, namely its trace, initial trace time and delay noise, is fixed by the
evaluation seed and does not depend on the actions. The future bandwidth of
every episode is therefore known in advance, and the bitrate sequence that
maximizes the QoE of each episode can be found by dynamic programming over the
chunk index, the buffer occupancy and the previous bitrate. We solve it for
all 8,192 evaluation episodes in about 2.5 minutes, running the recursion on
the GPU simulator.

With the GPU simulator (Figure~\ref{fig:train_scaling}, left), all three configurations reach a QoE of 1.58 to 1.60 within 2 minutes, and more environments converge faster: after 1 minute, 4,096 environments reach 1.560, compared with 1.474 for 256. With the CPU simulator (middle), this order is reversed. With 4,096 environments, training reaches only 1.308 after 1 minute and 1.480 after 2 minutes, barely above RobustMPC, whereas 256 environments reach 1.581 after 2 minutes, close to the GPU simulator. Over the full 12 minutes (right), the mean CPU curves first reach the QoE of the GPU simulator with 4,096 environments after 2.25, 3.25 and 7.5 minutes with 256, 1,024 and 4,096 environments. With 4,096 environments, the CPU simulator thus needs 3.75 times as long; across the 5 seeds, it needs between 5.0 and 8.0 minutes. Training has essentially converged after 2 minutes: the QoE of 1.601 reached by the GPU simulator is 0.148 above RobustMPC and closes 32\% of the gap between RobustMPC and the offline optimum ($0.148 / 0.465$); continuing on the CPU for up to 12 minutes adds little, with the curves averaging 1.60 to 1.62 between 6 and 12 minutes.

These results agree with the training-speed measurements in Table~\ref{tab:train_speed}. The GPU simulator pays off only with many environments: it trains 4.5 times faster than one CPU process with 4,096 environments, but more slowly with 256 or fewer, because each of its steps has a fixed launch cost. During training, moreover, a large part of the time is spent on the policy and the PPO update, which run on the GPU in both cases, so a faster environment shortens only part of each training step. Evaluation is different: it consists only of environment steps and forward passes, and thousands of episodes can run in parallel, so the environment dominates and the GPU simulator evaluates 8,192 episodes in 0.53~s instead of about 220~s (Table~\ref{tab:eval_noise}). This speed matters for our framework, in which candidate policies are evaluated repeatedly and the results are passed to the LLM that decides which policy to use; the evaluation must therefore be fast and, at the same time, precise. With 8,192 episodes, the evaluation noise is 0.026, whereas an evaluation limited to about 500 episodes, as is affordable on the CPU, has a noise of about 0.065 and cannot separate policies whose QoE differs by less than about 0.1.

\begin{table}[!htbp]
\centering
\caption{Summary of the GPU simulator's speed relative to the CPU simulator. CPU throughput is for one core and for all 48 hardware threads; the CPU evaluation time is extrapolated from 64 measured episodes.}
\label{tab:gpu_summary}
\small
\begin{tabular}{lll}
\toprule
Task & GPU & CPU \\
\midrule
Evaluating 8,192 episodes & 0.53~s & 220~s ($415\times$) \\
Simulation throughput & $4.9 \times 10^{7}$ steps/s & $1.7 \times 10^{5}$ steps/s, 1 core ($287\times$) \\
\quad (262,144 environments) & & $4.2 \times 10^{6}$ steps/s, 48 threads ($12\times$) \\
Training to QoE 1.601 & 2~min & 7.5~min ($3.75\times$) \\
\quad (4,096 environments) & & \\
\bottomrule
\end{tabular}
\end{table}

In summary, the GPU simulator brings its largest gain to evaluation, where it is two to three orders of magnitude faster than the CPU simulator, and enables analyses such as computing the offline optimum for the whole evaluation set. For training, it helps when many environments run in parallel, reaching the same QoE 3.75 times faster with 4,096 environments, but for a model of Pensieve's size the training converges within 2 minutes either way, and a CPU simulator with few environments trains nearly as fast. The GPU simulator thus matters most where evaluation is frequent, as when an LLM selects among policies based on their evaluated performance.

\newpage

%% file: policy_converge.tex
\section{Training to Convergence and Per-Scenario Evaluation}
\label{app:converge}

Because the batched environment makes long training runs and large evaluations cheap, we use it to ask two questions. First, when agents with different architectures are trained until their QoE stops improving, how far does each get, and how far apart do they end up? Second, do agents that score the same on a pooled test set also behave the same on the trace families inside the pool and on scenarios outside the training distribution?

We train Pensieve, Merina and PaMoE on 3G-Train-Pool with five training seeds each. Every run uses 4096 parallel environments and the PPO hyperparameters of Table~\ref{tab:ppo_params}, and performs 800 PPO updates. With a rollout length of 96 steps, one update consumes $96 \times 4096 = 393{,}216$ environment steps, so a run covers about $3.1 \times 10^{8}$ steps. Merina is trained in its two stages: its VAE is trained first for 300 updates, and a new policy is then trained on the frozen VAE for 800 updates, as for the other agents.

\begin{figure}[!htbp]
  \centering
  \includegraphics[width=0.5\columnwidth]{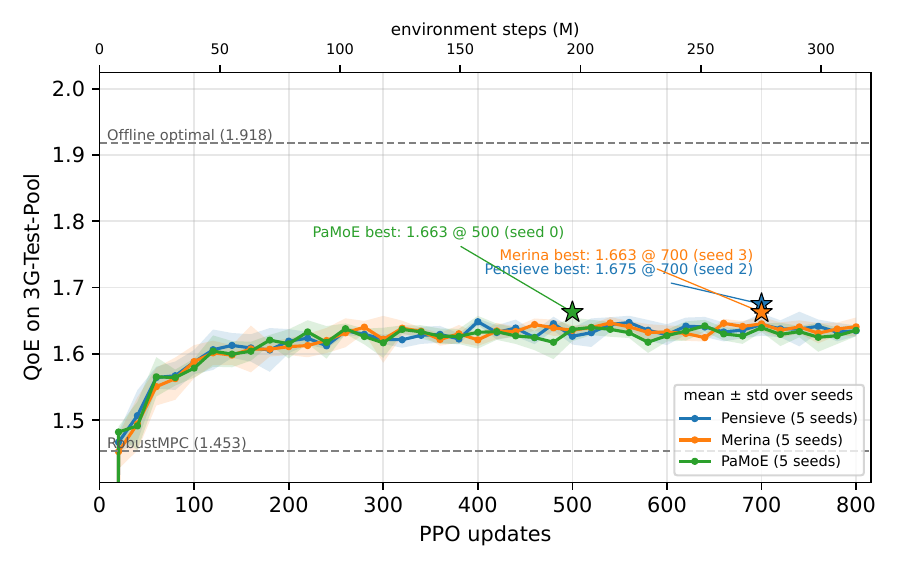}
  \caption{QoE on 3G-Test-Pool against the number of PPO updates (top axis: environment steps), as mean $\pm$ standard deviation over five training seeds. All points, RobustMPC and the offline optimum are evaluated on the same $4096 \times 2$ episodes (evaluation seed 123). Stars mark the best single checkpoint of each agent.}
  \label{fig:converge}
\end{figure}

We measure training progress in PPO updates. All agents use the same number of environments and the same rollout length, so after the same number of updates they have seen the same amount of data and taken the same number of gradient steps. Every 20 updates we evaluate the policy (greedy) on 3G-Test-Pool under the evaluation protocol (4096 environments $\times$ 2 episodes) and save a checkpoint. All these evaluations use the same evaluation seed (123), so every point of a curve is measured on the same episodes and the points can be compared directly. RobustMPC and the offline optimum in Figure~\ref{fig:converge} and Table~\ref{tab:converge} are computed on the same episodes.

Figure~\ref{fig:converge} shows the QoE of the three agents against the number of PPO updates. QoE rises steeply during the first 100 updates. After about 260 updates, the seed-averaged curves of all three agents stay between 1.61 and 1.65, and their $\pm 1$ standard-deviation bands overlap throughout. At the end of training the three agents are within 0.006 of each other (1.635, 1.641 and 1.635; Table~\ref{tab:converge}). This is smaller than the spread across training seeds of any one agent (standard deviation 0.008 to 0.014).

\begin{table}[!htbp]
  \centering
  \caption{Training to convergence on 3G-Train-Pool, evaluated on 3G-Test-Pool (evaluation seed 123). Final QoE: mean $\pm$ standard deviation over five training seeds after 800 updates. Best checkpoint: highest QoE of a single checkpoint over all seeds and evaluations. Time: training time per run in seconds, without evaluations; for Merina, the time of the VAE stage is given in parentheses.}
  \label{tab:converge}
  \small
  \setlength{\tabcolsep}{2.5pt}
  \begin{tabular}{lccc}
    \toprule
    Method & Final QoE & \shortstack{Best checkpoint\\(seed, updates)} & Time (s) \\
    \midrule
    Pensieve        & 1.635 $\pm$ 0.008 & 1.675 (2, 700) & 818 \\
    Merina          & 1.641 $\pm$ 0.014 & 1.663 (3, 700) & 839 (+306) \\
    PaMoE           & 1.635 $\pm$ 0.009 & 1.663 (0, 500) & 1731 \\
    \midrule
    RobustMPC       & 1.453 & -- & -- \\
    Offline optimum & 1.918 & -- & -- \\
    \bottomrule
  \end{tabular}
\end{table}

All three agents end well above RobustMPC (by 0.182 to 0.188) and well below the offline optimum (by 0.277 to 0.283). With the same inputs, training distribution and update budget, the components that Merina and PaMoE add to Pensieve (a VAE that encodes the network condition, and a mixture of expert policies) therefore do not raise the QoE that PPO reaches. The remaining gap to the optimum is about the same for all three. \textbf{Whether this conclusion extends beyond the pooled traces, to individual network scenarios and to out-of-distribution conditions, remains to be examined; we investigate this question in the following analysis.}

\paragraph{\textbf{Models and evaluation protocol:}}
For each agent, we keep the single checkpoint that achieves the highest QoE across all seeds and evaluations, marked by a star in Figure~\ref{fig:converge}: Pensieve reaches its best QoE of 1.675 with seed 2 after 700 updates, Merina reaches 1.663 with seed 3 after 700 updates, and PaMoE reaches 1.663 with seed 0 after 500 updates. We keep one model per agent rather than averaging over seeds, because a deployed system would run a single model.

Each model is then evaluated with five evaluation seeds (123 to 127), each giving a separate set of $4096 \times 2$ episodes. For every evaluation seed, RobustMPC and the offline optimum are computed on the same episodes as the agents. We report the mean and the standard deviation over the five seeds. Since all methods see the same episodes under a given seed, we also compare methods seed by seed before calling a difference real. The checkpoints were chosen on 3G-Test-Pool with evaluation seed 123, so their scores on this pool could be optimistic. Over the five evaluation seeds, however, the pool scores of the three models (1.676, 1.661 and 1.664; Table~\ref{tab:breakdown}) are within 0.002 of the scores at which they were chosen. This selection therefore did not noticeably inflate the in-distribution results. The out-of-distribution scenarios were not used for selection at all.

\begin{table*}[!htbp]
  \centering
  \caption{QoE per scenario: mean $\pm$ standard deviation over five evaluation seeds (123 to 127) of $4096 \times 2$ episodes each. The offline optimum is the mean over the same five seeds. Bold: best method other than the offline optimum. Trace families are the test splits.}
  \label{tab:breakdown}
  \providecommand{\qs}[2]{#1\,{\scriptsize$\pm$#2}}
  \small
  \setlength{\tabcolsep}{5pt}
  \begin{tabular}{llccccc}
    \toprule
    Scenario & Video & Pensieve & Merina & PaMoE & RobustMPC & Offline optimum \\
    \midrule
    \multicolumn{7}{l}{\emph{Within the training distribution}} \\
    FCC-16 & Envivio & \qs{0.699}{0.020} & \qs{\textbf{0.702}}{0.020} & \qs{0.697}{0.021} & \qs{0.548}{0.020} & 0.878 \\
    FCC-18 & Envivio & \qs{3.113}{0.012} & \qs{\textbf{3.123}}{0.011} & \qs{3.119}{0.010} & \qs{3.057}{0.012} & 3.265 \\
    Oboe & Envivio & \qs{\textbf{2.168}}{0.016} & \qs{2.156}{0.016} & \qs{2.155}{0.016} & \qs{2.062}{0.017} & 2.339 \\
    Puffer-21 & Envivio & \qs{\textbf{1.008}}{0.008} & \qs{0.980}{0.012} & \qs{0.985}{0.011} & \qs{0.544}{0.013} & 1.423 \\
    Puffer-22 & Envivio & \qs{\textbf{0.950}}{0.016} & \qs{0.918}{0.014} & \qs{0.924}{0.013} & \qs{0.653}{0.017} & 1.189 \\
    3G-Test-Pool & Envivio & \qs{\textbf{1.676}}{0.006} & \qs{1.661}{0.006} & \qs{1.664}{0.005} & \qs{1.448}{0.009} & 1.917 \\
    \midrule
    \multicolumn{7}{l}{\emph{Outside the training distribution}} \\
    HSR & Envivio & \qs{\textbf{3.059}}{0.016} & \qs{3.046}{0.017} & \qs{3.047}{0.014} & \qs{2.929}{0.021} & 3.633 \\
    Norway3G & BBB & \qs{$-$0.505}{0.020} & \qs{$-$0.256}{0.016} & \qs{\textbf{$-$0.153}}{0.012} & \qs{$-$0.630}{0.020} & 0.141 \\
    Lumos4G & BBB & \qs{2.292}{0.015} & \qs{2.247}{0.023} & \qs{1.698}{0.014} & \qs{\textbf{2.772}}{0.017} & 3.299 \\
    Lumos5G & BBB & \qs{1.180}{0.004} & \qs{1.583}{0.008} & \qs{1.048}{0.005} & \qs{\textbf{3.807}}{0.001} & 4.209 \\
    SolisWi-Fi & BBB & \qs{1.763}{0.010} & \qs{1.624}{0.013} & \qs{1.137}{0.009} & \qs{\textbf{2.152}}{0.010} & 2.365 \\
    Ghent & BBB & \qs{2.768}{0.005} & \qs{2.018}{0.022} & \qs{1.965}{0.007} & \qs{\textbf{3.028}}{0.007} & 3.468 \\
    Lab & BBB & \qs{3.358}{0.007} & \qs{3.168}{0.024} & \qs{2.494}{0.009} & \qs{\textbf{3.662}}{0.006} & 3.988 \\
    \bottomrule
  \end{tabular}
\end{table*}

\begin{figure*}[!htbp]
  \centering
  \includegraphics[width=0.75\textwidth]{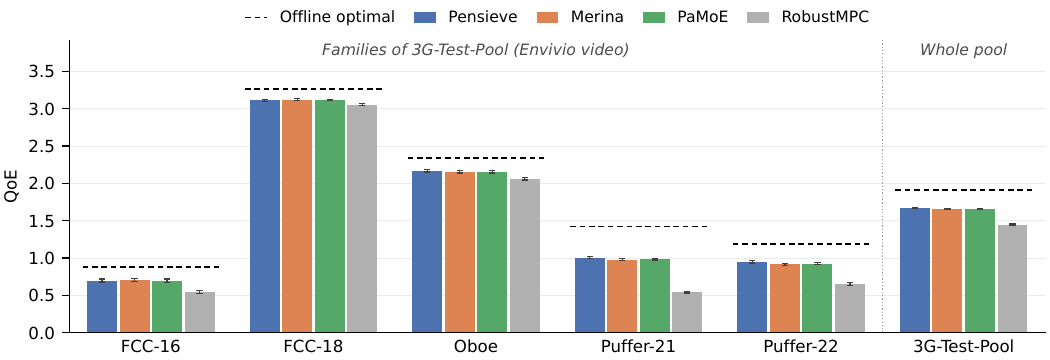}
  \caption{QoE of the best checkpoint of each agent on the five trace families of 3G-Test-Pool and on the whole pool (test splits, Envivio video). Bars and error bars: mean and standard deviation over five evaluation seeds of $4096 \times 2$ episodes each. The dashed lines show the offline optimum, averaged over the same five seeds.}
  \label{fig:breakdown-in}
\end{figure*}

\paragraph{\textbf{Within the training distribution:}}
3G-Test-Pool consists of the test traces of five families: FCC-16, FCC-18, Oboe, Puffer-21 and Puffer-22. A pooled average could hide an agent that is better on one family and worse on another, so we evaluate the models on each family separately (Figure~\ref{fig:breakdown-in}, Table~\ref{tab:breakdown}). 

On FCC-16, FCC-18 and Oboe the three agents are within 0.013 of each other. On Puffer-21 and Puffer-22, Pensieve is ahead of the other two by 0.023 to 0.032, and it is ahead under each of the five evaluation seeds. Every agent beats RobustMPC on every family. The best agent's lead ranges from 0.066 on FCC-18 to 0.464 on Puffer-21. The gap to the offline optimum varies with the family: from the best method it is 0.142 on FCC-18 and 0.171 on Oboe, but 0.415 on Puffer-21. Within the training distribution, then, the three agents perform almost identically on every family.

\begin{figure*}[!htbp]
  \centering
  \includegraphics[width=0.75\textwidth]{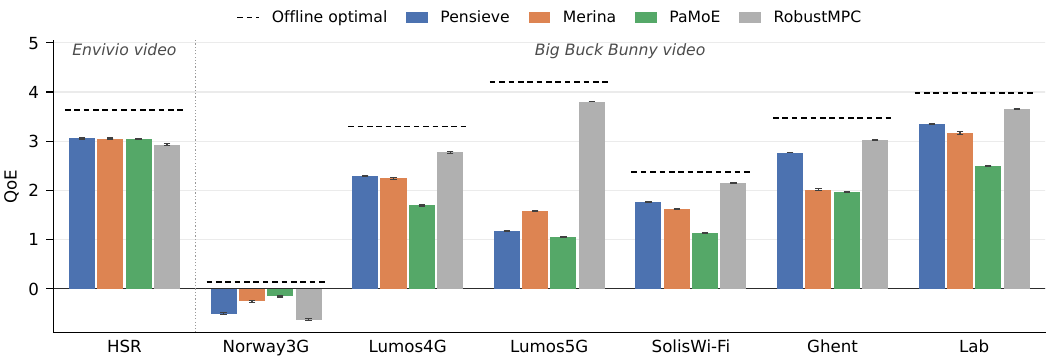}
  \caption{QoE of the same checkpoints on scenarios outside the training distribution: HSR (unseen traces, Envivio video) and six scenarios with unseen traces and the Big Buck Bunny video. Bars and error bars: mean and standard deviation over five evaluation seeds of $4096 \times 2$ episodes each. The dashed lines show the offline optimum, averaged over the same five seeds.}
  \label{fig:breakdown-ood}
\end{figure*}

\paragraph{\textbf{Outside the training distribution:}}
We evaluate the same models on two groups of unseen scenarios (Figure~\ref{fig:breakdown-ood}, Table~\ref{tab:breakdown}). HSR \cite{meng2019pitree} uses the same Envivio video as in training, so only the network changes. Norway3G, Lumos4G, Lumos5G, SolisWi-Fi, Ghent \cite{van2016http} and Lab \cite{meng2019pitree} come from other trace datasets, including 4G, 5G and Wi-Fi traces, and use the Big Buck Bunny video. At the same bitrate level, its chunks are on average 3.3 to 9.2 times larger than those of the Envivio video, so both the network and the video change.

On HSR the picture is the same as within the training distribution. The agents are within 0.013 of each other, Pensieve is ahead under each evaluation seed, and every agent beats RobustMPC, by 0.117 to 0.130. On the Big Buck Bunny scenarios the agents differ substantially. The gap between the best and the worst agent ranges from 0.352 on Norway3G to 0.864 on Lab, and the ranking changes from scenario to scenario: PaMoE is the best agent on Norway3G but the worst on the other five scenarios; Merina is the best on Lumos5G; Pensieve is the best on Lumos4G, SolisWi-Fi, Ghent and Lab.

RobustMPC beats all three agents on five of the six scenarios. Its lead over the best agent ranges from 0.260 on Ghent to 2.224 on Lumos5G. The exception is Norway3G: every method has a negative QoE there, RobustMPC the lowest ($-0.630$), and even the offline optimum reaches only 0.141. Unlike the learned policies, RobustMPC plans with the chunk sizes of the video being played and a throughput estimate, and nothing in it depends on the conditions seen during training. \textbf{Agents that are indistinguishable within the training distribution can therefore behave very differently outside it, and a test set drawn from the training distribution alone does not reveal this.}

\paragraph{\textbf{A hint for routing:}}
No single method in our evaluation is the best everywhere. Table~\ref{tab:routing} shows, for each group of scenarios, the average QoE of every method. It also shows the average obtained by choosing, in each scenario, the best of the three agents (Pensieve, Merina, PaMoE) or the best of all four methods (Pensieve, Merina, PaMoE, RobustMPC).

\begin{table*}[!htbp]
  \centering
  \caption{Unweighted average QoE over the scenarios of each group in Table~\ref{tab:breakdown} (within: the five trace families, without the pool; outside: all seven scenarios), so every scenario counts equally; this differs from the pool, where families are weighted by their number of traces. Best agent / best method per scenario: average of the highest QoE in each scenario among the three agents / among all four methods; the choice uses knowledge of the scenario, so these are upper bounds.}
  \label{tab:routing}
  \small
  \begin{tabular}{lccccccc}
    \toprule
    Scenarios & Pensieve & Merina & PaMoE & RobustMPC & \shortstack{Best RL agent\\per scenario} & \shortstack{Best method\\per scenario} & \shortstack{Offline\\optimal} \\
    \midrule
    Within (5)  & 1.588 & 1.576 & 1.576 & 1.373 & 1.590 & 1.590 & 1.819 \\
    Outside (7) & 1.988 & 1.918 & 1.605 & 2.532 & 2.096 & 2.618 & 3.015 \\
    \bottomrule
  \end{tabular}
\end{table*}

Choosing the best method per scenario shows that learned policies and model-based control are complementary across conditions, but this oracle choice is only an upper bound that still falls short of the offline optimum, so switching between them merits further study.

\newpage